%% file: planerectr++.tex
\documentclass[10pt,journal,compsoc]{IEEEtran}

\usepackage{graphicx}
\usepackage{amsmath}
\usepackage{amssymb}
\usepackage{booktabs}

\usepackage{multirow}
\usepackage{xcolor}
\usepackage{array}
\usepackage{multirow}
\usepackage{mathtools}
\usepackage{overpic}
\usepackage{stmaryrd}
\usepackage{url}
\usepackage{xspace}
\usepackage[ruled]{algorithm2e}

\usepackage{hyperref}
\usepackage{times}
\usepackage{epsfig}
\usepackage{bbm}
\usepackage{bbding}
\usepackage{subfigure}
\usepackage{amsthm}
\usepackage{makecell}
\usepackage{mathrsfs}

\newcommand{\compresslist}{%
 \setlength{\itemsep}{0pt}%
 \setlength{\parskip}{0pt}%
 \setlength{\parsep}{0pt}%
 }
\usepackage[capitalize]{cleveref}
\crefname{section}{Sec.}{Secs.}
\Crefname{section}{Section}{Sections}
\crefname{table}{Tab.}{Tabs.}
\Crefname{table}{Table}{Tables}
\crefname{figure}{Fig.}{Figs.}
\Crefname{figure}{Figure}{Figures}
\crefname{equation}{Eq.}{Eqs.}
\Crefname{equation}{Equation}{Equations}
\crefname{appendix}{Appx.}{Appxs.}
\Crefname{Appendix}{Appendix}{Appendices}
\crefname{algorithm}{Alg.}{Algs.}
\Crefname{algorithm}{Algorithm}{Algorithms}

\ifCLASSOPTIONcompsoc
  \usepackage[nocompress]{cite}
\else
  \usepackage{cite}
\fi

\input{mycommand}

\begin{document}


\title{PlaneRecTR++: Unified Query Learning for Joint 3D Planar Reconstruction and Pose Estimation}

%
%
%
%

\author{
Jingjia~Shi*, Shuaifeng~Zhi*$^\dag$, Kai~Xu$^\dag$
\thanks{* The first two authors contributed equally to this work.}
\thanks{$^\dag$ Shuaifeng Zhi and Kai Xu are corresponding authors.}
\thanks{Jingjia Shi, Shuaifeng Zhi and Kai Xu are with National University of Defense Technology, Changsha, China.}
}

\IEEEtitleabstractindextext{%

\input{abstract}
}

\maketitle

\IEEEdisplaynontitleabstractindextext

%
\IEEEpeerreviewmaketitle

\input{intro}

\input{related}

\input{overview}

\input{planerectr}
\input{corrattn}
\input{experiments}
\input{2views_experiments}

\input{conclusion}

\input{acknowledgement}

\ifCLASSOPTIONcaptionsoff
  \newpage
\fi



%

\bibliographystyle{IEEEtran}
\bibliography{robotvision}

\input{biography}

\end{document}

%% file: mycommand.tex
\makeatletter
\DeclareRobustCommand\onedot{\futurelet\@let@token\@onedot}
\def\@onedot{\ifx\@let@token.\else.\null\fi\xspace}

\def\eg{\emph{e.g}\onedot}

\def\ie{\emph{i.e}\onedot}

\def\etc{\emph{etc}\onedot}

\makeatother

\SetKwInOut{KwParam}{Parameter}

\newcolumntype{P}[1]{>{\centering\arraybackslash}p{#1}}
\newlength{\wdth}

\newcommand{\ptitle}[1]{\noindent\textbf{#1}\hspace{5pt}}

\newcommand{\paraspace}{\vspace{0pt}}
\newcommand{\jiajia}[1]{\textcolor{black}{#1}}

%% file: abstract.tex
\begin{abstract}
The challenging task of 3D planar reconstruction from images involves several sub-tasks including frame-wise plane detection, segmentation, parameter regression and possibly depth prediction, along with cross-frame plane correspondence and relative camera pose estimation. Previous works adopt a divide and conquer strategy, addressing above sub-tasks with distinct network modules in a two-stage paradigm. Specifically, given an initial camera pose and per-frame plane predictions from the first stage, further exclusively designed modules relying on external plane correspondence labeling are applied to merge multi-view plane entities and produce refined camera pose. Notably, existing work fails to integrate these closely related sub-tasks into a unified framework, and instead addresses them separately and sequentially, which we identify as a primary source of performance limitations. Motivated by this finding and the success of query-based learning in enriching reasoning among semantic entities, in this paper, we propose PlaneRecTR++, a Transformer-based architecture, which for the first time unifies all tasks of multi-view planar reconstruction and pose estimation within a compact single-stage framework, eliminating the need for the initial pose estimation and supervision of plane correspondence. Extensive quantitative and qualitative experiments demonstrate that our proposed unified learning achieves mutual benefits across sub-tasks, achieving a new state-of-the-art performance on the public ScanNetv1, ScanNetv2, NYUv2-Plane, and MatterPort3D datasets. Codes are available at \url{https://github.com/SJingjia/PlaneRecTR-PP}.

\end{abstract}


%% file: intro.tex

\IEEEraisesectionheading{\section{Introduction}\label{sec:intro}}
\IEEEPARstart{I}{mmersing} in a virtual 3D world involves efficient reasoning of surrounding scenes, whose properties need to be frequently updated. Though tremendous efforts have been devoted to create authentic 3D geometry from multi-view image observations or even a single image, a trade-off still exists between reconstruction quality and efficiency, depending on the underlying scene representations. 
3D maps composed of sparse primitives such as point clouds are light-weight to maintain but lack topological structures, while dense geometry like volumetric grids and meshes are computationally intensive to acquire and maintain. In this spectrum, planar representation has been proven to be a reliable alternative, which is compact, efficient, expressive, and generalizable enough to be deployed ubiquitously as well. Therefore, in practice it would always be ideal to infer planar information purely from a video sequence, or even a single image.
Being a challenging and yet fundamentally ill-posed computer vision problem, single image plane recovery has been extensively researched and focused so far. 
Early attempts have been made using image processing techniques to extract low-level primitives such as line segments, vanishing points to extract planar structures from an input image \cite{delage2007automatic, lee2009geometric}. Furthermore, multi-view 3D plane reconstruction \cite{Salas-Moreno:etal:ISMAR2014,hsiao2018dense} are investigated where plane-related constraints are introduced to regularize both camera poses as well as dense geometry, \eg, the well-known Manhattan-world assumption \cite{li2023hong,zhou2024neural}.

\input{figures/teaser}

As Convolutional Neural Networks (CNNs) have become the mainstream paradigm to tackle computer vision problems in the past few years \cite{xia2022metalearning,xia2024blind}, their excellence in performance gradually spreads to plane estimation task. Pioneering works such as PlaneNet \cite{Liu:etal:CVPR2018:Planenet} and PlaneRCNN \cite{Liu:etal:CVPR2019:Planercnn} propose an efficient solution of piece-wise planar structure recovery from a single image using CNNs in a top-down manner.  There have also been bottom-up solutions such as PlaneAE \cite{Yu:etal:CVPR2019:PlaneAE} and PlaneTR \cite{Tan:etal:ICCV2021:Planetr} which obtain plane-level masks by a post-clustering procedure on top of learned deep pixel-wise embeddings. Recently, as another emergent fundamental paradigm, Transformers \cite{Vaswani:etal:NIPS2017} have made great progress on a wide range of vision tasks \cite{wang2020axial, dosovitskiy2020image, wang2022towards}. The success of vision Transformers not only comes from its realization of global/long-range interaction across images via attention mechanisms, another important factor is the design of query-based set predictions, initially proposed in Detection Transformer (DETR) \cite{carion2020end} to enable reasoning between detected instances and their global context, which has been further proven to be particularly effective in high-level vision tasks such as instance and semantic segmentation \cite{cheng2022masked, xu2022fashionformer}, video panoptic segmentation \cite{weber2021step, yuan2022polyphonicformer}, \etc.
PlaneTR \cite{Tan:etal:ICCV2021:Planetr} is one early attempt of using such ideas in single-view image plane reconstruction. It is also inspired by structure-guided learning to integrate additional geometric cues like line segments during training, leading to state-of-the-art performance on the ScanNetv1 dataset and the unseen NYUv2-Plane dataset. Based on high-quality monocular plane predictions, following works have explored their extension to establish a plane-aware structure from motion framework, involving across-view plane correspondence learning and camera pose estimation \cite{jin2021sparseplanes,rockwell20228posevit,agarwala2022planeformers,tan2023nopesac}. 

However, all of the above mentioned single/multi-view plane recovery methods somewhat disentangle the prediction of principle components required for plane reconstruction.
\jiajia{For} single-view \jiajia{methods}, PlaneRCNN \cite{Liu:etal:CVPR2019:Planercnn} learns to predict plane offset from a monocular depth prediction branch while other attributes such as plane mask and normal are estimated separately from colour images, PlaneTR \cite{Tan:etal:ICCV2021:Planetr} also predicts a monocular depth map apart from the Transformer module, which is later used for acquiring plane segmentation masks using clustering associate embedding \cite{Yu:etal:CVPR2019:PlaneAE}. 
For multi-view ones, SparsePlanes \cite{jin2021sparseplanes}, PlaneFormers \cite{agarwala2022planeformers} and NOPE-SAC \cite{tan2023nopesac} inherit the above multi-step methods as the monocular plane predictor and further adopt a two-stage framework. Initially, they compute planes of two frames in the same coordinate system (from the predicted initial pose). Subsequently, complex manual design optimization algorithms \cite{jin2021sparseplanes} or additional neural modules \cite{agarwala2022planeformers, tan2023nopesac} are employed to achieve plane correspondence and refine pose hypotheses by incorporating externally provided camera states and explicit correspondence ground truth.
Accordingly, these closely related prediction tasks are usually interleaved, and none of them successfully unifies multi-view plane recovery within a single compact model. We conjecture this could be one performance bottleneck for existing data-driven approaches.

Motivated by this finding, we seek to borrow recent advance in query learning and aim to design a single, compact and unified model to jointly learn all plane-related tasks, and we expect such design would achieve a mutual benefits among tasks and advance the existing performance of both monocular and multi-view plane segmentation and reconstruction. Extensive experiments results on the ScanNet, MatterPort3D and NYUv2-Plane benchmark datasets show that \textit{without using any external priors} \cite{Liu:etal:CVPR2019:Planercnn, Tan:etal:ICCV2021:Planetr, jin2021sparseplanes, agarwala2022planeformers, tan2023nopesac} during training, our unified querying learning model, named PlaneRecTR++, achieves new state-of-the-art performance with a concise structure. Additionally, we have also found that such framework can implicitly discover the spatial plane correspondences to enable precise 3D plane reconstruction.

To summarize, our contributions are as follows:
\begin{itemize}\compresslist
    \item We propose a first single unified framework to address the challenging 3D plane recovery task, where all closely related sub-tasks are jointly optimized and inferred in a multi-task manner motivated by query-based learning.

    \item Without any external priors and supervisions other than input images, we propose a novel plane-aware attention structure to first tackle sparse view plane reconstruction in a purely end-to-end manner, producing robust cross-view plane correspondences and camera poses.
    
    \item  Our proposed method achieves significant gains in terms of model performance and compactness. Extensive numerical and visual comparisons on four public benchmark datasets demonstrate state-of-the-art performance of our proposed unified query learning, taking full advantages of plane-related cues to achieve mutual benefits.
\end{itemize}

A previous version of this work was published at ICCV2023 \cite{shi2023planerectr}. This paper extends the conference version with the following new contributions. First, to amplify the effectiveness of unified query learning in plane reconstruction, we extend PlaneRecTR \cite{shi2023planerectr} to jointly tackle multi-view plane recovery and camera pose estimation in a purely end-to-end manner, \ie PlaneRecTR++, enabling superior performance on the ScanNetv2 and MatterPort3D datasets. Second, we propose a plane-aware cross attention module to implicitly learn plane correspondences, achieving mutual benefits without requiring pose initialization and correspondence labelling. Third, we conduct extensive and comprehensive experiments with detailed ablation analysis to provide a thorough understanding of PlaneRecTR++. 
\jiajia{The plane embeddings guided by query learning from PlaneRecTR++ not only retain comparable single-view plane recovery capability to \cite{shi2023planerectr}, but also exhibit superior cross-view consistency, enabling precise plane matching and pose inference.}

%% file: figures/teaser.tex

\begin{figure}[!th]
\centering
\resizebox{0.85\linewidth}{!}{ 
  \begin{overpic}[width=1.0\linewidth]{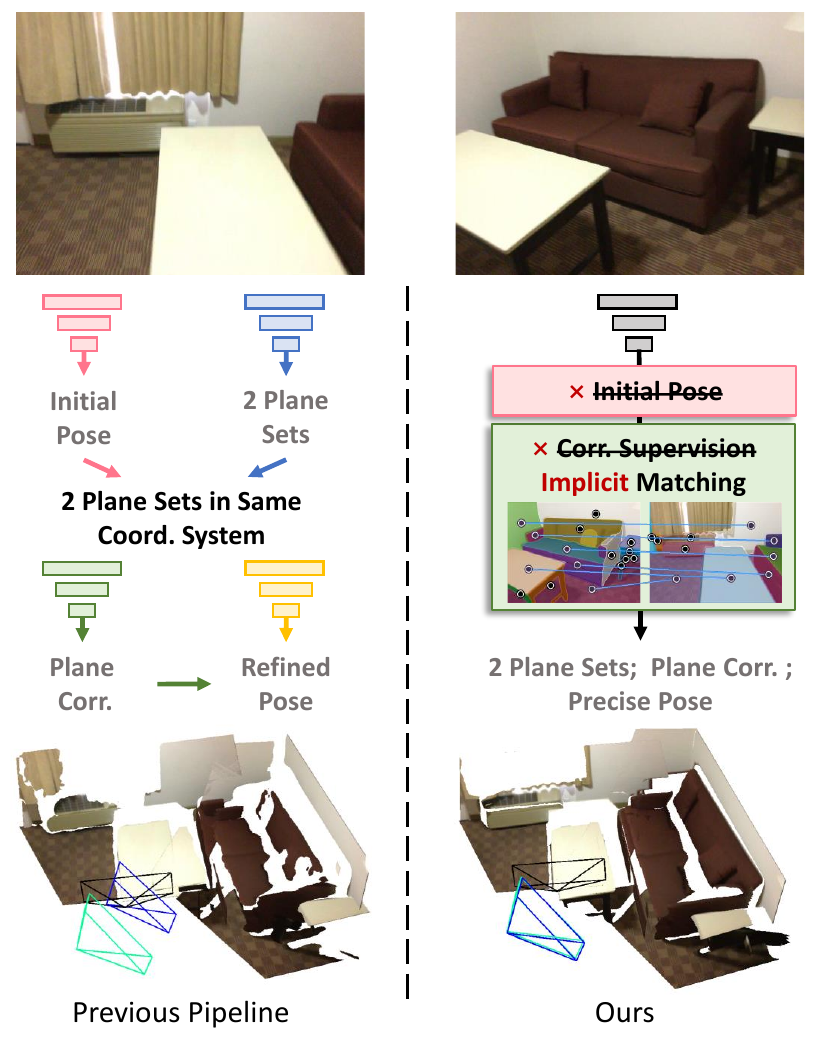}
  \end{overpic} }
  \vspace{-5pt}
  \caption{Previous methods \cite{jin2021sparseplanes, agarwala2022planeformers, tan2023nopesac} with multiple modules V.S. our single stage model. The 3D visualization of previous pipeline is based on the leading NOPE-SAC \cite{tan2023nopesac}. The \textbf{Green} and the \textbf{Blue} frustums show the ground truth and predicted cameras of the first image respectively.The fixed \textbf{Black} frustums show the camera of the second image. }
  \vspace{-10pt}
  \label{fig:teaser}
\end{figure}

%% file: related.tex

\section{Related Work}
\label{sec:related}
\subsection{3D Plane Recovery from a Single Image}
Recovering monocular planes enables 3D planar reconstruction and structural scene understanding. Traditional methods often rely on strong assumptions of scenes \cite{delage2007automatic, fouhey2014unfolding} (\eg, the Manhattan world assumption), or require manual extraction of primitives \cite{delage2007automatic,fouhey2014unfolding, lee2009geometric}, such as superpixels and line segments, which may not be applicable to complex real-world scenes. 

PlaneNet \cite{Liu:etal:CVPR2018:Planenet} is the first to propose an end-to-end learning framework for this task, it also releases a large dataset of planar depth maps utilizing the ScanNetv1 dataset \cite{Dai:etal:CVPR2017}. PlaneRecover \cite{yang2018recovering} presents an unsupervised learning approach that specifically targets outdoor scenes. However, both PlaneNet and PlaneRecover can only predict a fixed number of planes. PlaneRCNN \cite{Liu:etal:CVPR2019:Planercnn} tackles this limitation and extracts an arbitrary number of planes with planar parameters and segmentation masks using a proposal-based instance segmentation framework, \ie, Mask R-CNN \cite{He:etal:ICCV2017}. It also proposes a segmentation refinement network as well as a warping loss between frames to improve performance.
These proposal-based methods require multiple steps to successively tackle sub-tasks of 3D plane recovery including plane detection, segmentation, parameter and depth estimations, \etc. 

On the other hand, PlaneAE \cite{Yu:etal:CVPR2019:PlaneAE} leverages a proposal-free instance segmentation approach, which uses mean shift clustering to group embedding vectors within planar regions. PlaneTR \cite{Tan:etal:ICCV2021:Planetr} inherits the design of DETR \cite{carion2020end} to concurrently detect plane instances and estimate plane parameters, followed by plane segmentations generated by a pixel clustering strategy like PlaneAE \cite{Yu:etal:CVPR2019:PlaneAE}. 
Specifically, its Transformer branch only predicts instance-level plane information, thus post-processing like clustering is still required to carry out pixel-wise segmentation. The global depth is inferred by another convolution branch. 
 
Therefore, existing advanced methods, whether based on direct CNN prediction or embedding clustering, still divide the whole 3D plane recovery task into several steps. In contrast, our plane query learning offers a unified solution for the aforementioned sub-tasks within the intra-frame component and can be seamlessly extended to address inter-frame requirements in an end-to-end manner.

\subsection{3D Planar Reconstruction from Sparse Views}

Built upon advancement in monocular plane recovery, numerous works \cite{jin2021sparseplanes, agarwala2022planeformers, tan2023nopesac} have emerged to address the challenging two-view planar reconstruction with unknown camera poses, aiming to build a coherent 3D planar reconstruction.

SparsePlanes \cite{jin2021sparseplanes} is the first learning-based approach for planar reconstruction and pose estimation from sparse views. 
\jiajia{With monocular planes derived from PlaneRCNN \cite{Liu:etal:CVPR2019:Planercnn} and $1024$ pose hypotheses estimated from a dense pixel attention network, SparsePlanes employs them for a complex two-step optimization to calculate plane correspondences and a final pose.}
Furthermore, PlaneFormers \cite{agarwala2022planeformers} utilizes predicted monocular planes and top \jiajia{$9$} pose hypotheses from SparsePlanes as input, \jiajia{and replaces} the handcrafted optimization by \jiajia{$9$} learnable \jiajia{planeformer} modules, mitigating the \jiajia{intricate} optimization issue \cite{jin2021sparseplanes}.  
NOPE-SAC \cite{tan2023nopesac} improves monocular plane quality by replacing PlaneRCNN \cite{Liu:etal:CVPR2019:Planercnn} with a modified PlaneTR \cite{Tan:etal:ICCV2021:Planetr}, and 
achieves an initial coarse pose from direct regression using the similar pixel attention \cite{jin2021sparseplanes}.
It also introduces differentiable optimal transport \cite{sarlin2020superglue} for plane matching and proposes one-plane pose hypotheses based on corresponding plane pairs, resolving conflicts between numerous pose hypotheses and limited 3D plane correspondences during SparsePlanes' pose refinement.

However, both PlaneFormers and the leading \jiajia{NOPE-SAC} still adopt a multi-stage pipeline derived from SparsePlanes, requiring bootstrapping from external initial pose and correspondence supervision. 
The reason behind this lies in the existence of a chicken-and-egg relationship between explicitly learning camera pose and plane correspondence.  In all previous methods \cite{jin2021sparseplanes, agarwala2022planeformers, tan2023nopesac}, the capability of plane embeddings learned by ground truth correspondence supervision is \jiajia{still} insufficient to accurately track the same plane instance under sparse views.  An additional model is necessary to provide the initial pose(s), so that monocular \jiajia{planes} could be merged under a unified coordinate system, thereby assisting the original plane embedding in enhancing matching accuracy and ultimately refining the initial pose.
Our proposed PlaneRecTR++ deviates from such dilemma using unified plane query learning, actively inferring plane estimations, correspondences, and camera pose in a single-stage framework, \textit{without relying on initial pose guidance and supervision for plane matching}.

\subsection{Correspondence and Camera Pose Estimation}

Camera pose estimation between adjacent images is a fundamental step in multi-view 3D reconstruction\cite{hartley2003multiple}. Early studies focused primarily on extracting sparse keypoint correspondences \cite{Lowe:IJCV2004, bay2008surf} to compute the essential matrix using the five-point solver \cite{nister2004fivepoint}. 
Subsequently, significant efforts have been devoted to learning-based approaches for robust keypoint detection \cite{detone2018superpoint, dusmanu2019d2net, tyszkiewicz2020disk} and matching \cite{sarlin2020superglue, sun2021loftr, zhou2020learn}. However, pose computed solely from the essential matrix lacks a real scale and causes further challenges in the sparse view setting, where only a limited number of correct correspondences could be found.

Moreover, existing methods for directly learning relative pose from images usually require concatenating pairwise frames \cite{en2018rpnet} or computing affinity volume\cite{cai2021extreme, jin2021sparseplanes, tan2023nopesac}, resulting in a significant computational burden. Additionally, these methods either rely on extensively overlapping images\cite{en2018rpnet, cai2021extreme}, or adapt to sparser views but yield limited precision, thus serving solely as an initial pose prior \cite{jin2021sparseplanes, agarwala2022planeformers, tan2023nopesac}.
Recently proposed Pose Vit \cite{rockwell20228posevit}, a Transformer structure with tokenized uniform image patches as well as their quadratic positional bias as input, attempts to learn proximal patch correspondence information using attention mechanism, ultimately allowing a direct regression of rotation and translation with scale in a wide baseline.

Our approach draws inspiration from attention-based modules \cite{sarlin2020superglue, sun2021loftr,rockwell20228posevit}, but further extends the standard cross attention by a plane-aware attentive design. With our learned plane embeddings as the only input for pose prediction, our method implicitly learns genuine plane correspondences and is able to recover precise camera poses within a compact module.

%% file: overview.tex
\input{figures/planerectr++_overview}
\section{PlaneRecTR++ Overview}
\label{sec:overview}
Our PlaneRecTR++ is an end-to-end unified query learning architecture, designed for the challenging task of joint planar reconstruction and relative camera pose estimation. In Figure \ref{fig:planerectr++_overview}, the only input to PlaneRecTR++ are two RGB images $I_1$ and $I_2$ of different views, with their learnable queries serving as the bridge jointly connecting distinct sub-tasks. Specifically, plane queries guide the learning of unified plane embeddings $\mathcal{E}_\text{plane}$ for each possible plane candidate within input frames, whose mutual interactions enable their decoding to final planar attributes, cross-view plane correspondences, and relative camera pose. Please note that the above results are obtained by PlaneRecTR++ in one shot without \jiajia{external pose prior or correspondences supervision} \cite{jin2021sparseplanes, agarwala2022planeformers, tan2023nopesac}. 

To better dissect our method, we partition PlaneRecTR++ into two components based on the learning scope: intra-frame and inter-frame plane query learning. The intra-frame component aims to recover per-view 3D planes. It unifies various sub-tasks of monocular plane recovery including plane detection, parameter prediction, segmentation and depth estimation, while eliminating the need for multi-step prediction in previous methods \cite{jin2021sparseplanes,agarwala2022planeformers,tan2023nopesac}. Such design philosophy brings not only simplicity and compactness of overall framework but also mutual benefits among closely related tasks. Furthermore, in the inter-frame component, we integrate these learned unified plane embeddings from different views to form an attentive planar association matrix to approximate plane correspondences and capture \jiajia{features} of paired planes for directly pose regression. This design implicitly motivates multi-view consistency of plane embeddings and is proven to be sufficient for precise plane tracking, without any direct supervision. 
As a result, PlaneRecTR++ manages to unify separate multi-stage tasks required in the previous sparse views pipelines \cite{jin2021sparseplanes,agarwala2022planeformers,tan2023nopesac}, including monocular plane recovery, pose initialization, plane matching, and pose refinement. This joint optimization of all sub-tasks naturally enhances efficiency and final performance.

The training process of PlaneRecTR++ also comprises two phases: \jiajia{(1) A} monocular pre-training phase, where monocular images are utilized to train the intra-frame component for single-view plane recovery; \jiajia{(2) A} joint training phase, where paired images are used to train the complete PlaneRecTR++, optimizing both components in an end-to-end manner for planar reconstruction and pose estimation on challenging sparse-view datasets. Please note that our method could also be trained and converged well from scratch, \ie, only applying phase (2), however, we \jiajia{find} the two-phase training achieves better overall performance without losing its virtue in end-to-end unified query learning. Therefore we stick to the two-phase training unless otherwise mentioned.

In the subsequent Section \ref{sec:intra_frame_method} and \ref{sec:inter_frame_method}, we elaborate on architectural designs, training objectives and inference process of intra-frame and inter-frame components, respectively.

%% file: figures/planerectr++_overview.tex
\begin{figure*}[t!]
\centering
\includegraphics[width=0.95\linewidth]{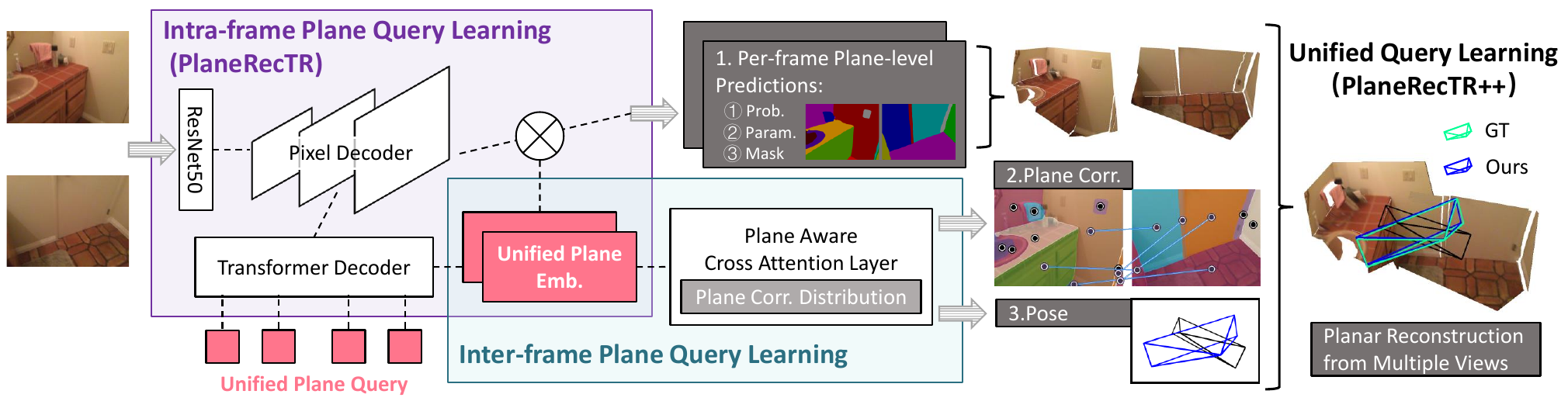}
\vspace{-5pt}
\caption{Overview of \jiajia{the proposed PlaneRecTR++. Our end-to-end model couples intra-frame and inter-frame components through learnable plane queries, unifying numerous interdependent and mutually constraining sub-tasks. Within a single forward pass from input images, PlaneRecTR++ accomplishes joint plane reconstruction and camera pose estimation without initial pose prior.}}
\label{fig:planerectr++_overview}
\vspace{-8pt}
\end{figure*}

%% file: planerectr.tex
\input{figures/planerectr_network}

\section{Intra-Frame Plane Query Learning}
\label{sec:intra_frame_method}

In this section, we present the details of intra-frame plane query learning, \ie, PlaneRecTR. We start by first introducing its architecture in Section~\ref{sec:model}, and then discuss the training process and loss functions in Section \ref{sec:training}. Finally, we describe the inference process of recovering 3D planes from a single view in Section \ref{sec:planeinference}.

\subsection{Transformer-based Unified Query Learning for Single-view Plane Recovery}
\label{sec:model}

Inspired by the successes of DETR \cite{carion2020end} and Mask2Former \cite{cheng2022masked} in object detection and segmentation, we find that, it is \jiajia{feasible} to tackle the challenging monocular planar reconstruction task using a \textit{single}, \textit{compact} and \textit{unified} framework, thanks to the merits of query-based reasoning \jiajia{for enabling} joint modeling of multiple tasks.

As shown in Figure \ref{fig:planerectr_network}, intra-frame plane query learning component consists of three main modules: (1) A pixel-level module to learn dense pixel-wise deep embedding of the input colour image. (2) A Transformer-based unified query learning module to jointly predict, for each of $N$ learnable plane queries, its corresponding plane embeddings $\mathcal{E}_\text{plane}$ as well as four target properties, including plane classification probability $p_i$, plane parameter $n_i$, mask embedding, and depth embedding ($i\in[1,2,..., N]$). Specifically, $p_i$ is the probability to judge whether the $i^\text{th}$ query corresponds to a plane or not; $n_i\doteq \tilde{n_i}/d_i \in \mathbb{R}^3$,  where $\tilde{n_i} \in \mathbb{R}^3$ is its plane normal and $d_i$ is the distance from the $i^\text{th}$ plane to camera center, \ie, offset. (3) A plane-level module to generate plane-level mask $m_i$ and plane-level depth $d_i$ through mask and depth embedding ($i\in[1,2,..., N]$). We then remove non-plane query hypothesis while combining the remaining ones for the final image-wise plane recovery. These three modules will be described in detail below.

\paraspace
\ptitle{Pixel-Level Module.}
Given an \jiajia{input} image of size $H \times W$, we use the pre-trained ResNet-50 \cite{He:etal:CVPR2016} as backbone to extract dense image feature maps, unless otherwise mentioned. \jiajia{Subsequently}, a multi-scale convolutional pixel decoder \jiajia{\cite{cheng2022masked}} is used to produce a set of dense feature maps with four scales, denoted as follows:
\begin{equation}
\begin{aligned}
\setlength{\abovedisplayskip}{3pt}
\setlength{\belowdisplayskip}{3pt}
\mathbb{F}&=\{F_{1}\in \mathbb{R}^{C_{1}\times H/32\times W/32}, F_{2}\in \mathbb{R}^{C_{2}\times H/16\times W/16}, \\
& F_{3}\in \mathbb{R}^{ C_{3}\times H/8\times W/8}, \mathcal{E}_\text{pixel}\in \mathbb{R}^{ C_{\mathcal E}\times H_{\mathcal E}\times W_{\mathcal E}}\},
\end{aligned}
\end{equation}
\jiajia{where $C_1$, $C_2$, $C_3$, $C_{\mathcal E}$ are feature dimensions.} The first three feature maps $\{F_{1}, F_{2}, F_{3}\}$ are fed to the Transformer module, while the last one $\mathcal{E}_\text{pixel}$, a dense per-pixel embedding of resolution $H_{\mathcal E}=H/4$ and $W_{\mathcal E}=W/4$, is exclusively used for computing plane-level binary masks and plane-level depths.

\paraspace
\ptitle{Transformer Module.}
We use the Transformer decoder with masked attention proposed in \cite{cheng2022masked}, which computes \textit{unified plane embeddings} $\mathcal{E}_\text{plane} \in \mathbb{R}^{N \times C_{\mathcal E}}$ from above mentioned multi-scale feature maps $\{F_{1}, F_{2}, F_{3}\}$ and $N$ learnable plane queries. The predicted $\mathcal{E}_\text{plane}$ are then independently projected to four target properties by four MLPs. Overall, the Transformer module predicts required planar attributes through $N$ plane queries. 

\paraspace
\ptitle{Plane-Level Module.}
 As shown in Figure \ref{fig:planerectr_network}, we obtain a dense plane-level binary mask $m_i \in [0, 1]^{H_{\mathcal E} \times W_{\mathcal E}}$/depth prediction $d_i \in \mathbb{R}^{H_{\mathcal E} \times W_{\mathcal E}}$ by a dot product between the $i^\text{th}$ mask/depth embedding and the dense per-pixel embedding $\mathcal{E}_\text{pixel}$ from previous two modules, respectively.
We finally obtain $N$ plane-level predictions $\{y_i=(p_i,n_i, m_i, d_i)\}_{i=1}^N$, \jiajia{each of which} contains all the necessary information to recover a possible 3D plane.

\subsection{Training Objective and Configuration}
\label{sec:training}
\ptitle{Plane-level Depth Training.}
Previous methods tend to predict a global image-wise depth by separate network branches to calculate plane offset \cite{Liu:etal:CVPR2019:Planercnn} or use depth as additional cues to formulate segmentation \cite{Yu:etal:CVPR2019:PlaneAE,Tan:etal:ICCV2021:Planetr}.
In contrast, our method tries to achieve mutual benefits between planar semantic and geometric reasoning, 
\jiajia{and leverages learnable plane queries} to unify all components of plane recovery in a concise multi-task manner. As a result, we explicitly predict dense plane-level depths, binary-masks, plane probabilities and parameters from a shared feature space, which is produced and refined via attention mechanism of the Transformer.

\paraspace
\ptitle{Bipartite Matching.}
During training, one important step is to build optimal correspondences between $N$ predicted planes and $M$ ground truth planes ($N \geq M$).
Following bipartite matching of \cite{cheng2022masked, Tan:etal:ICCV2021:Planetr}, we search for a permutation $\hat{\sigma}$ 
by minimizing a matching defined cost function $D$:
\begin{equation}
\label{eq: bi-matching}
\hat{\sigma}=\underset{\sigma}{\arg \min } \sum_{i=1}^{N} D\left(\hat{y}_{i}, y_{\sigma(i)}\right), 
\end{equation}
\vspace{-0.5cm}
\begin{align}
\label{eq: bi_matching2}
D= & \mathbbm{1}_{\{\hat{p}_{i}=1\}} \Big[
 -\omega_{1} \, p_{\sigma(i)}
+ \omega_{2} L_{1}\left(\hat{{n}}_{i}, {n}_{\sigma(i)}\right) \nonumber\\
& + \omega_{3} L_{1}\left(\hat{{d}}_{i}, {d}_{\sigma(i)}\hat{m}_{i}\right) + \omega_{4} L_{ce}  +  \omega_{5} L_{dice} \Big],
\end{align}
where $\hat{y}_{i} = (\hat{p}_i,\hat{n}_i, \hat{m}_i, \hat{d}_i)$ are the $i^\text{th}$ ground-truth plane attributes, we augment the ground truth instances with non-planes where $\hat{p}_{i}=0$ if $i\textgreater M$; $\sigma(i)$ indicates the matched index of the predicted planes to the ground truth $\hat{y}_{i}$; $\mathbbm{1}$ is an indicator function taking 1 if $\hat{p}_{i}=1$ is true and 0 otherwise; $\omega_{1}, \omega_{2}, \omega_{3}, \omega_{4}$ and $\omega_{5}$ are weighting terms and set to 2, 1, 2, 5, 5, respectively. 
Here we additionally consider the influence of mask and depth quality using a mask binary cross-entropy loss $L_{ce}$ \cite{cheng2022masked}, a mask dice loss $L_{dice}$ \cite{milletari2016v} and an $L_1$ depth loss, respectively.

\paraspace
\ptitle{Loss Functions.}
After bipartite matching, the final training objective $L$ is composed of \jiajia{the} following four parts:
\begin{equation}
\small 
\mathcal{L} = 
\sum\limits_{i=1}^{M}\left( \lambda\mathcal{L}_{\text {cls }}^{(i)} + \mathcal{L}_{\text {param }}^{(i)} + \mathcal{L}_{\text{mask}}^{(i)} + \lambda\mathcal{L}_{\text {depth}}^{(i)} \right),
\end{equation}
where $\lambda$ is a weighting factor and is set to 2 in this paper. $\mathcal{L}_{\text {cls}}$ and $\mathcal{L}_{\text {param}}$ are a plane classification loss and a plane parameter loss, in a similar form to previous work \cite{Tan:etal:ICCV2021:Planetr}.

However, different from PlaneTR \cite{Tan:etal:ICCV2021:Planetr}, the remaining two loss terms in our paper $\mathcal{L}_{\text{mask}}$ and $\mathcal{L}_{\text {depth}}$ are designed to explicitly learn dense planar masks and depths. Specifically,
we introduce the plane segmentation mask prediction loss, as a combination of a cross-entropy loss and a dice loss:
\begin{equation}
\mathcal{L}_{\text {mask }}^{(i)} =
 \mathbbm{1}_{\{\hat{p}_{i}=1\}}   \,  \beta_{1} L_{ce} + \nonumber
 \mathbbm{1}_{\{\hat{p}_{i}=1\}}   \,  \beta_{2} L_{dice},
\end{equation}
where $\beta_{1}, \beta_{2} = 5$.
The depth loss is in a typical $L_1$ form, penalizing the discrepancy of depth value within planar regions:
\begin{equation}
\mathcal{L}_{\text {depth }}^{(i)} =
 \mathbbm{1}_{\{\hat{p}_{i}=1\}}  L_{1}\left(\hat{{d}}_{i}, {d}_{\sigma(i)}\hat{m}_{i}\right),
\end{equation}

\subsection{Inference Process of Monocular 3D Plane Recovery}
\label{sec:planeinference}

For the $N$ plane-level predictions $\{y_i\}_{i=1}^N$ predicted by the network, we first drop non-plane candidates according to the plane classification probability $p_i$, leading to a valid subset of $K$ planes (\ie, $K\leq N$). 
For each pixel within planar regions, we calculate 
the most likely plane index
$\mathop{\arg\max}\limits_{i}\{m_i\}_{i=1}^{K}$ to obtain the final image-wise segmentation mask.
Note that plane-level depths are not involved during inference and we use plane parameters and segmentation to infer planar depths. 
We experimentally found that this design also leads to more structural and smooth geometric predictions than that relying on direct depth predictions.

%% file: figures/planerectr_network.tex
\begin{figure*}[t!]
\centering
\includegraphics[width=0.95\linewidth]{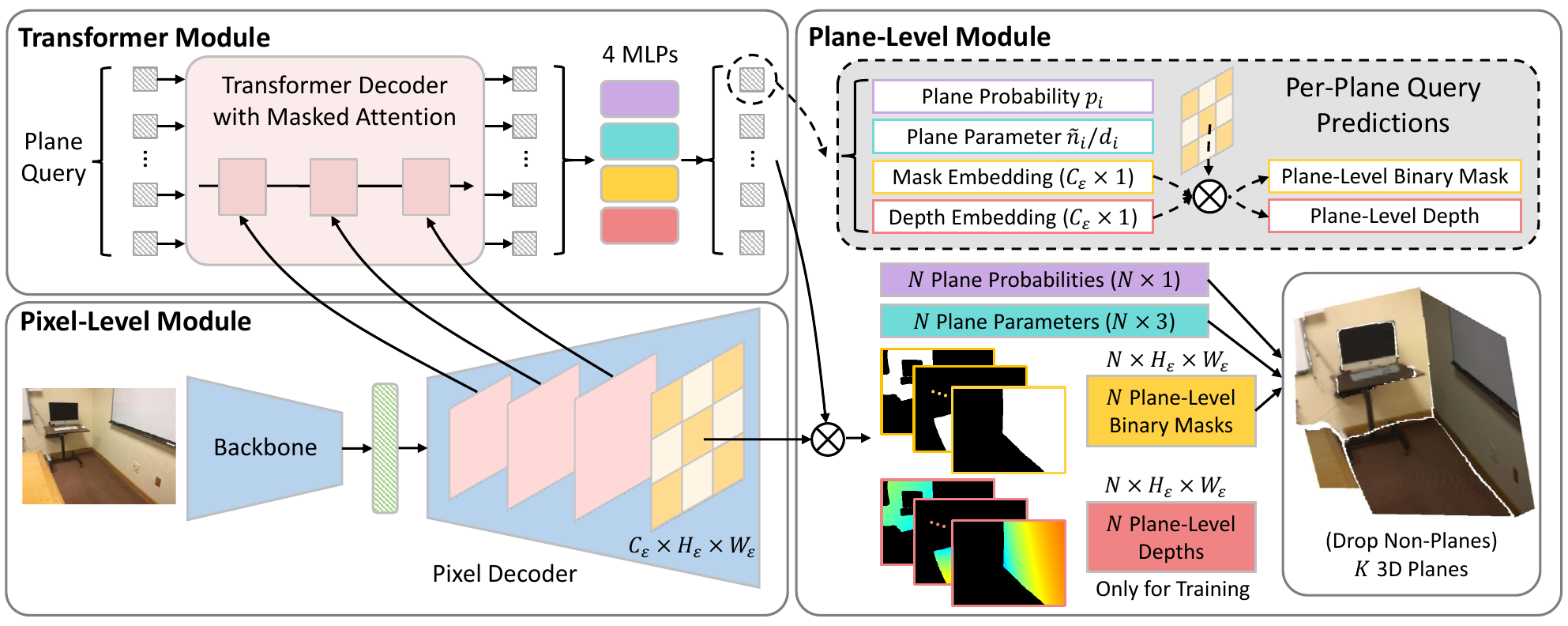}
\vspace{-9pt}
\caption{Overview of intra-frame plane query learning (PlaneRecTR). Our \jiajia{intra-frame component} consists of three main modules:  (1) A pixel-level module to extract dense pixel-wise image features; (2) A Transformer module to jointly predict 4 plane-related properties from each plane query, including plane classification probability, plane parameter, mask and depth embedding;
(3) A plane-level module to calculate dense plane-level binary masks/depths, then filter non-plane predictions and produce the final  3D plane recovery.}
\label{fig:planerectr_network}
\vspace{-8pt}
\end{figure*}

%% file: corrattn.tex
\input{figures/crossplaneattn_network}
\section{Inter-Frame Plane Query Learning}
\label{sec:inter_frame_method}
In this section, we \jiajia{present} how we extend our \jiajia{monocular} framework PlaneRecTR to a novel multi-view setup, while still retaining the virtue of query-based learning. 
We introduce an inter-frame query learning component on top of unified plane embeddings of per-frame. A plane aware cross attention layer is proposed to achieve inter-frame plane interactions, within which dual softmax \cite{sun2021loftr, rockwell20228posevit} and bilinear attention \cite{kim2018bilinear, rockwell20228posevit} mechanisms are used to align the intermediate attention structure with a plane correspondence matrix and enable plane-level feature fusion between views.  
Most importantly, we further modify the key, query, and value forms of standard multi-head attention \cite{Vaswani:etal:NIPS2017} to effectively utilize the complete representation of unified plane embeddings, which better accommodates multi-view plane properties without requiring any additional inputs such as position encoding \cite{sun2021loftr, rockwell20228posevit}. This simple adjustment guarantees that our attention structure truly accomplishes plane matching, allowing the network to spontaneously focus on genuine paired planes from two views.

As illustrated in Figure \ref{fig:crossplaneattn_network}, we employ two plane-aware cross attention layers and an MLP head to construct a simple and lightweight pose regression module. A plane-aware attention layer first takes two-view plane embeddings $\mathcal{E}_\text{plane}^1$ and $\mathcal{E}_\text{plane}^2$ as input, and actively learns their correspondences within the network. Subsequently, our model directly regresses a relative camera pose from probabilistic paired plane embeddings. Conceptually, this entire process aligns with the logical framework of solutions based on classical two-view geometry \cite{hartley2003multiple}, 
thereby offering enhanced interpretability.

\subsection{Cross Attention for Unified Plane Embeddings}
\label{sec:plane_cross_attention}

\ptitle{Preliminaries: Standard Cross Attention.} 
The Transformer cross attention layer \cite{Vaswani:etal:NIPS2017} updates the input value term by mapping query from the $i$-th input and key-value pair from another $j$-th input through a weighted summation, typically using a scaled dot-product similarity function $\operatorname{S}(\cdot,\cdot)$. In addition, attention often employs a multi-head strategy \cite{Vaswani:etal:NIPS2017} where query, key, and value (denoted $Q_i$, $K_j$, $V_j$ 
$\in \mathbb{R}^{N \times C_{\mathcal E}}$, respectively)
are divided, along channel dimensions, into $N_h$ segments $\{q_i^h\}_{h=1}^{N_h}$, $\{k_j^h\}_{h=1}^{N_h}$, $\{v_j^h\}_{h=1}^{N_h} \in \mathbb{R}^{N \times \frac{C_{\mathcal E}}{N_h}}$, in order to enhance the expressiveness and diversity of results without incurring extra computational costs. 

The multi-head similarity function ($\operatorname{S}$) and  cross attention layer ($\operatorname{MCA}$) are defined in Equations \ref{eq:standard_crossattn1} and \ref{eq:standard_crossattn2}:
\begin{small}
\begin{equation}
\label{eq:standard_crossattn1}
\operatorname{S}(q_i^h, k_j^h) =  \operatorname{softmax}\left(\frac{q_i^h{k_j^h}^T}{\sqrt{C_{\mathcal E}/N_h}}, 1\right),
\end{equation}
\end{small}
\begin{small}
\begin{align}
\label{eq:standard_crossattn2}
\operatorname{MCA}(Q_i, K_j, V_j) = \operatorname{Linear}(\operatorname{Concat}(\{\operatorname{S}(q_i^h, k_j^h)v_j^h\}_{h=1}^{N_h})),
\end{align}
\end{small}where $\operatorname{softmax}(\cdot, k)$ applies softmax operation across the $k$-th axis; $\operatorname{Linear}$ and $\operatorname{Concat}$ mean linear projection and channel-wise concatenation, respectively.
Our \jiajia{method} targets equations \ref{eq:standard_crossattn1} and \ref{eq:standard_crossattn2} for intuitive and efficient plane-specific modifications, achieving implicit plane matching and direct pose regression.

\paraspace
\ptitle{Plane Correspondence Probability Function.}
To overcome the limitations of the multi-stage two-view plane reconstruction paradigm \cite{jin2021sparseplanes, agarwala2022planeformers, tan2023nopesac}, here we specifically devise an inter-frame correspondence attention structure to learn reliable plane embeddings, which \jiajia{enables} the network to autonomously acquire probabilities of plane correspondence and conduct pose inference in a single forward pass. This design also eliminates the dependency on either ground truth correspondence supervision or initial pose.

Specifically, in contrast to the similarity function in Equation \ref{eq:standard_crossattn1},
we utilize a dual-softmax operation instead of a single softmax on the \emph{unsplit} query and key embeddings $Q_i, K_j$, aiming to keep integral embedding information when constructing plane-wise correspondence probability, as shown in Equation \ref{eq:dual_softmax} below.
\begin{equation}
\label{eq:dual_softmax}
\small
\operatorname{C}(Q_i, K_j) =\operatorname{softmax}(\frac{Q_iK_j^T}{\sqrt{C_{\mathcal E}}} , 1) \odot \operatorname{softmax}(\frac{Q_iK_j^T}{\sqrt{C_{\mathcal E}}} , 2)\jiajia{,}
\end{equation}
We compute a 2D correspondence matrix $\operatorname{C}(Q_i, K_j)$,  where the element at the $m^{th}$ row and $n^{th}$ column $\operatorname{C}_{mn}(Q_i, K_j)$ denotes the probability that the $m^{th}$ plane embedding from the $i^{th}$ image $I_i$ corresponds to the $n^{th}$ plane embedding from  $j^{th}$ image $I_j$, indicating their likelihood of representing the same plane instance. For the task of two-view plane reconstruction, we stick to configurations of $\{i=1,j=2\}$ and $\{i=2,j=1\}$.

The noteworthy aspect lies in our simple modification to the input query and key's formats, which yield benefits that better align with the characteristics of plane instance. The key $K_j$ and query $Q_i$ in our model are derived from a linear mapping of the unified plane embeddings $\mathcal{E}_\text{plane}$ obtained by intra-frame query learning. We want to highlight the practical significance of an intact $\mathcal{E}_\text{plane}$ representing plane entities. Specifically, we preserve the integrity of $Q_i$ and $K_j$ rather than dividing them into multiple heads to fully leverage the representation power of unified plane embeddings encoding comprehensive information (geometry, appearance, location, context, etc.). This facilitates learning genuine correspondence distribution between planes instead of only abstractly capturing similarities among different subspace representations at various positions.

This simple design has been experimentally validated (Section \ref{sec:sparseview_ablation} and \ref{sec:unified_plane_emb}) to significantly increase the discriminative multi-view consistency of the unified plane embeddings $\mathcal{E}_\text{plane}$, which is of much higher quality than previous methods. Consequently, it facilitates the integration of real plane pairs' information without initial pose for direct precise pose estimation and enables explicit utilization in achieving accurate fusion of plane meshes from two views, thereby completing the overall reconstruction.

\paraspace
\ptitle{Inter-frame Plane Aware Cross Attention.}
\label{sec:bilinear_attntion}
The standard cross attention offers an efficient approach to selectively utilize one of the input plane sequences, but overlooks the interaction between two inputs. Therefore, we adopt bilinear attention \cite{kim2018bilinear, rockwell20228posevit} to incorporate planar information from both views through the plane correspondence probability distribution $\operatorname{C}(Q_i, K_j)$:
\begin{align}
\label{eq:bilinear_attentin}
\operatorname{PCA}(Q_i, K_j, V_i, V_j) = &\operatorname{Linear}(\operatorname{Concat}( \nonumber\\
& \{(v_i^h)^T \operatorname{C}(Q_i, K_j)v_j^h\}_{h=1}^{N_h} ))\jiajia{,}
\end{align}

Unlike the query and key terms, the value is still  divided into $N_h$ segments along the feature dimension, maintaining the advantages of multi-head attention. The value segments share the correspondence attention (Equation \ref{eq:dual_softmax}) from \emph{unsplit} query and key, allowing our model to selectively attend to actual corresponding plane embedding pairs across distinct sub-spaces.

As shown in Figure \ref{fig:crossplaneattn_network}, two parallel plane aware cross attention layers (Equation \ref{eq:bilinear_attentin}) are used to capture integrated features of the corresponding planes from $I_1$ to $I_2$ as well as from $I_2$ to $I_1$. It is worth noting that pose ViT \cite{rockwell20228posevit} employs the identical  features on both sides of bilinear attention matrix. However, we intuitively choose to cross-place embedding sequences of distinct images to model correspondences, thereby enhancing learning efficiency and yielding improved results. We will validate various design disparities through subsequent ablation studies in Section \ref{sec:sparseview_ablation}.

\subsection{Pose Regression}
\label{sec:pose_loss}
Each plane aware cross attention layer ultimately outputs a feature map of size $N_h\times \frac{C_{\mathcal E}}{N_h} \times \frac{C_{\mathcal E}}{N_h}$. The corresponding plane feature maps from two parallel cross attention layers are concatenated and then mapped to a relative camera pose $T=(t, q) \in SE(3)$ using a simple MLP with two hidden layers. Here, $t \in \mathbb{R}^3$ represents translation in real units, and $q \in \mathbb{R}^4$ denotes a unit quaternion representing rotation transformation satisfying $\Vert q \Vert = 1 $.

We use lietorch \cite{teed2021lietorch} to calculate the geodesic distance $\mathcal{G} \in \mathbb{R}^6$ between the predicted pose $T$ and the ground truth pose $T^{\ast}$ as the loss for backpropagation:
\begin{equation}
\mathcal{G} (T, T^{\ast}) = 
\operatorname{Log} ( T.\operatorname{inv()} \cdot  T^{\ast}),
\end{equation}
\begin{equation}
\mathcal{L}_{pose} = \lambda_t \Vert \mathcal{G}_{1:3} \Vert + \lambda_q \Vert \mathcal{G}_{4:6} \Vert\jiajia{.}
\end{equation}
where $\lambda_t = 5$, $\lambda_q = 15$.
It should be noted that $\mathcal{L}_{pose}$ serves as the sole objective for our inter-frame component. Without requiring correspondence supervision, the proposed framework allows for the discovery of plane correspondence, and transforms abstract similarity attention distribution within network into a concrete probabilistic distribution of plane correspondences. Furthermore, the bilinear structure effectively utilizes integrated features of inter-frame planes, leading to accurate pose recovery without relying on external initial poses.

\subsection{Inferring Sparse Views Planar Reconstruction}


After intra-frame plane query learning, we can independently recover the 3D plane sets of two images in their respective camera coordinate systems using the corresponding unified plane embeddings as described in Section \ref{sec:model} and \ref{sec:planeinference}.

During inference, we extract the learned correspondence matrix $\operatorname{C}(Q_i, K_j)$ from the network and filter out low-probability correspondences using a threshold $\theta$. We then employ the mutual nearest neighbor (MNN) criterion \cite{sun2021loftr} to obtain a hard assignment between the two plane sets. Like previous methods for sparse view planar reconstruction \cite{jin2021sparseplanes,agarwala2022planeformers,tan2023nopesac}, based on above estimated camera pose and plane matching results, we transform the plane attributes into canonical viewpoint for final reconstruction and evaluation.  Specifically, we merge normals, offsets, and textures of paired monocular 3D planes \cite{jin2021sparseplanes} to achieve a geometrically precise and smooth reconstruction using sparse views. Those paired planes whose deviations in merged normals or offsets exceed predefined thresholds are removed during inference. 

%% file: figures/crossplaneattn_network.tex
\begin{figure*}[t!]
\centering
\includegraphics[width=0.95\linewidth]{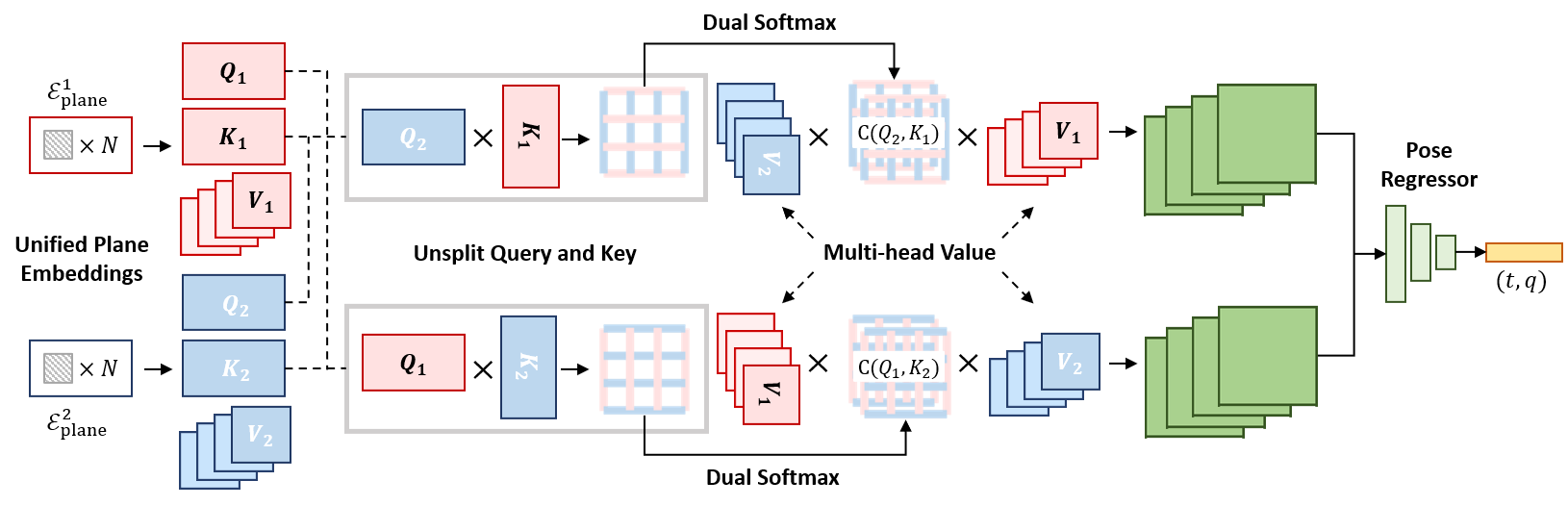}
\vspace{-10pt}
\caption{Overview of inter-frame plane query learning. \jiajia{Our inter-frame component encompasses two parts: (1) Two parallel plane-aware cross attention layers to predict potential cross-view plane correspondences without any association labels; (2) A simple MLP-based pose regressor to directly estimate the relative camera pose.}}
\label{fig:crossplaneattn_network}
\vspace{-10pt}
\end{figure*}

%% file: experiments.tex
\section{Single View Experiments}
\label{sec:experiments}

In this section, we first perform experimental evaluations on the monocular 3D plane recovery task with two public datasets. Then, we ablate the structural designs of our intra-frame component \jiajia{(PlaneRecTR)} to demonstrate that multiple sub-tasks of plane recovery can benefit each other through unified query learning.

\paraspace
\ptitle{Datasets.} 
We train and evaluate \jiajia{the monocular} component on two popular benchmarks: \jiajia{(1)} \textbf{ScanNetv1} \cite{Dai:etal:CVPR2017} dataset, which is a large-scale RGB-D video collection of 1,513 indoor scenes. We use piece-wise planar ground-truth  generated by PlaneNet \cite{Liu:etal:CVPR2018:Planenet}, which contains 50,000 training and 760 testing images with resolution 256 $\times$ 192\jiajia{; (2)} \textbf{NYUv2-Plane} \cite{Silberman:etal:ECCV2012} dataset, which is a planar variant of the original NYUv2 dataset \cite{Silberman:etal:ECCV2012} provided by \cite{Liu:etal:CVPR2018:Planenet}. 
It has 654 testing images with resolution 256 $\times$ 192. 

\paraspace
\ptitle{Evaluation Metrics.}
For the entire task of 3D plane recovery, we adopt per-plane and per-pixel recalls \jiajia{for evaluation} following \cite{Tan:etal:ICCV2021:Planetr, Yu:etal:CVPR2019:PlaneAE}. The per-plane/pixel recall metric is defined as the percentage of the correctly predicted ground truth planes/pixels. 
A plane is considered to be correctly predicted if its segmentation Intersection over Union (IoU), depth, normal and plane offset errors satisfy pre-defined thresholds. Specifically, the \jiajia{IoU} threshold is set to 0.5, while the error thresholds of depth/surface normal vary from 0.05m/2.5° to 0.6m/30°, with an increment of 0.05m/2.5°.

To evaluate plane segmentation, we apply three popular metrics\cite{Arbelaez:etal:TPAMI2010, yang2018recovering}: variation of information (VI), rand index (RI) and segmentation covering (SC). For plane parameter, we \jiajia{find} the best match by minimizing the $L_1$ cost between $K$ predicted planes and $M$ ground truth planes, and then separately compute the average errors of plane normal and offset. As to the NYUv2-Plane dataset, depth accuracy is evaluated on structured planar regions.

\input{tables/planerectr_variants}

\input{figures/planerectr_scannet_3dres}

\paraspace
\ptitle{Network Variants.}
To reflect the contributions of our key design choices, we consider (1) explicit plane-level binary mask prediction (M), (2) plane-level depth prediction (D), (3) plane parameter estimation (P), and thus denote network varieties in Table \ref{tab:planerectr_variants}.

\subsection{Implementation Detail}
\label{sec:monocular_imp_detail}
Our intra-frame component is implemented using Detectron2 \cite{wu2019detectron2}. We train it on the ScanNetv1 training set with a total of 34 epochs on a single NVIDIA TITAN V GPU. We use AdamW optimizer \cite{loshchilov2017decoupled} with an initial learning rate of $0.0001$ and a weight decay of $0.05$. The batch size is set to $16$.

\subsection{Results on the ScanNetv1 Dataset}
\ptitle{Qualitative Results.}
\jiajia{In Figure \ref{fig:planerectr_scannet_3dres}}, we present qualitative results of our single-view plane reconstruction on a variety of unseen ScanNetv1 scenes in both 2D and 3D domains. \jiajia{Our method} is able to predict accurate plane segmentation masks and plane parameters, as well as reasonable 3D plane reconstructions. We further show detailed visual comparisons to \jiajia{the leading} PlaneTR \cite{Tan:etal:ICCV2021:Planetr} in Figure \ref{fig:planerectr_compare_3dres}. Although PlaneTR integrates extra structural cues like line segments, it is exciting to see that our approach, mainly benefiting from joint modelling of 
plane geometry and segmentation, 
is able to predict crisper segmentation and discriminate planes sharing similar normals 
(columns (b), (c))
, with more complete and holistic structures.

\input{tables/planerectr_scannet_nyu_seg}

\input{tables/planerectr_nyu_depth}
\input{figures/planerectr_compare_3dres}

\input{figures/planerectr_scannet_recall}
\paraspace
\ptitle{Quantitative Results.}
We conduct extensive quantitative evaluations towards previous state-of-the-art learning-based plane recovery methods: PlaneNet \cite{Liu:etal:CVPR2018:Planenet}, PlaneRCNN \cite{Liu:etal:CVPR2019:Planercnn}, PlaneAE \cite{Yu:etal:CVPR2019:PlaneAE} and PlaneTR \cite{Tan:etal:ICCV2021:Planetr}. Like \cite{Tan:etal:ICCV2021:Planetr}, PlaneRCNN is shown here mainly as a reference because of its learning with a different ScanNetv1 training set-up. 
We use public implementation of PlaneTR \cite{Tan:etal:ICCV2021:Planetr} and its provided pre-trained weights to report corresponding performance. 

In terms of segmentation accuracy, Table \ref{tab:planerectr_scannet_nyu_seg} shows that on the challenging ScanNetv1 we achieve a new state-of-the-art plane segmentation performance, outperforming \jiajia{the} leading PlaneTR with a relative large margin especially in the VI metric. To further demonstrate the flexibility of our method, we have shown improved versions of PlaneRecTR by replacing ResNet-50 backbone \cite{He:etal:CVPR2016} with the same HRNet-32 backbone \cite{WangSCJDZLMTWLX19:HRNet} as PlaneTR or a SwinTransformer-B model \cite{liu2021swin}. The performance gap widens and shows that our framework could benefit from ongoing research in developing more powerful fundamental vision models.

\input{figures/planerectr_backbones}

As to 
performance of entire plane recovery task, 
we display per-pixel and per-plane recalls of depth and plane normal on the ScanNetv1 dataset, respectively (see Figure \ref{fig:planerectr_scannet_recall}).
With varying thresholds from 0.05 to 0.6 meters and from 2.5 to 30 degrees for depth and normal evaluations, our method consistently outperforms all baselines, indicating more precise predictions of plane parameters, segmentation mask and planar depths. We want to further highlight the significant improvement w.r.t. per-plane recall, our methods could efficiently discover structural planes of various scales, in contrast to \cite{Tan:etal:ICCV2021:Planetr} which tends to miss planes with small areas or sharing similar geometry (see Figure \ref{fig:planerectr_compare_3dres}).

\subsection{Results on the NYUv2-Plane Dataset}

The NYUv2-Plane dataset is chosen here mainly to verify the generalization capability of our method on unseen novel scenes. As shown in Table \ref{tab:planerectr_scannet_nyu_seg}, our method still achieves leading plane segmentation accuracy in all metrics without any fine-tuning. Please check the bottom part of Figure \ref{fig:planerectr_compare_3dres} for more detailed visual comparisons.
In Table \ref{tab:planerectr_nyu_depth},
we focus on pixel-wise depth accuracy in planar regions, thus we treat PlaneTR \cite{Tan:etal:ICCV2021:Planetr} as the targeting baseline for a fair comparison and others as reference.
Our base pipeline performs on-par with PlaneTR and outperforms PlaneNet \cite{Liu:etal:CVPR2018:Planenet} and PlanAE \cite{Yu:etal:CVPR2019:PlaneAE}. When our backbone is replaced from ResNet-50 \cite{He:etal:CVPR2016} to the same HRNet-32 \cite{WangSCJDZLMTWLX19:HRNet} as PlaneTR, our performance is significantly better than PlaneTR. PlaneRCNN \cite{Liu:etal:CVPR2019:Planercnn}, with a ResNet-101 backbone, achieves better performance, partly caused by its utilization of neighbouring multi-view information and higher resolution images during training, while ours is trained with a single image without using extra cues \cite{Liu:etal:CVPR2019:Planercnn,Tan:etal:ICCV2021:Planetr}. 
Nevertheless, we have found that the benefit using more discriminative backbones (\eg, HRNet-32, Swin-B) also transfers to our generalization ability in novel scenes, resulting in a huge performance boost to reach a leading planar depth accuracy. 

\input{tables/planerectr_all_ablation}

\input{figures/ablation_planerectr_mask_compare}

\subsection{Ablation Studies}
\label{sec:ablation}
\jiajia{We own the above improvements to the explicit joint modelling of plane geometry and segmentation via our unified query learning.} In this section, we conduct detailed ablation studies on the ScanNetv1 dataset and show the effectiveness of the key designs.

\paraspace
\ptitle{Effects of Jointly Predicting Plane-Level Depths.}
One key difference to previous Transformer-based \cite{Tan:etal:ICCV2021:Planetr} and CNN-based baselines \cite{Liu:etal:CVPR2018:Planenet,Yu:etal:CVPR2019:PlaneAE,Liu:etal:CVPR2019:Planercnn} is that, instead of learning monocular (planar) depth prediction in an individual branch, we predict plane-level depths based on shared representation for plane detection, segmentation and parameter prediction. It turns out that this simple design choice leads to promising performance increment, achieve mutual benefits among tasks. Bottom two rows of Table \ref{tab:planerectr_all_ablation} show that augmenting depth prediction task to query learning affects per-pixel and per-plane recalls of depth and normal, plane parameter errors. It is worth noting the significant gain in recall rates especially under the strict (lower) thresholds.  

\paraspace
\ptitle{Effects of Predicting Dense Plane-Level Masks.}
Compared to \jiajia{the} Transformer-based PlaneTR, we  differ in obtaining plane-level masks via a dense prediction instead of post-clustering from dense embeddings and depth information. Rightmost two columns of Table \ref{tab:planerectr_all_ablation} have shown that explicitly learning per-plane masks significantly boosts the accuracy of plane parameter estimation of our method, even when direct depth supervision is not available.

\paraspace
\ptitle{Plane Segmentation and Geometry Prediction.}
We also investigate whether the powerful segmentation framework could further benefit from learning plane geometry (plane parameters and depths). The experiment results show that there are marginal numerical gains \jiajia{for} plane segmentation \jiajia{when} comparing \jiajia{PlaneRecTR} to PlaneRecTR (-P-D). Fortunately, we do observe clear qualitative improvements. As shown in Figure \ref{fig:ablation_planerectr_mask_compare}, our complete model manages to distinguish detailed planar structures and even predict more fine-grained and sound plane segmentation than ground truth annotation, which, however, could possibly be penalized during quantitative evaluation (bottom row of Figure \ref{fig:ablation_planerectr_mask_compare} where bedstead labelling is missing).

\paraspace
\ptitle{Backbone Impacts.}
Like other vision frameworks, the backbone feature extractor also has an active role in the final performance of PlaneRecTR. We see this as an exciting advantage, since our concise architectural design continues to benefit from ongoing research in fundamental vision models. PlaneRecTR exhibits a stable performance increase and achieves superior SOTA performance when adopting more powerful backbones such as HRNet32 and SwinB, as shown in Table \ref{tab:planerectr_scannet_nyu_seg} and \ref{tab:planerectr_nyu_depth} \jiajia{as well as Figure \ref{fig:planerectr_scannet_recall} and \ref{fig:planeretr_backbones}}.

%% file: tables/planerectr_variants.tex
\begin{table}
\centering
\caption{Our PlaneRecTR variants during experiments. (`-X' indicates the network is trained without task `X')}
\vspace{-5pt}
\resizebox{0.95\linewidth}{!}{ 
    \begin{tabular}{c|cccc}
    \toprule
        Name & Backbone &  Task 'M' & Task 'P' & Task 'D' \\ 
        \hline
        PlaneRecTR          & ResNet-50 & $\checkmark$ & $\checkmark$ & $\checkmark$\\
        PlaneRecTR (-D)  & ResNet-50 & $\checkmark$ & $\checkmark$ & -\\
        PlaneRecTR (-M-D)  & ResNet-50 & - & $\checkmark$ & - \\
        PlaneRecTR (-P-D)  & ResNet-50 & \checkmark & - & - \\
        PlaneRecTR (HRNet-32)  & HRNet-32 \cite{WangSCJDZLMTWLX19:HRNet} & $\checkmark$ & $\checkmark$ & $\checkmark$\\
        PlaneRecTR (Swin-B)  & Swin-B \cite{liu2021swin} & $\checkmark$ & $\checkmark$ & $\checkmark$\\
        \bottomrule
    \end{tabular}
}
\vspace{-5pt}
\label{tab:planerectr_variants}
\end{table}

%% file: figures/planerectr_scannet_3dres.tex
\begin{figure*}
\centering
\renewcommand\tabcolsep{15pt}
\begin{tabular}{rccccccc}
\raisebox{20pt}{\rotatebox[origin=c]{90}{Input}}
\includegraphics[width=0.12\linewidth]{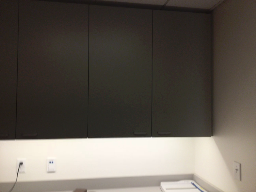}
\includegraphics[width=0.12\linewidth]{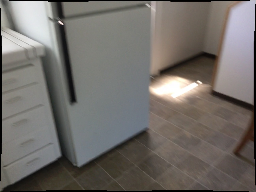}
\includegraphics[width=0.12\linewidth]{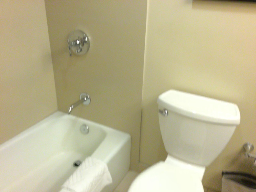}
\includegraphics[width=0.12\linewidth]{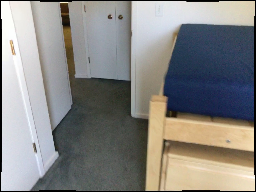}
\includegraphics[width=0.12\linewidth]{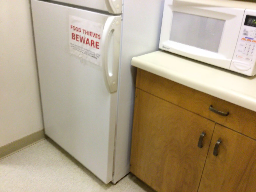}
\includegraphics[width=0.12\linewidth]{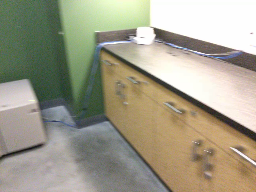}
\includegraphics[width=0.12\linewidth]{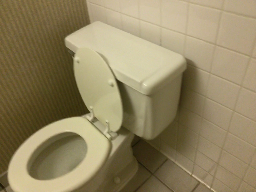}\\
\raisebox{20pt}{\rotatebox[origin=c]{90}{Mask}}
\includegraphics[width=0.12\linewidth]{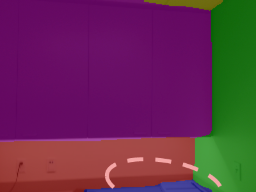}
\includegraphics[width=0.12\linewidth]{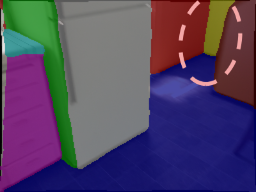}
\includegraphics[width=0.12\linewidth]{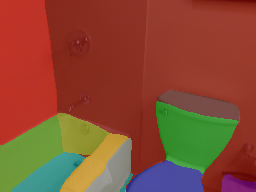}
\includegraphics[width=0.12\linewidth]{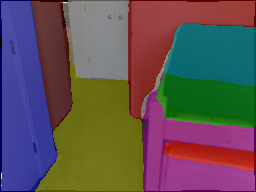}
\includegraphics[width=0.12\linewidth]{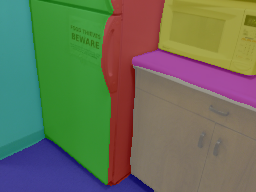}
\includegraphics[width=0.12\linewidth]{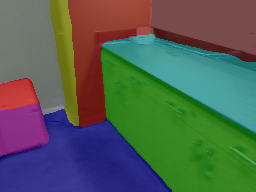}
\includegraphics[width=0.12\linewidth]{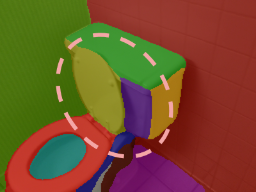}\\
\raisebox{20pt}{\rotatebox[origin=c]{90}{GT Mask}}
\includegraphics[width=0.12\linewidth]{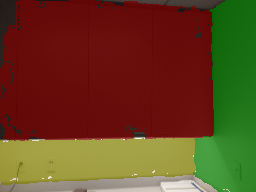}
\includegraphics[width=0.12\linewidth]{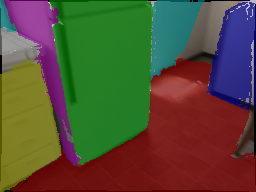}
\includegraphics[width=0.12\linewidth]{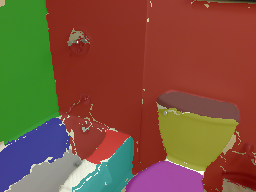}
\includegraphics[width=0.12\linewidth]{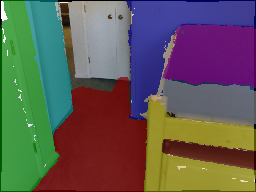}
\includegraphics[width=0.12\linewidth]{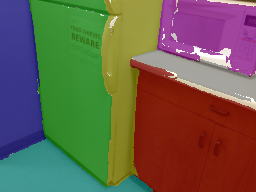}
\includegraphics[width=0.12\linewidth]{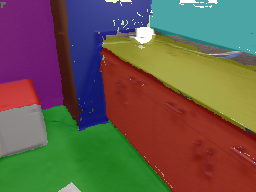}
\includegraphics[width=0.12\linewidth]{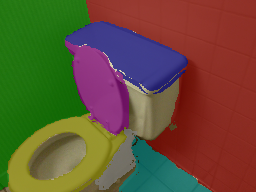}\\
\raisebox{20pt}{\rotatebox[origin=c]{90}{Depth}}
\includegraphics[width=0.12\linewidth]{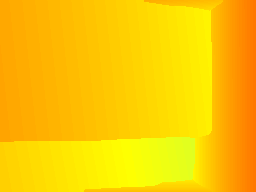}
\includegraphics[width=0.12\linewidth]{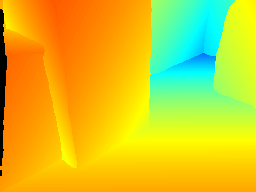}
\includegraphics[width=0.12\linewidth]{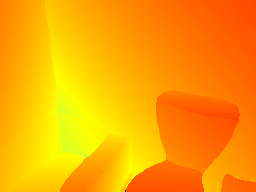}
\includegraphics[width=0.12\linewidth]{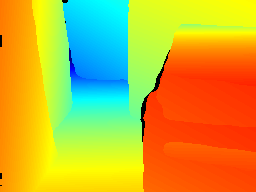}
\includegraphics[width=0.12\linewidth]{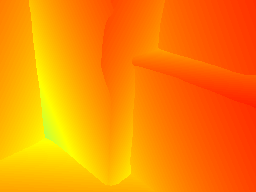}
\includegraphics[width=0.12\linewidth]{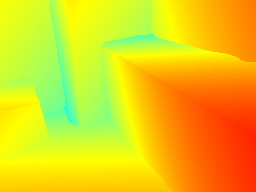}
\includegraphics[width=0.12\linewidth]{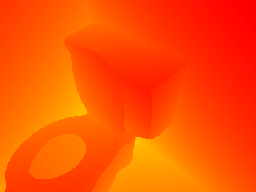}\\
\raisebox{20pt}{\rotatebox[origin=c]{90}{GT Depth}}
\includegraphics[width=0.12\linewidth]{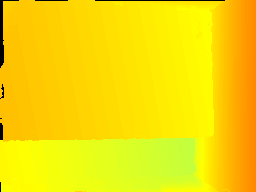}
\includegraphics[width=0.12\linewidth]{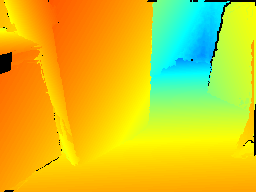}
\includegraphics[width=0.12\linewidth]{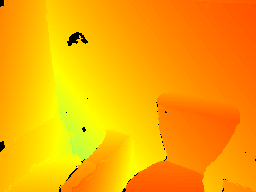}
\includegraphics[width=0.12\linewidth]{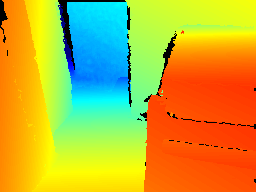}
\includegraphics[width=0.12\linewidth]{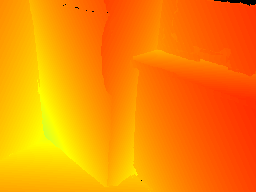}
\includegraphics[width=0.12\linewidth]{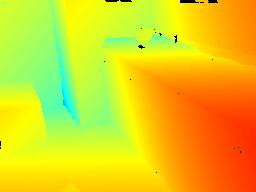}
\includegraphics[width=0.12\linewidth]{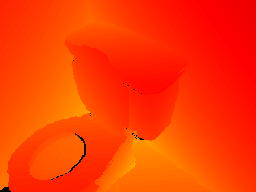}\\
\raisebox{20pt}{\rotatebox[origin=c]{90}{3D Models}}
\includegraphics[width=0.12\linewidth]{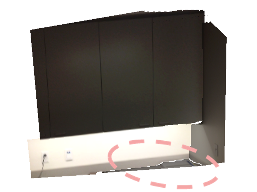}
\includegraphics[width=0.12\linewidth]{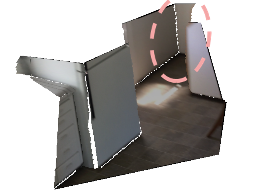}
\includegraphics[width=0.12\linewidth]{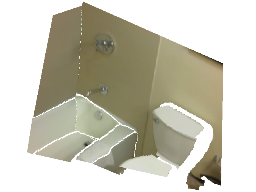}
\includegraphics[width=0.12\linewidth]{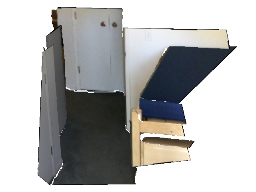}
\includegraphics[width=0.12\linewidth]{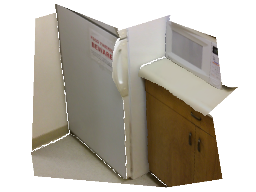}
\includegraphics[width=0.12\linewidth]{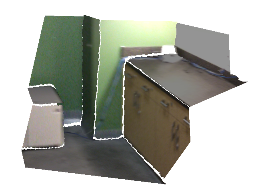}
\includegraphics[width=0.12\linewidth]{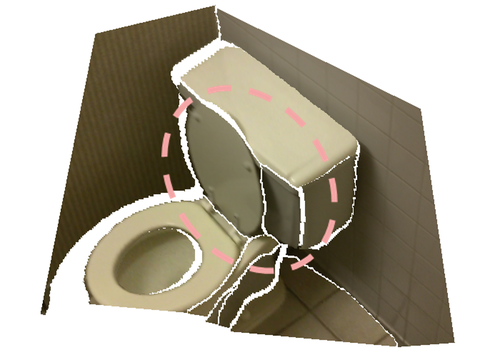}\\
\raisebox{20pt}{\rotatebox[origin=c]{90}{GT 3D Models}}
\includegraphics[width=0.12\linewidth]{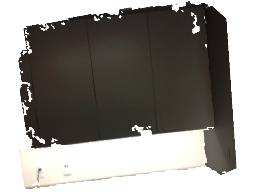}
\includegraphics[width=0.12\linewidth]{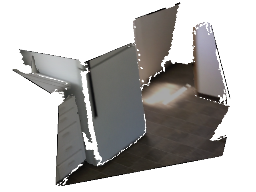}
\includegraphics[width=0.12\linewidth]{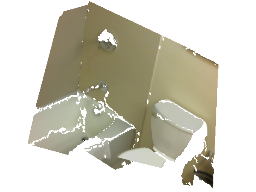}
\includegraphics[width=0.12\linewidth]{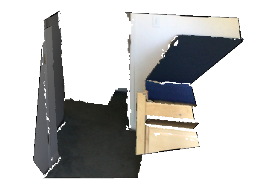}
\includegraphics[width=0.12\linewidth]{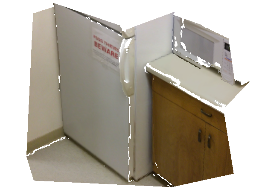}
\includegraphics[width=0.12\linewidth]{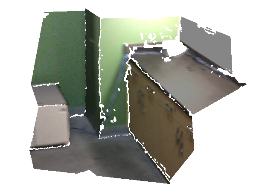}
\includegraphics[width=0.12\linewidth]{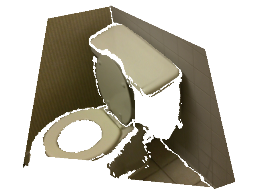}
\end{tabular}  
\vspace{2mm}
\caption{3D plane recovery results of PlaneRecTR on the ScanNetv1 dataset.}
\vspace{-2mm}
\label{fig:planerectr_scannet_3dres}
\end{figure*}

%% file: tables/planerectr_scannet_nyu_seg.tex
\begin{table}
\centering
\caption{Comparison of plane segmentation results. }
\vspace{-5pt}
\resizebox{0.95\linewidth}{!}{ 
    \begin{tabular}{c|ccc|ccc}
    \toprule
        \multirow{2}{*}{Method} & \multicolumn{3}{c|}{ScanNet} & \multicolumn{3}{c}{NYUv2-Plane} \\ 
                                & VI~$\downarrow$ & RI~$\uparrow$ & SC~$\uparrow$ & VI~$\downarrow$ & RI~$\uparrow$ & SC~$\uparrow$\\\midrule
        PlaneNet \cite{Liu:etal:CVPR2018:Planenet}        & 1.259 & 0.858 & 0.716 & 1.813 & 0.753 & 0.558 \\
        PlaneRCNN \cite{Liu:etal:CVPR2019:Planercnn}      & 1.337 & 0.845 & 0.690 & 1.596 & 0.839 & 0.612 \\
        PlaneAE \cite{Yu:etal:CVPR2019:PlaneAE}        & 1.025 & 0.907 & 0.791 & 1.393 & 0.887 & 0.681 \\
        PlaneTR \cite{Tan:etal:ICCV2021:Planetr}           & 0.767 & 0.925 & 0.838 & 1.163 & 0.891 & 0.712  \\ \bottomrule
        PlaneRecTR            & \textbf{0.698} & \textbf{0.936} & \textbf{0.854} & \textbf{1.130} & \textbf{0.905} & \textbf{0.722} \\
        PlaneRecTR (HRNet-32)
    & {\textbf{0.679}} & 
    {\textbf{0.937}} & {\textbf{0.857}}&  {\textbf{1.049}} & {\textbf{0.912}} & {\textbf{0.738}} \\
        PlaneRecTR (Swin-B)            & \underline{\textbf{0.651}} & \underline{\textbf{0.943}} & \underline{\textbf{0.866}} & \underline{\textbf{1.045}} & \underline{\textbf{0.915}} & \underline{\textbf{0.745}}  \\
         \bottomrule
    \end{tabular}
}
    \vspace{-5pt}
\label{tab:planerectr_scannet_nyu_seg}
\end{table}

%% file: tables/planerectr_nyu_depth.tex
\begin{table}
\centering
\caption{Depth accuracy comparison on the NYUv2 dataset. }
\vspace{-5pt}
\resizebox{0.95\linewidth}{!}{ 
    \begin{tabular}{c|ccc|ccc}
    \toprule
        Method & Rel$\downarrow$ &  ${\text{log}_{10}}\downarrow$ & RMSE$\downarrow$ & $\delta_{1}\uparrow$ & $\delta_{2}\uparrow$ & $\delta_{3}\uparrow$ \\ 
        \hline
        {PlaneNet \cite{Liu:etal:CVPR2018:Planenet}}        & {0.236} & {0.124} & {0.913} & {53.0} & {78.3} & {90.4} \\
        {PlaneAE \cite{Yu:etal:CVPR2019:PlaneAE}}      & {0.205} & {0.097} & {0.820} & {61.3} & {87.2} & {95.8}\\ 
        {PlaneRCNN \cite{Liu:etal:CVPR2019:Planercnn}}        & {\textbf{0.183}} & {\textbf{0.076}} & {\textbf{0.619}} & {\textbf{71.8}} & {\textbf{93.1}} & {\textbf{98.3}} \\

        {PlaneTR \cite{Tan:etal:ICCV2021:Planetr}}      & {0.199} & {0.100} & {0.700} & {59.6} & {86.6} & {96.3} \\ \hline

        PlaneRecTR         & 0.202 & 0.100  & 0.729 & 59.5 & 87.4 & 96.0 \\    
        {PlaneRecTR (HRNet-32)}               
        & {\textbf{0.178}} & {\textbf{0.087}} & {\textbf{0.635}} & {\textbf{66.6}} & {\textbf{91.1}} & {\textbf{ 97.7}} \\ 
        PlaneRecTR (Swin-B)           & \underline{\textbf{0.157}} & \underline{\textbf{0.073}} & \underline{\textbf{0.547}} & \underline{\textbf{74.2}} & \underline{\textbf{94.2}} & \underline{\textbf{99.0}}  \\
         \bottomrule
    \end{tabular}
}
\vspace{-10pt}
\label{tab:planerectr_nyu_depth}
\end{table}

%% file: figures/planerectr_compare_3dres.tex
\begin{figure*}
\centering
\subfigure[{\scriptsize Input}]{
\begin{minipage}[b]{0.115\linewidth}
\includegraphics[width=1\linewidth]{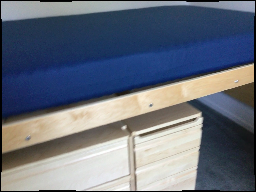}
\includegraphics[width=1\linewidth]{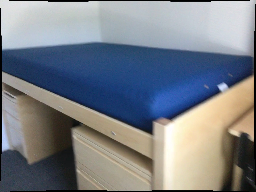}
\includegraphics[width=1\linewidth]{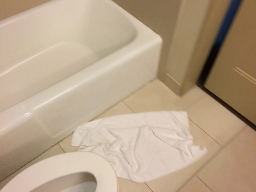}
\includegraphics[width=1\linewidth]{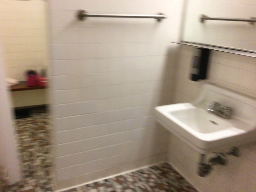}
\includegraphics[width=1\linewidth]{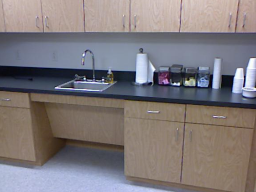}
\includegraphics[width=1\linewidth]{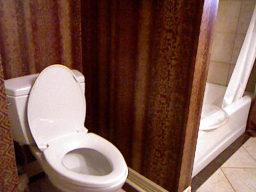}
\includegraphics[width=1\linewidth]{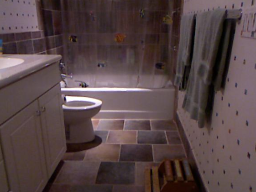}
\end{minipage}}
\subfigure[{\scriptsize PlaneTR Mask}]{
\begin{minipage}[b]{0.115\linewidth}
\includegraphics[width=1\linewidth]{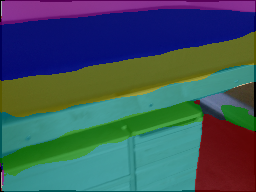}
\includegraphics[width=1\linewidth]{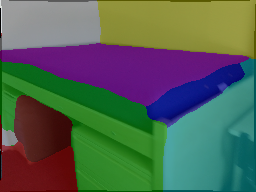}
\includegraphics[width=1\linewidth]{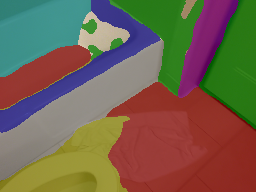}
\includegraphics[width=1\linewidth]{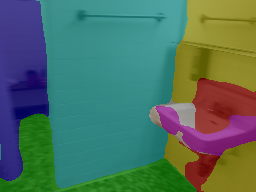}
\includegraphics[width=1\linewidth]{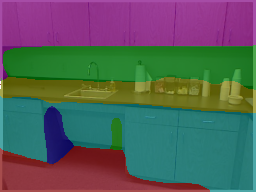}
\includegraphics[width=1\linewidth]{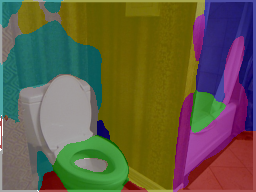}
\includegraphics[width=1\linewidth]{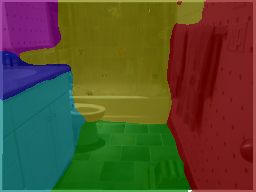}
\end{minipage}
}
\subfigure[{\scriptsize Ours Mask}]{
\begin{minipage}[b]{0.115\linewidth}
\includegraphics[width=1\linewidth]{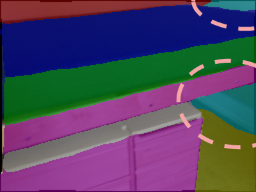}
\includegraphics[width=1\linewidth]{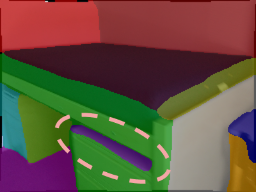}
\includegraphics[width=1\linewidth]{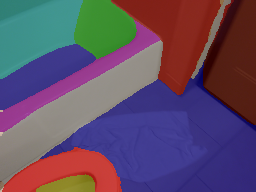}
\includegraphics[width=1\linewidth]{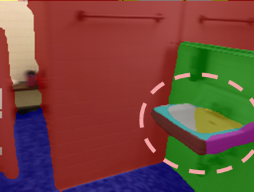}
\includegraphics[width=1\linewidth]{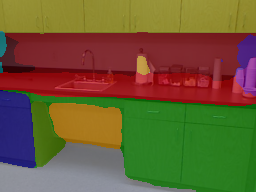}
\includegraphics[width=1\linewidth]{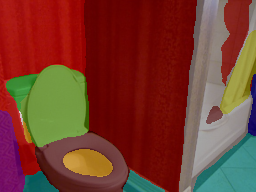}
\includegraphics[width=1\linewidth]{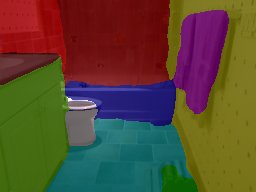}
\end{minipage}
}
\subfigure[{\scriptsize GT Mask}]{
\begin{minipage}[b]{0.115\linewidth}
\includegraphics[width=1\linewidth]{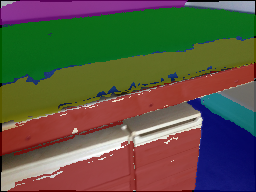}
\includegraphics[width=1\linewidth]{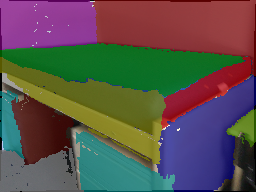}
\includegraphics[width=1\linewidth]{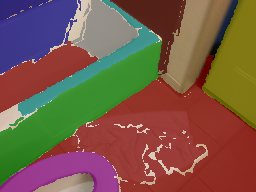}
\includegraphics[width=1\linewidth]{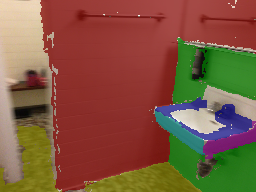}
\includegraphics[width=1\linewidth]{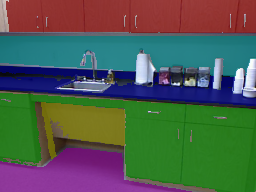}
\includegraphics[width=1\linewidth]{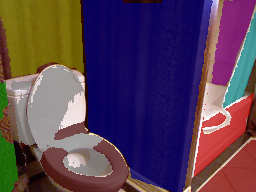}
\includegraphics[width=1\linewidth]{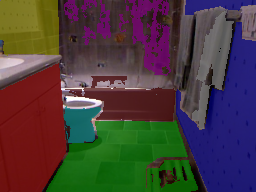}
\end{minipage}
}
\subfigure[{\scriptsize PlaneTR 3D}]{
\begin{minipage}[b]{0.115\linewidth}
\includegraphics[width=1\linewidth]{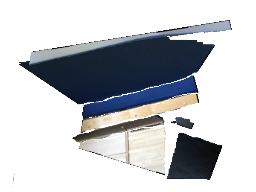}
\includegraphics[width=1\linewidth]{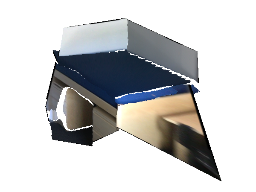}
\includegraphics[width=1\linewidth]{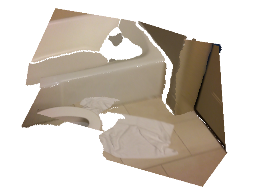}
\includegraphics[width=1\linewidth]{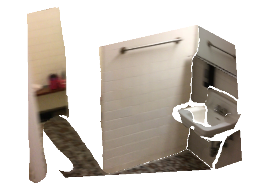}
\includegraphics[width=1\linewidth]{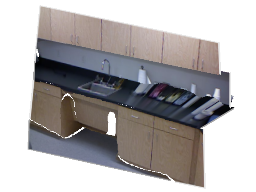}
\includegraphics[width=1\linewidth]{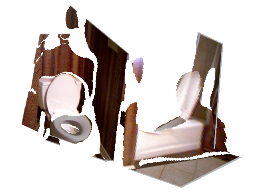}
\includegraphics[width=1\linewidth]{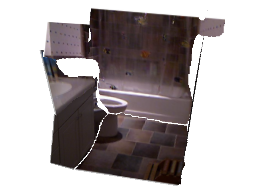}
\end{minipage}
}
\subfigure[{\scriptsize Ours 3D}]{
\begin{minipage}[b]{0.115\linewidth}
\includegraphics[width=1\linewidth]{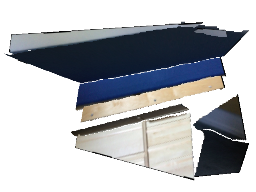}
\includegraphics[width=1\linewidth]{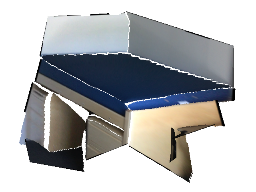}
\includegraphics[width=1\linewidth]{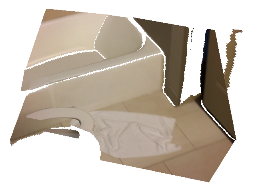}
\includegraphics[width=1\linewidth]{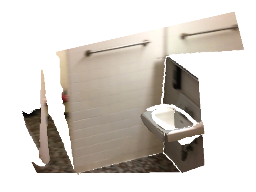}
\includegraphics[width=1\linewidth]{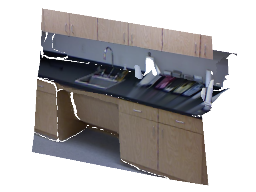}
\includegraphics[width=1\linewidth]{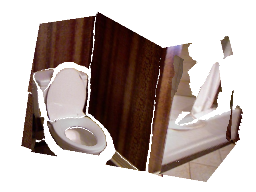}
\includegraphics[width=1\linewidth]{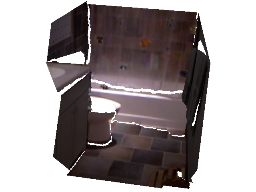}
\end{minipage}}
\subfigure[{\scriptsize GT 3D}]{
\begin{minipage}[b]{0.115\linewidth}
\includegraphics[width=1\linewidth]{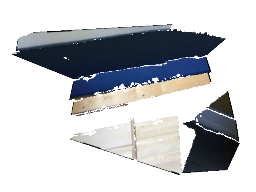}
\includegraphics[width=1\linewidth]{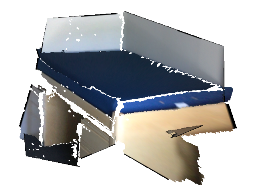}
\includegraphics[width=1\linewidth]{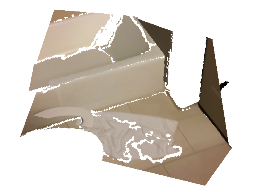}
\includegraphics[width=1\linewidth]{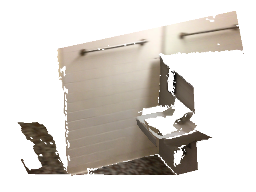}
\includegraphics[width=1\linewidth]{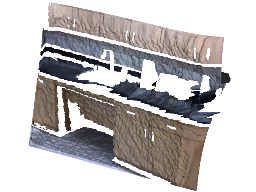}
\includegraphics[width=1\linewidth]{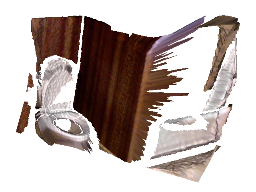}
\includegraphics[width=1\linewidth]{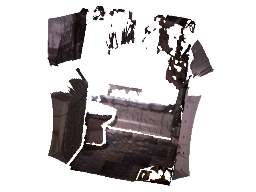}
\end{minipage}}
\caption{Comparison of plane reconstruction results on the ScanNetv1 (top \jiajia{4} rows) and NYUv2-Plane (bottom \jiajia{3} rows) datasets. 
}
\vspace{-5pt}
\label{fig:planerectr_compare_3dres}
\end{figure*}

%% file: figures/planerectr_scannet_recall.tex
\begin{figure*}[!t]
\centering
\includegraphics[width=0.23\linewidth]{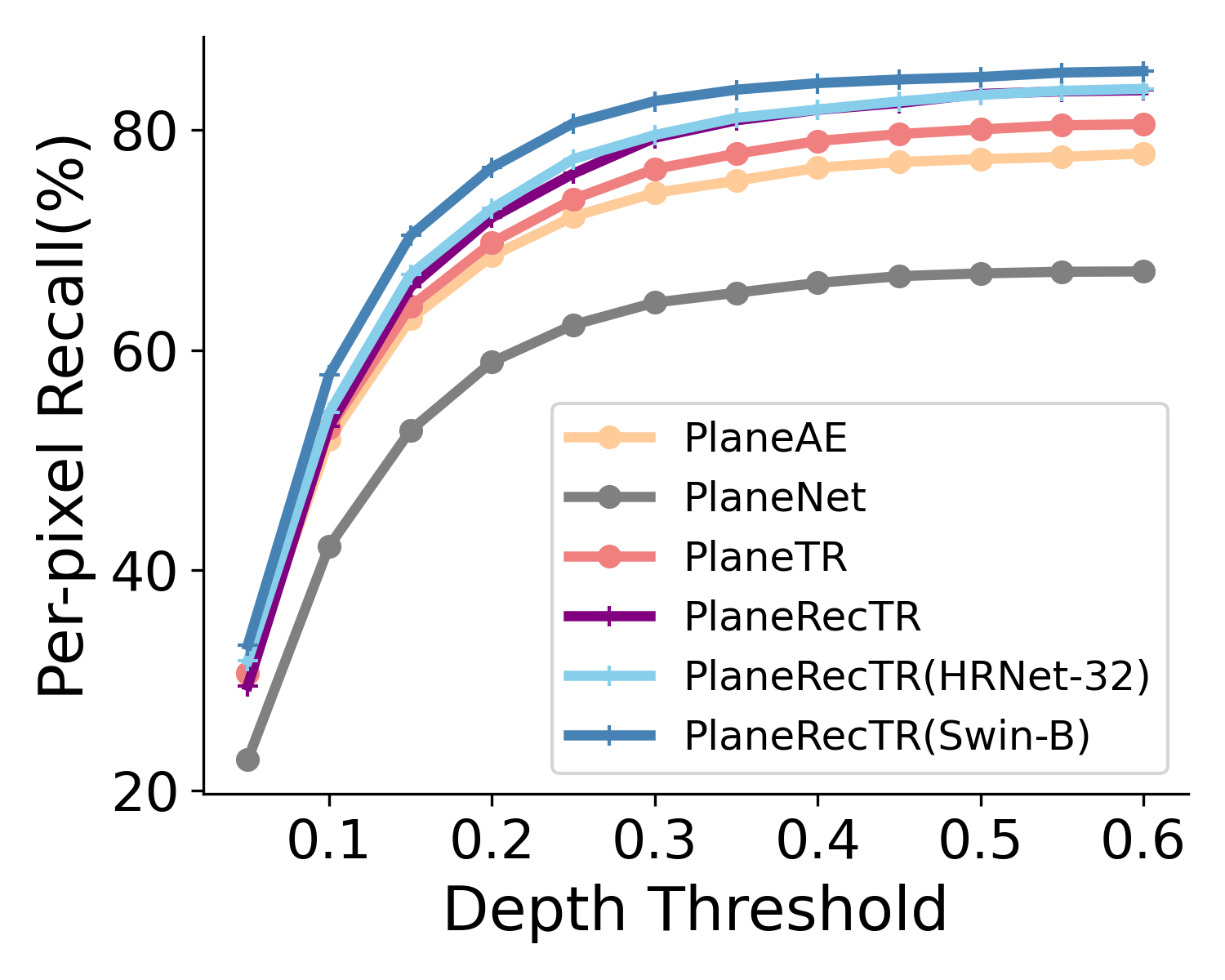} \hspace{-4pt}
\includegraphics[width=0.23\linewidth]{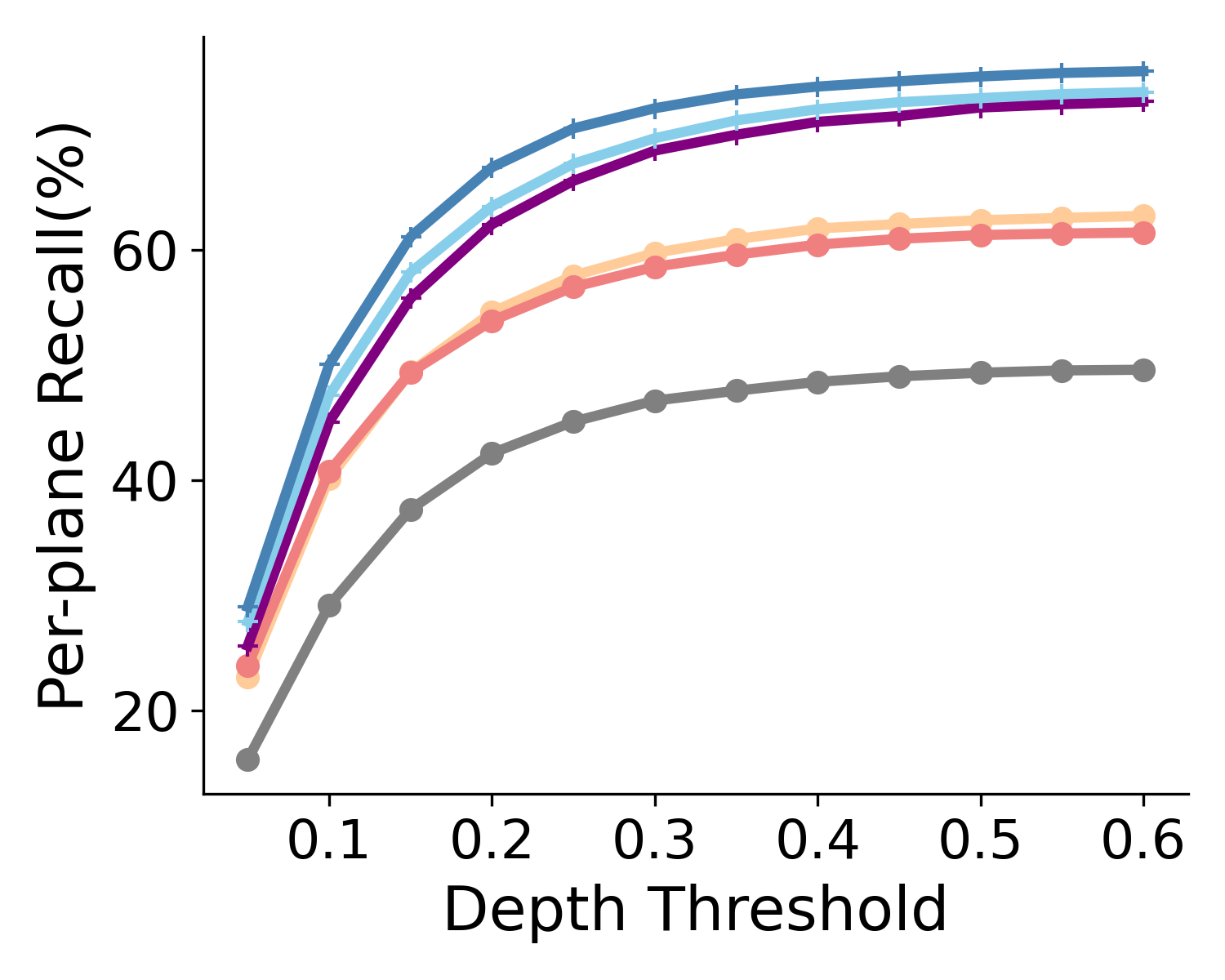}
\includegraphics[width=0.23\linewidth]{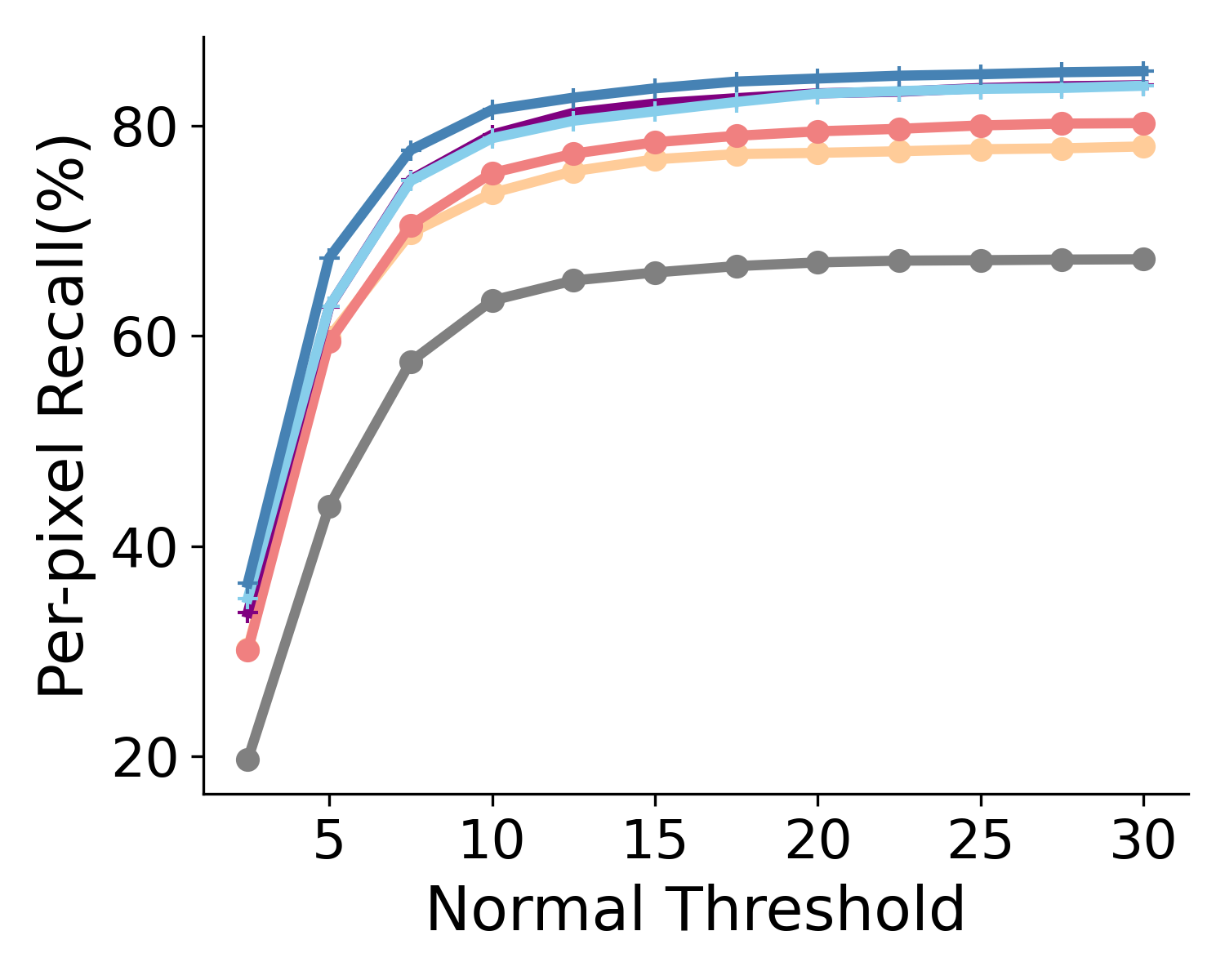}
\includegraphics[width=0.23\linewidth]{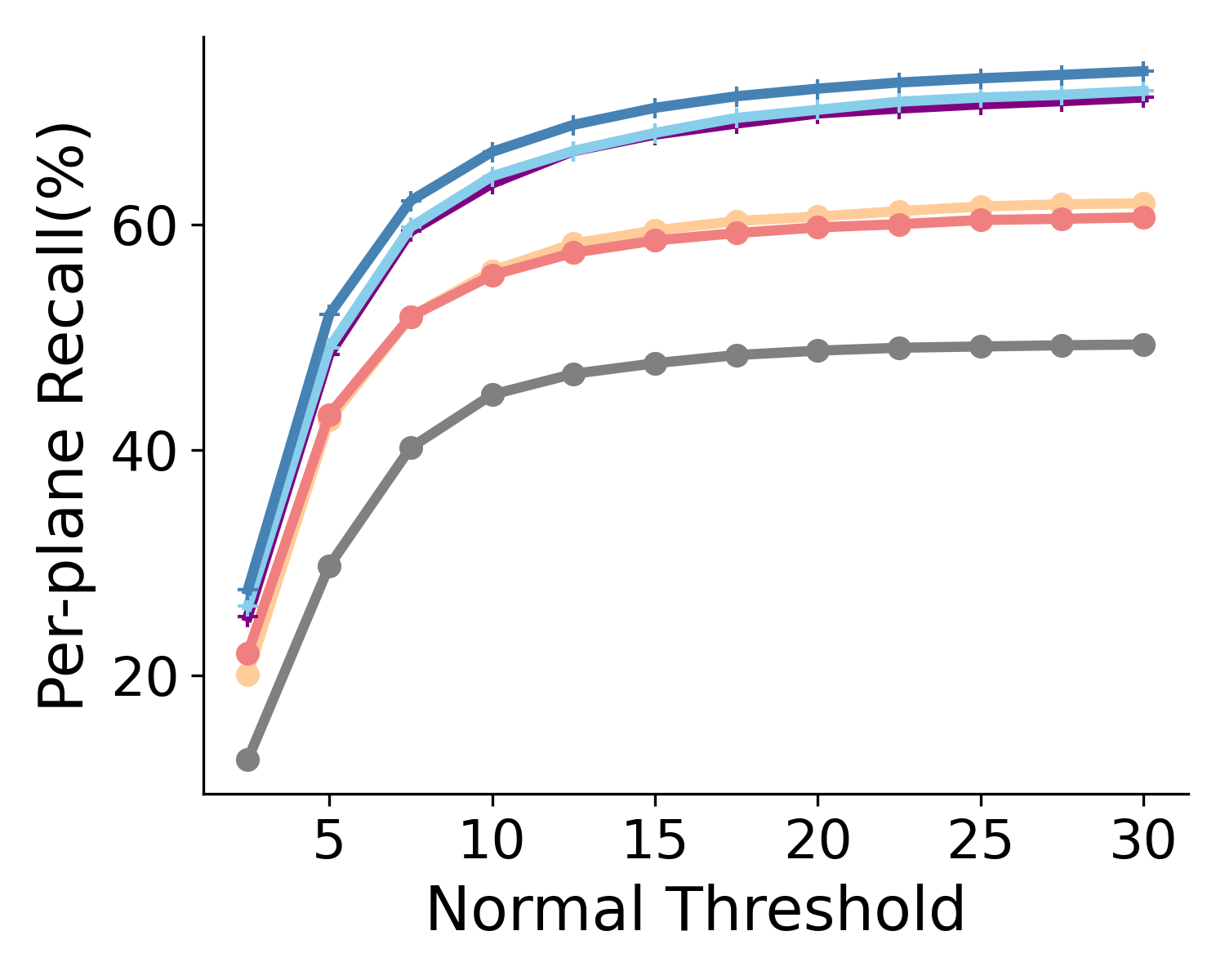}
\vspace{-5pt}
\caption{Per-pixel and per-plane recalls on the ScanNetv1 dataset.
}
\vspace{-10pt}
\label{fig:planerectr_scannet_recall}
\end{figure*}

%% file: figures/planerectr_backbones.tex
\begin{figure*}
\centering
\subfigure[{\scriptsize Input}]{
\begin{minipage}[b]{0.115\linewidth}
\includegraphics[width=1\linewidth]{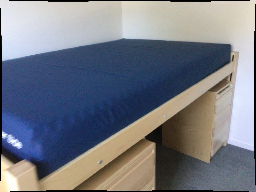}
\includegraphics[width=1\linewidth]{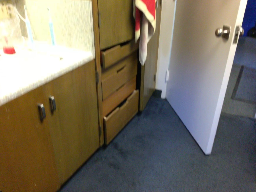}
\includegraphics[width=1\linewidth]{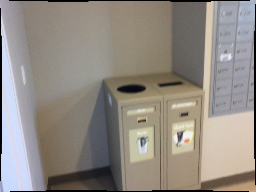}
\includegraphics[width=1\linewidth]{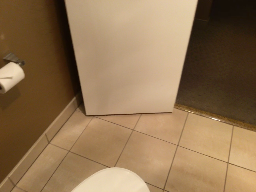}
\includegraphics[width=1\linewidth]{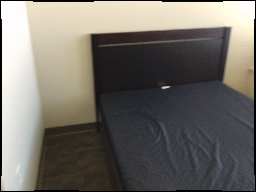}
\end{minipage}}
\subfigure[{\scriptsize ResNet50 Mask}]{
\begin{minipage}[b]{0.115\linewidth}
\includegraphics[width=1\linewidth]{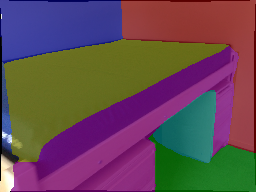}
\includegraphics[width=1\linewidth]{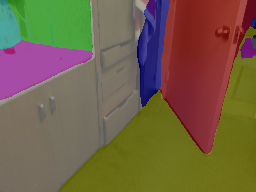}
\includegraphics[width=1\linewidth]{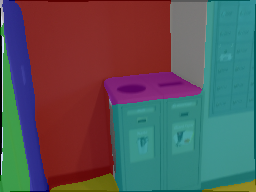}
\includegraphics[width=1\linewidth]{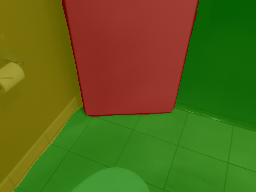}
\includegraphics[width=1\linewidth]{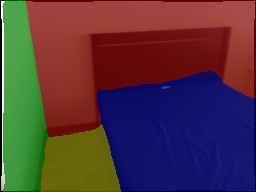}

\end{minipage}
}
\subfigure[{\scriptsize Swin-B Mask}]{
\begin{minipage}[b]{0.115\linewidth}
\includegraphics[width=1\linewidth]{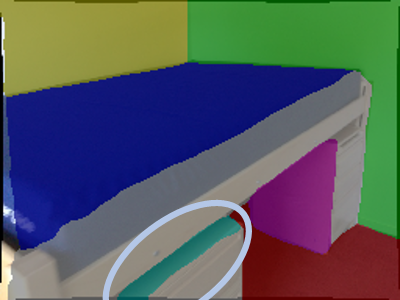}
\includegraphics[width=1\linewidth]{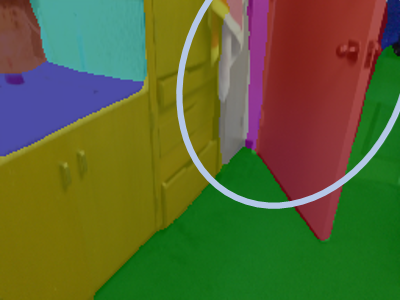}
\includegraphics[width=1\linewidth]{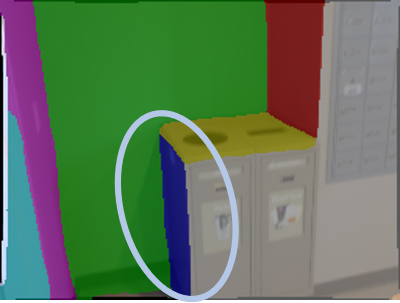}
\includegraphics[width=1\linewidth]{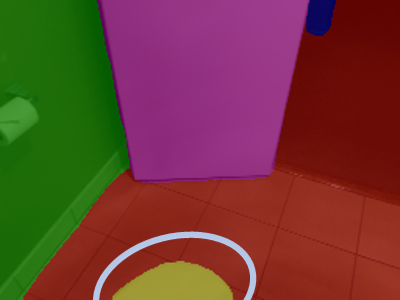}
\includegraphics[width=1\linewidth]{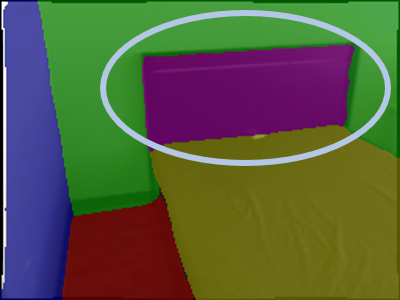}

\end{minipage}
}
\subfigure[{\scriptsize GT Mask}]{
\begin{minipage}[b]{0.115\linewidth}
\includegraphics[width=1\linewidth]{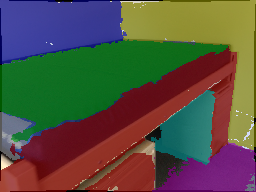}
\includegraphics[width=1\linewidth]{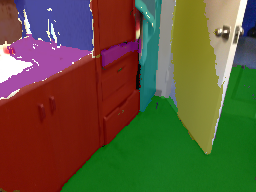}
\includegraphics[width=1\linewidth]{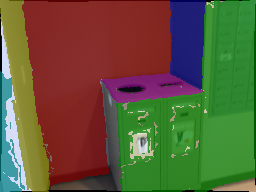}
\includegraphics[width=1\linewidth]{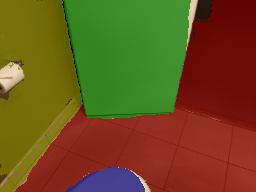}
\includegraphics[width=1\linewidth]{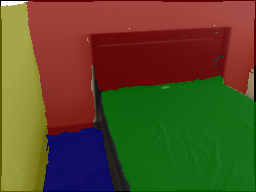}

\end{minipage}
}
\subfigure[{\scriptsize ResNet50 Depth}]{
\begin{minipage}[b]{0.115\linewidth}
\includegraphics[width=1\linewidth]{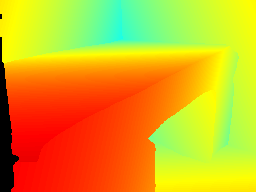}
\includegraphics[width=1\linewidth]{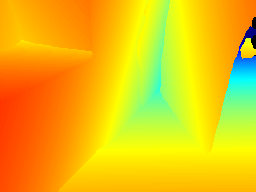}
\includegraphics[width=1\linewidth]{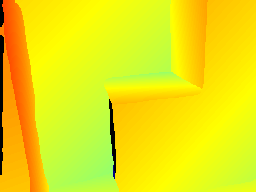}
\includegraphics[width=1\linewidth]{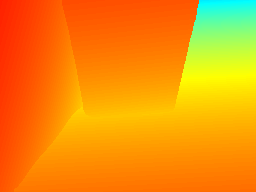}
\includegraphics[width=1\linewidth]{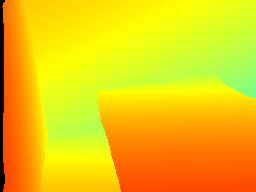}

\end{minipage}
}
\subfigure[{\scriptsize Swin-B Depth}]{
\begin{minipage}[b]{0.115\linewidth}
\includegraphics[width=1\linewidth]{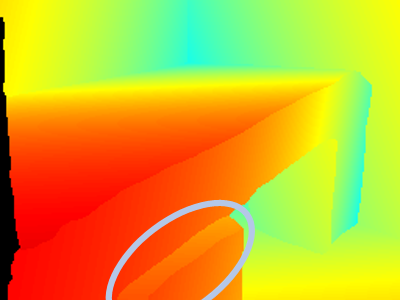}
\includegraphics[width=1\linewidth]{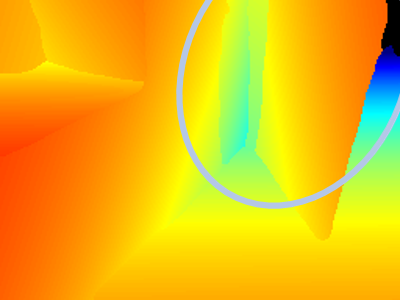}
\includegraphics[width=1\linewidth]{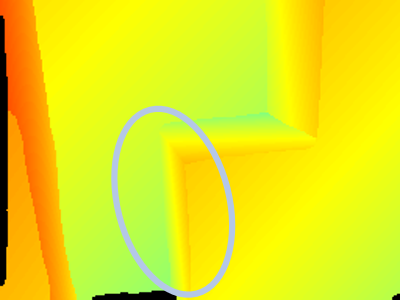}
\includegraphics[width=1\linewidth]{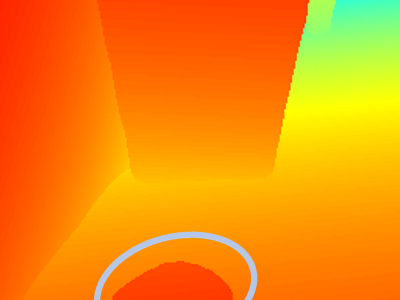}
\includegraphics[width=1\linewidth]{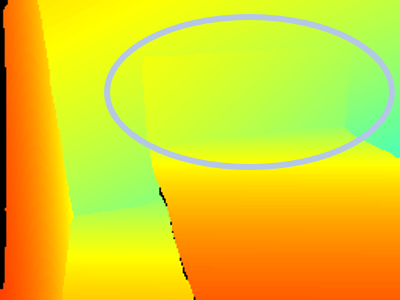}

\end{minipage}}
\subfigure[{\scriptsize GT Depth}]{
\begin{minipage}[b]{0.115\linewidth}
\includegraphics[width=1\linewidth]{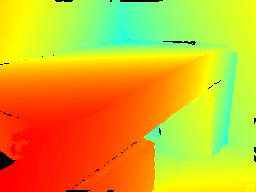}
\includegraphics[width=1\linewidth]{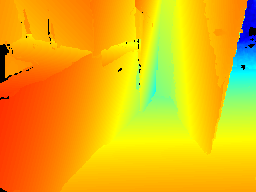}
\includegraphics[width=1\linewidth]{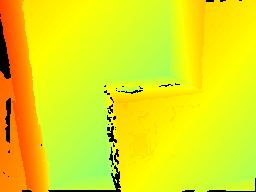}
\includegraphics[width=1\linewidth]{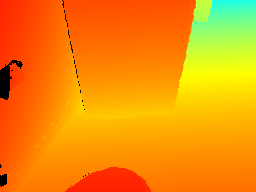}
\includegraphics[width=1\linewidth]{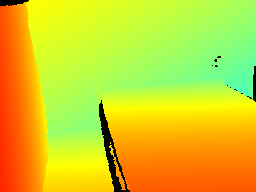}

\end{minipage}}
\caption{Comparison of plane reconstruction results with different backbones on the ScanNetv1 dataset.}
\vspace{-10pt}
\label{fig:planeretr_backbones}
\end{figure*}

%% file: tables/planerectr_all_ablation.tex
\begin{table}[t]
\centering
\caption{Ablations of our unified learning set-up on the ScanNetv1 dataset.}
\vspace{-5pt}
\LARGE
\resizebox{1.0\linewidth}{!}
{    \renewcommand{\arraystretch}{1.3} 
    \begin{tabular}{ccccccc}
    \toprule
        \multirow{3}{3cm}{\centering {}\\{\huge Method}} & \multicolumn{4}{c}{Per-Pixel/Per-Plane Recalls~$\uparrow$ } & \multicolumn{2}{c}{Plane Parameters}  \\ 
         \cmidrule(l{15pt}r{15pt}){2-5}
        &\multicolumn{2}{c}{Depth}&\multicolumn{2}{c}{Normal} & \multicolumn{2}{c}{Estimation Errors~$\downarrow$}\\
        \cmidrule(l{15pt}r{15pt}){2-3}\cmidrule(l{15pt}r{15pt}){4-5}\cmidrule(l{15pt}r{15pt}){6-7}

         & @0.10 m& @0.60 m& @$5^{\circ}$ & @$30^{\circ}$ & Normal (°) & Offset (mm)\\\midrule
         
        PlaneRecTR (-M-D) & - & - & - & - &  10.97 & 177.69\\
        PlaneRecTR(-D)&50.28/43.36  & 82.87/72.29 &61.42/47.80 & 83.44/71.11 & 10.27 (-0.70) & 166.85 (-10.84) \\
        PlaneRecTR& \textbf{53.07/45.07} & \textbf{83.60/72.84} & \textbf{62.75/48.48} & \textbf{83.85/71.33} & \textbf{10.23} (\textcolor{red}{-0.74}) & \textbf{165.13}(\textcolor{red}{-12.56}) \\\bottomrule
    \end{tabular}
}
\vspace{-10pt}
\label{tab:planerectr_all_ablation}
\end{table}

%% file: figures/ablation_planerectr_mask_compare.tex
\begin{figure}
\subfigure[{\scriptsize Input}]{
\begin{minipage}[b]{0.22\linewidth}
\includegraphics[width=1\linewidth]{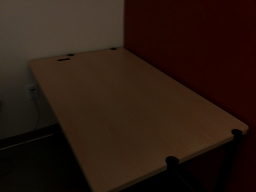}
\includegraphics[width=1\linewidth]{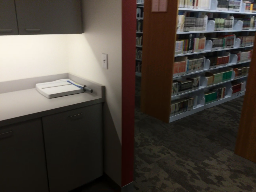}
\includegraphics[width=1\linewidth]{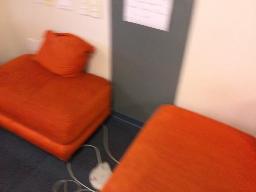}
\includegraphics[width=1\linewidth]{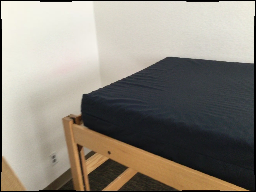}
\end{minipage}}
\subfigure[{\scriptsize Ours (-P-D)}]{
\begin{minipage}[b]{0.22\linewidth}
\includegraphics[width=1\linewidth]{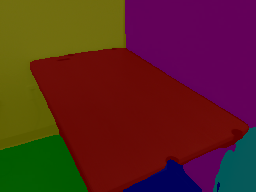}
\includegraphics[width=1\linewidth]{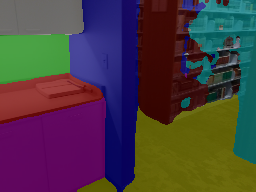}
\includegraphics[width=1\linewidth]{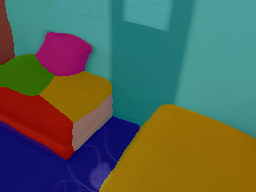}
\includegraphics[width=1\linewidth]{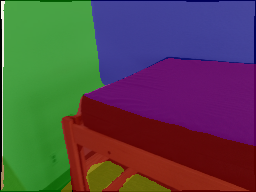}
\end{minipage}
}
\subfigure[{\scriptsize Ours}]{
\begin{minipage}[b]{0.22\linewidth}
\includegraphics[width=1\linewidth]{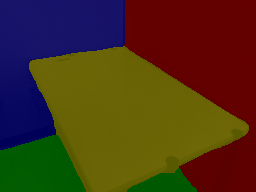}
\includegraphics[width=1\linewidth]{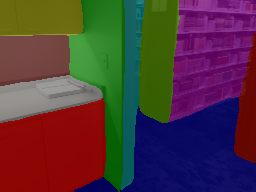}
\includegraphics[width=1\linewidth]{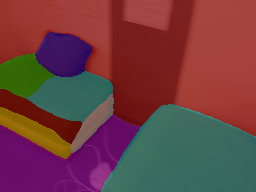}
\includegraphics[width=1\linewidth]{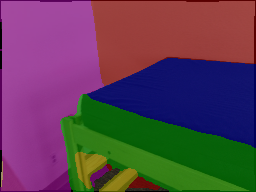}
\end{minipage}
}
\subfigure[{\scriptsize GT}]{
\begin{minipage}[b]{0.22\linewidth}
\includegraphics[width=1\linewidth]{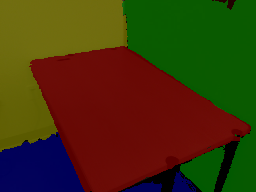}
\includegraphics[width=1\linewidth]{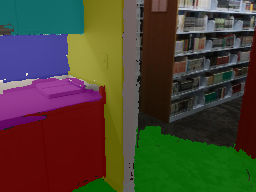}
\includegraphics[width=1\linewidth]{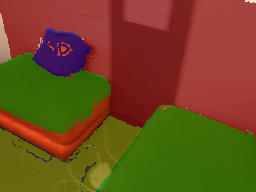}
\includegraphics[width=1\linewidth]{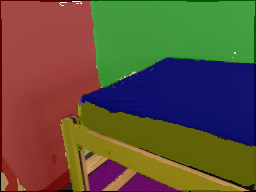}
\end{minipage}
}
\caption{Plane segmentation of PlaneRecTR(-P-D) and PlaneRecTR.}
\vspace{-10 pt}
\label{fig:ablation_planerectr_mask_compare}
\end{figure}

%% file: 2views_experiments.tex
\section{Sparse View Experiments}
\label{sec:corrattn_experiments}

\paraspace
\ptitle{Datasets.}
We adopt the setup of various existing works \cite{jin2021sparseplanes, tan2023nopesac} to ensure a fair comparison, 
\jiajia{which includes two large-scale sparse view datasets as benchmarks.}

\textbf{ScanNetv2 Dataset.}
We use a more challenging sparse view split of the ScanNetv2 dataset created by \cite{Dai:etal:CVPR2017,Dai:scannetv2web,Liu:etal:CVPR2019:Planercnn,tan2023nopesac}, consisting of 17,237/4,051 image pairs from 1,210/303 non-overlapping scenes for training/testing. 
Different to ScanNetv1 for monocular plane recovery, ScanNetv2 was proposed to contain a lower overlapping ratio between frames and more complex camera rotation distributions. The average overlap ratio of adjacent frames were 20.6\% and 18.6\% in the training and test sets, respectively. The image size is 640 × 480 for all baselines.

\textbf{MatterPort3D Dataset.}
In contrast to ScanNetv2, the sparse view version of the MatterPort3D dataset\cite{chang2017matterport3d} is created in a semi-simulated manner, whose RGB images are rendered from approximated 3D planar meshes during the generation process of plane annotations \cite{jin2021sparseplanes}, thereby mitigating real-world impact like lighting variation. 
However, these rendered RGB images consist solely of planes along with numerous small erroneous facets caused by approximated 3D planar meshes, increasing \jiajia{the difficulty for accurate plane detection. Therefore, evaluations on MatterPort3D may provide unfair advantages to all two-stage baselines \cite{jin2021sparseplanes, agarwala2022planeformers, tan2023nopesac} with a dense pixel-based pose initialization, and cause challenges to plane-based methods (like ours). We have confirmed such findings in subsequent experiments and regard MatterPort3D mainly as a complement benchmark compared to ScanNetv2 dataset.} The training set and testing set consist of 31,932 and 7,996 image pairs, respectively, exhibiting an overlap ratio of about 21.0\% \cite{tan2023nopesac}. The image size is also kept to 640 × 480.

\paraspace
\ptitle{Evaluation Metrics.}
To evaluate relative camera pose, we use geodesic distance and euclidean distance to measure rotation and translation error, respectively. We also employ three popular statistical measurements: the median to reflect overall prediction accuracy, the mean to account for large outlier errors, and the percentage of errors below a certain threshold.

To assess the overall planar reconstruction performance, we initially present the predicted results obtained from monocular images on the above datasets, utilizing two metrics: plane segmentation and plane recovery recall from Section \ref{sec:experiments}. For the merged 3D reconstruction results derived from two adopted views, we employ the average precision (AP) metric \cite{jin2021sparseplanes}, treating each reconstructed 3D plane as a detection target for evaluation.    A true positive of 3D plane detection necessitates three conditions: (i) (Mask) an intersection-over-union value to the ground-truth mask $\geq 0.5$;    (ii) (Normal) arccos value between the predicted and ground truth normal $\leq \alpha$ where $\alpha \in \{30^{\circ}, 15^{\circ}, 5^{\circ}\}$; and (iii) (Offset) absolute difference between predicted and ground truth offset $\leq \beta$ where $\beta \in \{1\text{m}, 0.5\text{m}, 0.2\text{m}\}$.
To further assess the network's capability in implicitly learning plane-aware correspondences during reconstruction, Section \ref{sec:unified_plane_emb} presents the number of true positives (TP) and calculates precision, recall, and F-score for plane correspondences.

\subsection{Baselines and their Variants}
3D planar reconstruction from sparse views demands the resolution of multiple sub-tasks, encompassing relative camera pose estimation, monocular planes recovery and multi-view plane matching.  We conducted a comparative analysis between representative baselines in terms of both overall results and intermediate outcomes.

\paraspace
\ptitle{Public Baselines} We compared our method with several state-of-the-art and leading sparse views planar reconstruction methods, including SparsePlanes \cite{jin2021sparseplanes}, PlaneFormers \cite{agarwala2022planeformers}, and NOPE-SAC \cite{tan2023nopesac} which are all multi-stage approaches. Specifically, all these baselines begin by estimating one or more initial camera poses from learned dense features of two input frames, and subsequently refine these camera poses using predicted planes via post optimization algorithms\cite{jin2021sparseplanes},  SIFT-like keypoints \cite{jin2021sparseplanes}, or separate network models\cite{agarwala2022planeformers, tan2023nopesac}.
Therefore, their primary contribution usually lies in an elaborated pose refinement module while treating monocular plane prediction and initial camera pose prediction as given knowledge from external and distinct modules. 
These facts are also our main differences by performing two-view planar reconstruction in a purely end-to-end manner from images without any external priors and matching supervision. 

As to the evaluation of camera pose estimation, we additionally consider two popular baselines SuperGlue \cite{sarlin2020superglue} and Pose ViT \cite{rockwell20228posevit}. SuperGlue is a competitive graph-based point matching network that leverages the estimated essential matrix with a RANSAC solver to obtain the relative camera pose. Therefore, SuperGlue inherits the intrinsic scale ambiguity of the Essential Matrix,  so we solely focus on comparing the rotation error during its evaluation. Pose Vit uses embeddings of tokenized image patches extracted from a ViT \cite{vit2021} and follows the principle of the Eight-Point Algorithm \cite{hartley1997defense} via another ViT module to directly estimate relative camera pose between two input images. Our proposed PlaneRecTR++ draws inspiration from its design philosophy but brings several important modifications to better fit unified query learning as well as end task. 
Specifically, we choose to use plane-aware embeddings instead of those from raw image patches,  even without requiring the Quadratic Position Encodings used in \cite{rockwell20228posevit}. Our method also introduces distinct differences in the attention structure, enhancing interpretability and yielding better performance.  

\input{tables/planerectr++_scannetv2_mp3d_seg}
\input{tables/planerectr++_scannetv2_mp3d_recall}

\paraspace
\ptitle{Improved Baseline Variants} The monocular plane predictor of SparsePlanes and PlaneFormers is based on PlaneRCNN\cite{Liu:etal:CVPR2019:Planercnn}, while \jiajia{NOPE-SAC} utilizes an improved version of the leading PlaneTR \cite{Tan:etal:ICCV2021:Planetr} (denoted as PlaneTRP) to recover monocular planes. Motivated by \cite{tan2023nopesac}, to make fair comparisons, we include improved baseline variants by replacing the PlaneRCNN backbone of SparsePlanes and PlaneFormers with a more powerful PlaneTRP module, termed SparsePlanes-TRP and PlaneFormers-TRP.


\input{tables/pose_comparison}
\input{tables/2views_planerec_ap}

\paraspace
\ptitle{NOPE-SAC and its Variants} 
The NOPE-SAC framework comprises four distinct modules\jiajia{: (1) A} Monocular Plane Predictor (PlaneTRP); \jiajia{(2) An attention} network for \jiajia{dense pixel-based} pose initialization; \jiajia{(3) A supervised} differentiable plane matching module based on optimal transport; \jiajia{(4) A neural RANSAC network for plane-level pose refinement.}
To provide a more comprehensive comparison in pose estimation to the leading NOPE-SAC \cite{tan2023nopesac}, we further introduce its two variants to better isolate the contribution of key modules.

\textbf{NOPE-SAC Init.} refers to the camera pose initialization network to kick off its overall procedure, \ie, module (2) above, which relies solely on dense pixel features. We present its pose accuracy to indicate the quality of learned pose prior \cite{tan2023nopesac}, acting as a similar role to external pose predictor in \cite{jin2021sparseplanes, agarwala2022planeformers}. 

\textbf{NOPE-SAC Ref.} \jiajia{refers to NOPE-SAC without module (2) and mainly relies on the pose refinement module for pose estimation. NOPE-SAC Ref. performs plane matching without initial pose and directly estimates relative pose from sparser views through the neural RANSAC network. Specifically, NOPE-SAC Ref. sets the initial pose to an identity matrix and cuts off initial pose related geometric score function during plane matching, while maintaining the appearance score function intact. Despite retaining a multi-stage approach,  NOPE-SAC Ref. shares consensus with ours in that it relies solely on \textit{monocular plane features} for plane matching and pose regression from \textit{input sparse views}.}

In contrast to NOPE-SAC and other existing baselines, one thing worth highlighting is that our unified plane embedding demonstrates multi-view consistency and implicitly accomplishes plane matching, \textit{without any external supervision for plane matching or the need for initial pose assistance}. In Section \ref{sec:unified_plane_emb}, we compare our method with optimal transport (OT) within NOPE-SAC\cite{tan2023nopesac} requiring external priors.

\subsection{Implementation Detail}

The two-phase training process of sparse views reconstruction, as described in Section \ref{sec:overview}, begins with pre-training the model on the Scannetv2 and Matterport3d datasets following Section \ref{sec:monocular_imp_detail}. Subsequently, we employ nearly identical training configurations to jointly train the entire model with 42 epochs and a 10-fold reduction in loss for monocular planes.

\input{figures/2view_rec}

\subsection{Evaluation of Monocular Planes}
\label{sec:eval_mono}
In this section, we compare the monocular plane predictions to first highlight the effectiveness of our intra-frame plane query learning component after joint optimization. In Table \ref{tab:planerectr++_scannetv2_mp3d_seg} and Table \ref{tab:planerectr++_scannetv2_mp3d_recall}, we initially evaluated the performance for single-view plane segmentation and geometry, similar to Section \ref{sec:experiments}. 

On both datasets, our method demonstrated superior monocular plane prediction accuracy compared to the state-of-the-art PlaneTRP \cite{Tan:etal:ICCV2021:Planetr, tan2023nopesac}. Furthermore, as discussed in Section \ref{sec:corrattn_experiments}, the MatterPort3D dataset \cite{chang2017matterport3d} presents a greater challenge for plane prediction compared to the ScanNetv2 dataset, leading to a moderate performance degradation for all methods.

We also present our monocular plane prediction results exclusively from the monocular pre-training phase (\ie, without the joint optimization stage using pose losses), denoted as Ours (Monocular). There is almost indistinguishable difference in terms of the monocular prediction accuracy (rows 2-3 of Table \ref{tab:planerectr++_scannetv2_mp3d_seg} and Table \ref{tab:planerectr++_scannetv2_mp3d_recall}). This observation signifies the feasibility of our overall pipeline that even though our unified plane embeddings have incorporated evident multi-view consistency for pose estimation (Section \ref{sec:pose_evaluation}), there still remains a comparable ability for monocular plane prediction after the comprehensive joint training phase.

\input{figures/2view_rec_compare}

\subsection{Relative Camera Pose Evaluation}
\label{sec:pose_evaluation}

\paraspace
\ptitle{Quantitative Results.} In Table \ref{tab:pose_comparison}, our PlaneRecTR++ demonstrates superior performance in camera pose \textit{across all metrics} on the realistic ScanNetv2 dataset compared to other methods. On the MatterPort3D dataset, our approach achieves overall comparable results to the \jiajia{leading NOPE-SAC}, exhibiting better translation accuracy while slightly lagging behind in rotation estimation.

\jiajia{The relatively poor rotation performance can primarily be attributed to the simulated characteristics of input images on the MatterPort3D dataset, which contains numerous erroneous tiny planes that challenge plane detection of all methods (see Section \ref{sec:eval_mono}) but enhance cross-view photometric consistency during rendering (see dataset introduction in Section \ref{sec:corrattn_experiments}).  
During inference on MatterPort3D, the dense pixel-based NOPE-SAC Init. and other pose initialization networks within baseline methods \cite{jin2021sparseplanes, agarwala2022planeformers, tan2023nopesac} could provide a superior initial rotation, transforming the input sparse views into closer views and thus lowering the difficulty of following plane-level pose predictions.
} 

\jiajia{It is noteworthy that even on this challenging MatterPort3D dataset, our median rotation error remains smaller than that of NOPE-SAC, which indicates smaller typical prediction errors.}

\paraspace
\ptitle{Qualitative Results.} 
Figure \ref{fig:2view_rec} visually illustrates the relative camera pose estimates of our PlaneRecTR++ from two different viewpoints (last two columns) on the ScanNetV2 and MatterPort3D datasets.  Our method can accurately recover a precise relative camera pose from input sparse views, even in scenarios with extremely low image overlap (rows 3-8), without relying on initial pose estimation and explicit corresponding plane pairs.
Figure \ref{fig:2view_rec_compare} presents the predicted pose comparison of ours and NOPE-SAC\cite{tan2023nopesac}, showing that our method recovers a more accurate relative camera pose than the leading baseline.

\subsection{3D Planar Reconstruction Evaluation}

\paraspace
\ptitle{Quantitative Results.}
The numerical evaluation on the final 3D reconstruction is shown in Table \ref{tab:3DPlane}.  Our unified single-stage approach demonstrates superior performance compared to all other multi-stage methods, particularly exhibiting a significant enhancement in the reconstruction accuracy on the real-world dataset ScanNetv2.

\paraspace
\ptitle{Qualitative Results.} 
Figure \ref{fig:2view_rec} presents the visualization of the 3D plane reconstruction results achieved by PlaneRecTR++ on the ScanNetv2 \cite{Dai:etal:CVPR2017, Dai:scannetv2web} and the MatterPort3D \cite{chang2017matterport3d} datasets. Note that our model implicitly learns plane matching during pose inference, and we further extract and process the probability distributions of plane correspondences from our model (see 3rd column). \jiajia{Our method exhibits superior performance in plane matching and reconstruction even in the presence of extremely sparse views, and is capable of identifying and exploiting the discontinuous ground (rows 5,7,8).}  In Figure \ref{fig:2view_rec_compare}, a \jiajia{further} comparison between our proposed PlaneRecTR++ and the current state-of-the-art NOPE-SAC \cite{tan2023nopesac} is provided on both datasets, demonstrating that our method yields more precise plane reconstructions and recovers more accurate relative camera poses from sparse views.
Even in more challenging scenarios characterized by inconsistent brightness (column 1,3), confusion caused by symmetrical repetitive patterns (column 6) and so on, our method can implicitly acquire robust correspondences for reconstruction and pose recovery, which is outperforms NOPE-SAC that relies on initial pose prior and matching supervision.

\input{tables/pose_model_ablation}
\input{tables/corr_ablation}

\paraspace
\ptitle{Inference Time.} 
We calculate the average inference time of the latest neural methods for joint planar reconstruction and pose estimation on a NVIDIA TITAN V GPU. 
Our single-stage PlaneRecTR++ ($0.258$ s) achieves higher inference speed than previous multi-stage NOPE-SAC ($0.274$ s) and PlaneFormers ($4.257$ s), thanks to the exemption from external pose initialization modules \cite{jin2021sparseplanes, agarwala2022planeformers, tan2023nopesac} and iterative refinement modules \cite{agarwala2022planeformers}.
\subsection{Ablation Studies of Model Designs}
\label{sec:sparseview_ablation}

We conducted extensive ablation studies to investigate the contributions of each design choice, particularly within our plane aware cross attention layer in Section \ref{sec:plane_cross_attention}. We focus on experimenting with the following two aspects: (1) cross embedding structure (CE) within the bilinear attention mechanism, and (2) the specialized design of query, key and value subdivision.

\paraspace
\ptitle{Cross Embeddings.}
Our method differs from Pose ViT \cite{rockwell20228posevit}, which also employs bilinear attention, in that we cross-place plane embeddings of different input images on both sides of the bilinear attention matrix, while Pose ViT places visual features and positional encodings of the same input image, which experimentally yields better results as shown in \cite{rockwell20228posevit}. We consider that our cross plane embeddings placement follows a more intuitive discipline and better performance. To validate our idea, we follow  \cite{rockwell20228posevit} and shift our plane aware cross attention layer's structure to the same embedding placement strategy.  The experimental findings (rows 1, 2 of Table \ref{tab:pose_model_ablation} and \ref{tab:corr_ablation}) show a significant degradation in performance without proposed cross embeddings set-up.

We believe the key reason for the differences between ours and Pose ViT lies in whether the network truly learns plane correspondences. In both methods, it is widely anticipated that the similarity attention matrix would serve as the function for the object assignment matrix. 
However, only our plane aware cross attention design empowers the network to effectively execute authentic plane-level matching, considering that it is less clear to generate plausible patch-wise correspondences of two sparse views.  Consequently, in our method, cross embedding placement naturally yields superior performance compared to Pose ViT, wherein this prerequisite is not met and may even impede efficient passing of visual features through cross-embedding attention.

\input{figures/heatmap}
\input{figures/multiview_3dres}

\paraspace
\ptitle{Query, Key and Value Designs.} We show the effectiveness of our query, key and value designs by evaluating two model variants on pose, plane correspondences and reconstruction, respectively. 

In  Table \ref{tab:pose_model_ablation}, on both datasets, pose accuracy of PlaneRecTR++ with the unsplit query and key design is more precise when maintaining VNum. is 4 (rows 1, 3). In Figure \ref{subfig:qk1v4} and \ref{subfig:qkv4}, the highlighted areas of the plane correspondence attention matrix $\rm{C}(Q_i, K_j)$ using our unsplit key and query, align well with the ground truth correspondence. However, the distribution of 4 similarity attention matrices, each computed using one of the 4 split query and key segments, does not effectively capture the ground truth pattern. Though the combination of 4 similarity matrices can approximate actual plane correspondence distribution, it is still inferior to ours caused by more introduced noisy matches with high probabilities. 
We consider its potential in capturing real correspondences via a post-hoc evaluation, where we select the one head with the highest matching accuracy to ground truth and compare it with our method.  As presented in rows 1, 3 of two datasets in Table \ref{tab:corr_ablation},  even after carefully selecting the best possible matches from the split query and key pairs, the performance is still inferior to ours adopting unsplit query and key pairs.

Moreoever, in Table \ref{tab:pose_model_ablation} and \ref{tab:corr_ablation}, when the key and query are guaranteed to be complete, the accuracy of all metrics with VNum. $=4$ still surpasses that with VNum. $=1$ (rows 1, 4), indicating a positive contribution from the partition of value term.

On the whole, we have validated that our design not only retains the advantages of multi-head attention in standard Transformer, but also effectively captures the distribution of plane correspondence and further enhancing model performance.



\vspace{-0.1cm}
\subsection{Studies of Unified Plane Embedding}
\label{sec:unified_plane_emb}

During the inter-frame plane query learning stage, we have actually conducted several experiments to enhance the capability of input plane embedding with auxiliary knowledge. Such attempts include concatenating cosine positional encoding \cite{sun2021loftr}, quadratic positional encoding of plane center \cite{rockwell20228posevit}, plane parameter encoding or plane appearance embedding along with original plane embedding. We also explored to filter out plane embedding sequence using their plane probability $p_i$, or to incorporate several self-attention layers \cite{rockwell20228posevit, Vaswani:etal:NIPS2017} to promote contextual features, or to introduce an explicit view consistency loss of planes and pose during training.
However, none of these variants yielded any obvious improvement in the current model's performance.

It became evident that our unified plane embedding, achieved through a simple combination of intra-frame and inter-frame plane query learning, already encompassed adequate information to address the task of sparse views planar reconstruction.

\input{figures/tsne}
\input{tables/corr_comparison}

\paraspace
\ptitle{Consistent Planar Attributes across Frames.}
After the initial single view training and following comprehensive sparse view training, rows 2,3 of Table \ref{tab:planerectr++_scannetv2_mp3d_seg} and Table \ref{tab:planerectr++_scannetv2_mp3d_recall} exhibit comparable performance in monocular plane detection on two datasets.   In the rows 3,4 of Table \ref{tab:corr_compare}, the former relies solely on similar appearance features from a single view and achieves poor matching results, whereas the latter computes a reasonable and precise plane correspondence. In Figure \ref{fig:tsne}, despite only being trained on input sparse views, our unified query embeddings exhibit promising consistency across more frames without the need for ground truth correspondence supervision. \jiajia{Figure \ref{fig:multiview_rec} exhibits our qualitative results that conspicuously outperform the leading NOPE-SAC on multiple views ($\geq 3$ views).}



\paraspace
\ptitle{Implicit Plane Matching.}
Compared with the differentiable optimal transport (OT) \cite{sarlin2020superglue} method from NOPE-SAC \cite{tan2023nopesac}, our approach owns the following advantages:
(1) Most importantly, we skip the requirement of pose initialization in all previous methods \cite{jin2021sparseplanes, agarwala2022planeformers, tan2023nopesac}. This means that there is no need for us to convert plane parameters into the same coordinate system before effectively utilizing planar geometry for matching.  We believe that this is a crucial factor for constructing a single-stage method.
(2) We do not require explicit supervision using \jiajia{the} ground truth correspondences. Instead, only through pose supervision, our carefully designed simple network structure actively learns multi-view consistency for plane embedding.  In a single forward pass, our method implicitly performs plane matching and probabilistically synthesizes pairwise plane features for pose prediction.  
It does not explicitly perform plane matching and input hard plane pairs to a pose refinement network \cite{jin2021sparseplanes, tan2023nopesac}.
(3) The correspondence attention matrix formed by our network can be processed using a simple MNN (maximum nearest neighbor) operation to obtain a hard plane assignment matrix.  The accuracy of this assignment matrix is comparable or even higher than previous methods with supervisions, as shown in Table \ref{tab:corr_compare}.

%% file: tables/planerectr++_scannetv2_mp3d_seg.tex
\begin{table}
\centering
\caption{Plane segmentation on the ScanNetV2 and MatterPort3D datasets. }
\vspace{-5pt}
\resizebox{0.95\linewidth}{!}{ 
    \begin{tabular}{c|ccc|ccc}
    \toprule
        \multirow{2}{*}{Method} & \multicolumn{3}{c|}{ScanNetV2} & \multicolumn{3}{c}{MatterPort3D} \\ 
                                & VI~$\downarrow$ & RI~$\uparrow$ & SC~$\uparrow$ & VI~$\downarrow$ & RI~$\uparrow$ & SC~$\uparrow$\\\midrule
        PlaneTRP \cite{Tan:etal:ICCV2021:Planetr, tan2023nopesac}          & 1.291 & 0.880 & 0.716 &1.458  & 0.897 & 0.683 \\ \midrule
        Ours (Monocular)       & 0.781  & \textbf{0.939} & 0.835 & \textbf{0.920}  & \textbf{0.934} & \textbf{0.773}  \\
        Ours
   &\textbf{0.777}  & 0.938  & \textbf{0.836} & 0.946 & 0.931 & 0.766  \\
    \bottomrule

    \end{tabular}
}
\vspace{-5pt}
\label{tab:planerectr++_scannetv2_mp3d_seg}
\end{table}

%% file: tables/planerectr++_scannetv2_mp3d_recall.tex
\begin{table}[t]
\centering
\centering
\caption{Per-pixel/plane recalls on the ScanNetv2 and MatterPort3D datasets.}
\vspace{-5pt}
\LARGE
\resizebox{1.0\linewidth}{!}
{    \renewcommand{\arraystretch}{1.3} 
    \begin{tabular}{ccccccc}
    \toprule
        \multirow{3}{3cm}{\centering {}\\{\huge Method}} & \multicolumn{4}{c}{Per-Pixel/Per-Plane Recalls~$\uparrow$ } & \multicolumn{2}{c}{Plane Parameters}  \\ 
         \cmidrule(l{15pt}r{15pt}){2-5}
        &\multicolumn{2}{c}{Depth}&\multicolumn{2}{c}{Normal} & \multicolumn{2}{c}{Estimation Errors~$\downarrow$}\\
        \cmidrule(l{15pt}r{15pt}){2-3}\cmidrule(l{15pt}r{15pt}){4-5}\cmidrule(l{15pt}r{15pt}){6-7}

         & @0.10 m& @0.60 m& @$5^{\circ}$ & @$30^{\circ}$ & Normal (°) & Offset (mm)\\\midrule

        \multicolumn{7}{c}{ScanNetv2 dataset} \\
        \midrule
         
        PlaneTRP\cite{Tan:etal:ICCV2021:Planetr, tan2023nopesac} & 14.17/10.97  & 64.73/52.20 & 37.38/27.59 &67.52/55.97  & 14.95 & 237.90 \\\midrule
        Ours (Monocular)& 24.27/18.91& \textbf{80.66}/\textbf{64.98} & 57.71/41.15 &\textbf{82.11}/\textbf{67.27} & \textbf{10.11} & 193.33  \\
        Ours & \textbf{25.51}/\textbf{19.72} &80.58/64.89  & \textbf{58.81}/\textbf{41.29}& 82.07/67.12  &10.32 & \textbf{191.09} \\\midrule

        \multicolumn{7}{c}{MatterPort3D dataset} \\
        \midrule
        PlaneTRP\cite{Tan:etal:ICCV2021:Planetr, tan2023nopesac} & 25.87/20.56 & 61.40/56.96 & 53.89/\textbf{47.51} & 64.85/62.89 & 10.67 & 390.38 \\\midrule
        Ours (Monocular) & \textbf{31.36}/\textbf{24.57} & \textbf{71.13}/\textbf{63.96}& 56.41/44.86&  \textbf{73.78}/67.79&8.64& 390.65\\
        Ours & 28.51/22.86 & 69.96/63.69 & \textbf{58.00}/46.91& 72.84/\textbf{67.99} & \textbf{8.47} & \textbf{384.59}\\
        \bottomrule
    \end{tabular}
}
\vspace{-10pt}
\label{tab:planerectr++_scannetv2_mp3d_recall}
\end{table}

%% file: tables/pose_comparison.tex
\begin{table*}[t!]
\centering
\caption{Comparison of relative camera pose on the ScanNetv2 dataset and the \jiajia{MatterPort3D} dataset.} 
\vspace{-5pt}
\resizebox{0.9\linewidth}{!}{ 
    \begin{tabular}{l|ccccc|ccccc}
    \toprule
        \multicolumn{1}{c|}{\multirow{2}{*}{Method}} & \multicolumn{5}{c|}{Translation} & \multicolumn{5}{c}{Rotation} \\ 
                                & Med.~$\downarrow$ & Mean~$\downarrow$ & ($\le$1m)~$\uparrow$ & ($\le$0.5m)~$\uparrow$& ($\le$0.2m)~$\uparrow$ & Med.~$\downarrow$ & Mean~$\downarrow$ & ($\le 30^{\circ}$)~$\uparrow$ & ($\le 15^{\circ}$)~$\uparrow$ & ($\le 10^{\circ}$)~$\uparrow$\\
        
        \midrule
        \multicolumn{10}{c}{ScanNetv2 dataset} \\
        \midrule
        SuperGlue~\cite{sarlin2020superglue} & - & - & - & - & - & 10.90 & 31.00 & 67.8\% & 56.0\% & 48.4\%\\
        NOPE-SAC Init.~\cite{tan2023nopesac}& 0.48 &0.72 &77.7\% &51.9\% &16.5\% &14.68 &26.75 &73.7\% &51.0\% &34.4\% \\
        Pose ViT~\cite{rockwell20228posevit}& 0.43 & 0.68 & 79.9\%  & 56.1\% & 18.7\% & 11.65 & 24.28  & 78.0\% & 59.7\% & 43.1\% \\
        \midrule
        SparsePlanes~\cite{jin2021sparseplanes}  & 0.56& 0.81& 73.7\% & 44.6\% & 10.7\% & 15.46& 33.38& 70.5\% & 48.7\% & 28.0\% \\
        PlaneFormers~\cite{agarwala2022planeformers} & 0.55 & 0.81 & 75.3\% & 45.5\% & 11.3\% & 14.34 & 32.08 & 73.2\% & 52.1\% & 32.3\% \\\midrule
        SparsePlanes-TRP~\cite{jin2021sparseplanes, Tan:etal:ICCV2021:Planetr} &  0.57 &  0.82 &  73.4\% &  43.6\% & 10.1\% & 14.57 & 32.36 & 72.8\% & 51.2\% & 30.1\% \\
        PlaneFormers-TRP~\cite{agarwala2022planeformers,Tan:etal:ICCV2021:Planetr} & 0.53 & 0.79 & 76.2\% & 47.0\% & 11.4\% & 13.81 & 31.58 & 74.5\% & 54.1\% & 33.6\% \\

        NOPE-SAC\jiajia{~\cite{tan2023nopesac}} & 0.41  & 0.65  & 82.1\% &  59.2\%  &  20.9\% & 8.29  & 22.30  &  82.4\%  &  73.0\%  &  59.2\% \\ 
        \midrule
        NOPE-SAC Ref.~\cite{tan2023nopesac}  & 0.57  & 0.81  & 75.4\% & 43.2\% & 7.3\%   &14.91 &37.05  & 66.8\%   & 50.3\% &  34.8\% \\
        PlaneRecTR++ (ours) & \textbf{0.24}  & \textbf{0.46}  & \textbf{88.6\%}  & \textbf{76.3\%}   &  \textbf{43.2\%} & \textbf{4.30}  & \textbf{17.16}  &  \textbf{87.6\%}  &  \textbf{84.1\%}  & \textbf{79.7\%} \\
        \midrule
        
        \multicolumn{10}{c}{MatterPort3D dataset} \\\midrule
        SuperGlue~\cite{sarlin2020superglue} & - & - & - & - & - & 3.88 & 24.17 & 77.8\% & 71.0\% & 65.7\%\\
        NOPE-SAC Init.~\cite{tan2023nopesac}& 0.69 & 1.08 & 65.0\% & 37.0\% & 10.1\% & 11.16 & 21.49 & 81.3\% & 60.5\% & 46.5\% \\
        Pose ViT~\cite{rockwell20228posevit}& 0.64 & 1.01 & 67.4\% & 39.9\% & 11.6\% & 8.01 & 19.13 & 85.4\% & 70.8\% & 57.8\% \\
        \midrule
        SparsePlanes~\cite{jin2021sparseplanes} & 0.63& 1.15 & 66.6\% & 40.4\% & 11.9\% & 7.33 & 22.78& 83.4\% & 72.9\% & 61.2\% \\
        PlaneFormers~\cite{agarwala2022planeformers} & 0.66 & 1.19 & 66.8\% & 36.7\% & 8.7\% & 5.96 & 22.20 & 83.8\% & 77.6\% & 68.0\% \\\midrule
        SparsePlanes-TRP~\cite{jin2021sparseplanes,Tan:etal:ICCV2021:Planetr} & 0.61 & 1.13 & 67.3\% & 41.7\% & 12.2\% & 6.87 & 22.17 & 83.8\% & 74.5\% & 63.3\% \\
        PlaneFormers-TRP~\cite{agarwala2022planeformers,Tan:etal:ICCV2021:Planetr} & 0.64 & 1.17 & 67.9\% & 38.7\% & 8.9\% & 5.28 & 21.90 & 83.9\% & 79.0\% & 70.8\% \\
        NOPE-SAC~\cite{tan2023nopesac} & 0.52  &  0.94  &  73.2\%  &   48.3\%  &   16.2\% & 2.77   &   \textbf{14.37}  &  \textbf{89.0\%}   &  \textbf{86.9\%}  &  \textbf{84.0\%} \\
        \midrule
        NOPE-SAC Ref.~\cite{tan2023nopesac}  &   1.53  & 1.92  &31.2\% & 11.8\%  & 2.5\%  & 3.88 &    27.95  & 76.8\%   &74.1\% & 70.9\%\\
        PlaneRecTR++ (ours) & \textbf{0.39}  & \textbf{0.86}  & \textbf{77.6\%}  & \textbf{58.5\%}   &  \textbf{24.3\%} & \textbf{2.60}  & 21.19  &  84.6\%  &  81.2\%  & 78.2\% \\

        \bottomrule
    \end{tabular}
}
\vspace{-5pt}
\label{tab:pose_comparison}
\end{table*}

%% file: tables/2views_planerec_ap.tex
\begin{table*}[!t]
\centering
\caption{Average Precision (AP) of 3D plane reconstruction given mask IoU ($\geq$0.5), normal angle error, and offset distance error. `All' means we consider all three conditions. `-Offset' and `-Normal' mean we ignore the offset and the normal conditions respectively.}
\vspace{-2mm}
\resizebox{0.8\linewidth}{!}{ 
    \begin{tabular}{l|ccc|ccc|ccc}
    \toprule
        \multicolumn{1}{c|}{\multirow{2}{*}{Method}} & \multicolumn{3}{c|}{Offset$\le$1m,~Normal$\le 30^{\circ}$} & \multicolumn{3}{c|}{Offset$\le$0.5m,~Normal$\le 15^{\circ}$} & \multicolumn{3}{c}{Offset$\le$0.2m,~Normal$\le 5^{\circ}$} \\
        & All & -Offset & -Normal & All & -Offset & -Normal & All & -Offset & -Normal \\\midrule
        \multicolumn{10}{c}{ScanNetv2 dataset} \\\midrule
        SparsePlanes~\cite{jin2021sparseplanes}  & 33.08& 34.12& 40.51& 21.69& 25.59& 32.20& 2.52&  4.50& 14.85\\
        PlaneFormers~\cite{agarwala2022planeformers} & 34.64& 35.47& 41.37& 24.48& 27.19& 34.69& 3.93& 5.52& 18.58\\\midrule
        SparsePlanes-TRP~\cite{jin2021sparseplanes,Tan:etal:ICCV2021:Planetr}  & 35.32& 36.50& 41.92& 24.71& 29.55& 33.50& 3.21& 6.07& 15.32\\
        PlaneFormers-TRP~\cite{agarwala2022planeformers,Tan:etal:ICCV2021:Planetr} & 36.82& 37.87& 43.01& 27.41& 30.72& 36.31& 4.83& 7.02& 19.94\\
        NOPE-SAC \cite{tan2023nopesac}& 39.61 & 40.45 & 44.04 & 31.39 & 35.07 & 38.05 & 6.76 & 10.12 & 21.50 \\
        \midrule
        PlaneRecTR++ (ours) & \textbf{51.08}  & \textbf{51.73} & \textbf{55.08} & \textbf{44.24} &  \textbf{46.20} & \textbf{51.25} & \textbf{18.19} & \textbf{21.68} &  \textbf{36.86}\\
        \midrule

        \multicolumn{10}{c}{MatterPort3D dataset} \\\midrule
        SparsePlanes~\cite{jin2021sparseplanes} & 36.02 & 42.01 & 39.04 & 23.53 & 35.25 & 27.64 & 6.76 & 17.18 & 11.52\\
        PlaneFormers~\cite{agarwala2022planeformers} & 37.62 & 43.19 & 40.36 & 26.10 & 36.88 & 29.99 & 9.44 & 18.82 & 14.78 \\\midrule
        SparsePlanes-TRP~\cite{jin2021sparseplanes,Tan:etal:ICCV2021:Planetr} & 40.35 & 46.39 & 43.03 & 27.81 & 40.65 & 31.38 & 9.02 & 22.80 & 13.66 \\
        PlaneFormers-TRP~\cite{agarwala2022planeformers,Tan:etal:ICCV2021:Planetr} & 41.87 & 47.50 & 44.43 & 30.78 & 42.82 & 34.03 & 12.45 & 25.98 & 17.34 \\
        NOPE-SAC~\cite{tan2023nopesac} &  43.29 & 49.00 & 45.32 & 32.61 & \textbf{44.94} & 35.36 & \textbf{14.25} & \textbf{30.39} & 18.37 \\
        \midrule
        PlaneRecTR++ (ours) & \textbf{45.22}  &  \textbf{49.71}  & \textbf{48.23} & \textbf{34.66} & 43.90 &  \textbf{38.85} & 13.70 & 26.03  & \textbf{19.69} \\
        
        \bottomrule
    \end{tabular}
}
\vspace{-5pt}
\label{tab:3DPlane}
\end{table*}

%% file: figures/2view_rec.tex
\begin{figure*}[htb!]
\centering
\centering
    \subfigure[Image 1]{
    \begin{minipage}[b]{0.13\linewidth}

        \includegraphics[height=50pt]{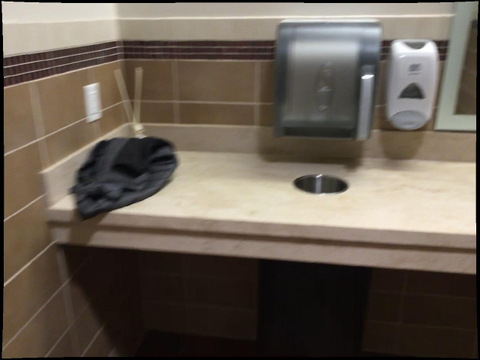}\vspace{2pt}
        \includegraphics[height=50pt]{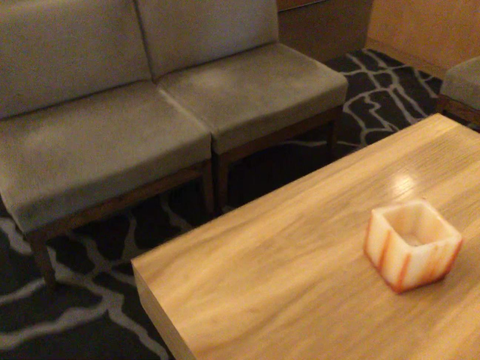}\vspace{2pt}
        \includegraphics[height=50pt]{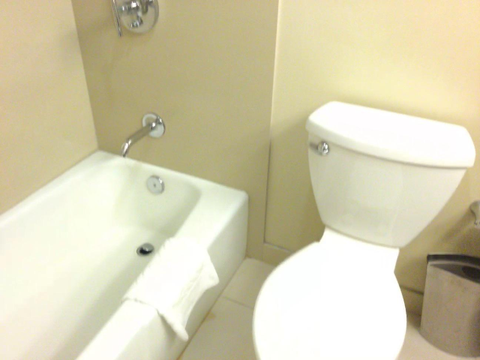}\vspace{2pt}

        \includegraphics[height=50pt]{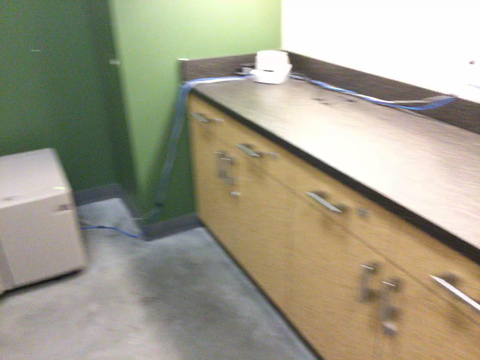}\vspace{2pt}
        \includegraphics[height=50pt]{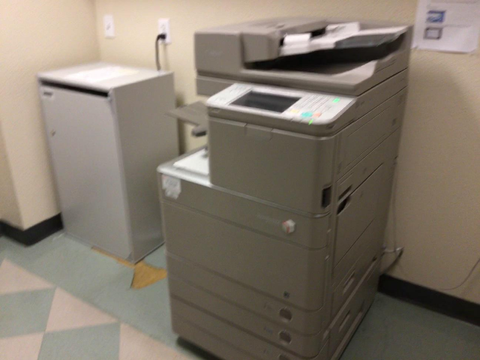}\vspace{2pt}

         \includegraphics[height=50pt]{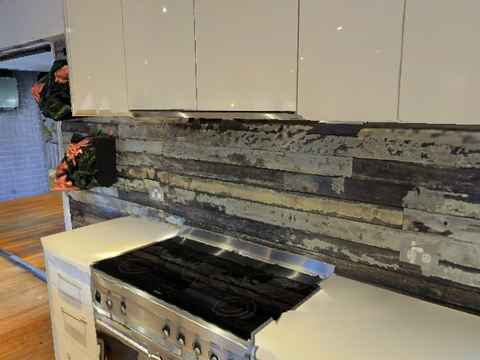}\vspace{2pt}
        \includegraphics[height=50pt]{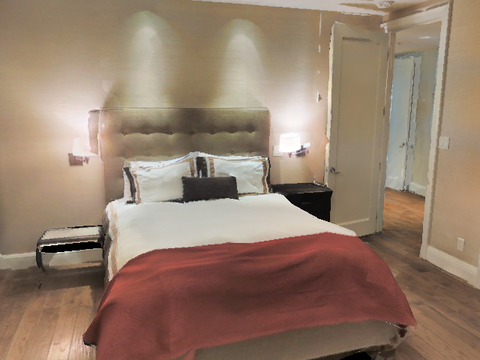}\vspace{2pt}

        \includegraphics[height=50pt]{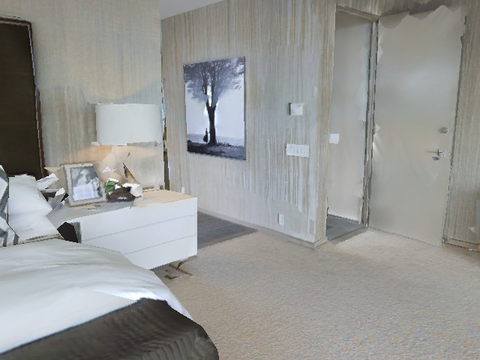}\vspace{2pt}
        \includegraphics[height=50pt]{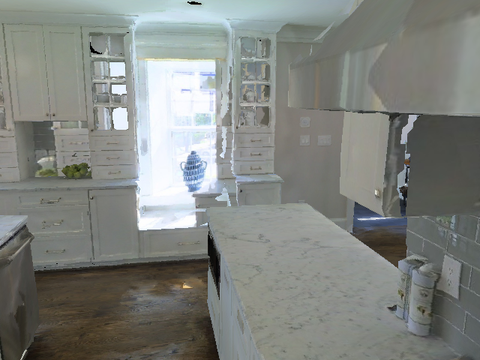}\vspace{2pt}
        \includegraphics[height=50pt]{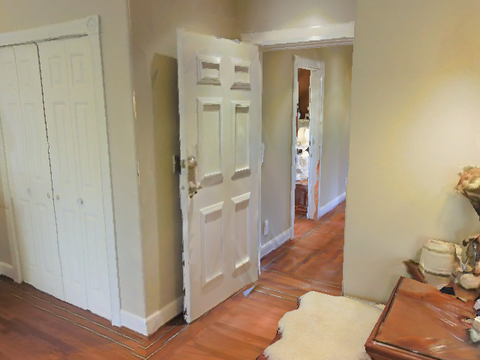}\vspace{2pt}
    \end{minipage}}
    \subfigure[Image 2]{
    \begin{minipage}[b]{0.13\linewidth}
        \includegraphics[height=50pt]{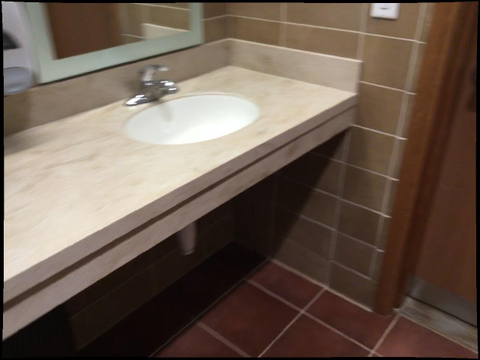}\vspace{2pt}
        \includegraphics[height=50pt]{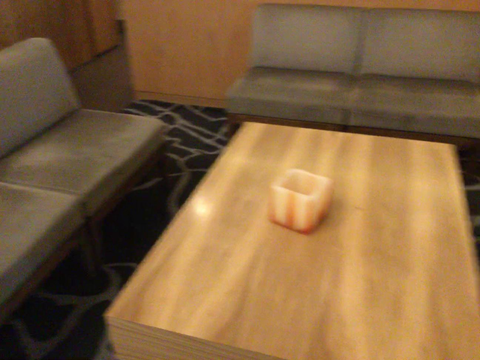}\vspace{2pt}
        \includegraphics[height=50pt]{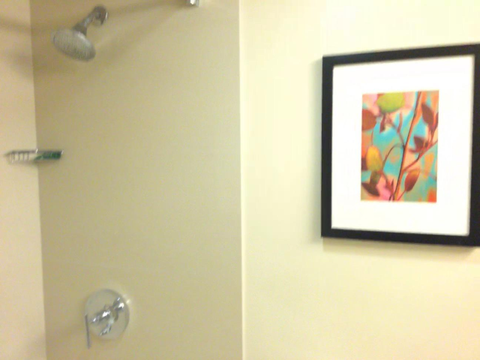}\vspace{2pt}

        \includegraphics[height=50pt]{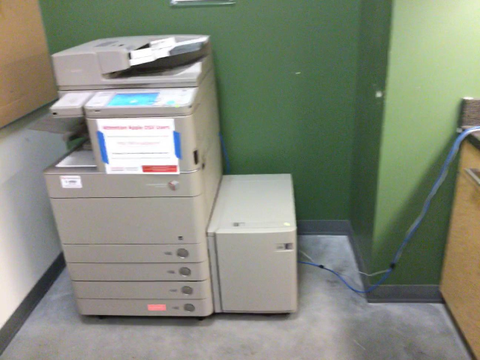}\vspace{2pt}
        \includegraphics[height=50pt]{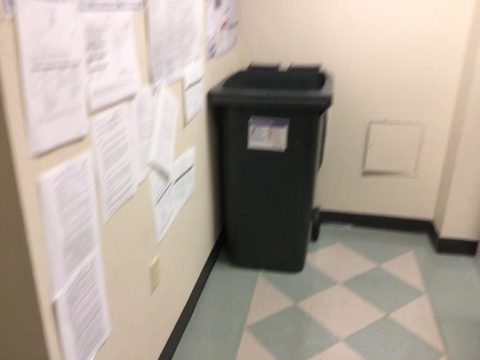}\vspace{2pt}

        \includegraphics[height=50pt]{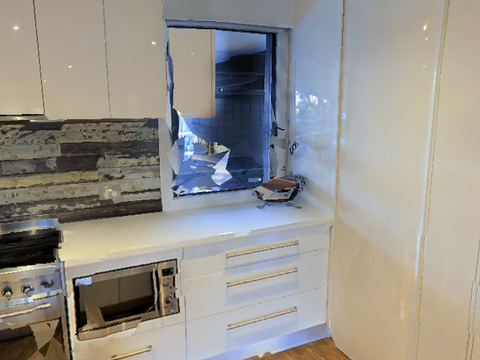}\vspace{2pt}
        \includegraphics[height=50pt]{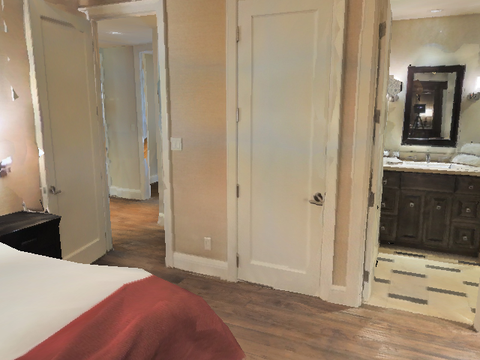}\vspace{2pt}

        \includegraphics[height=50pt]{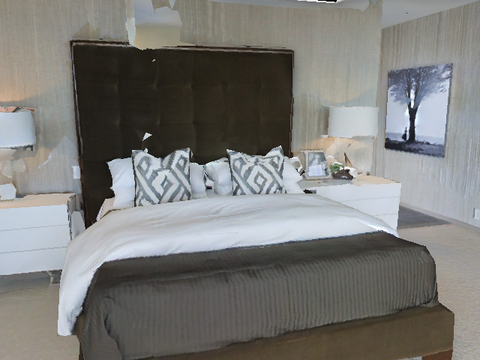}\vspace{2pt}
        \includegraphics[height=50pt]{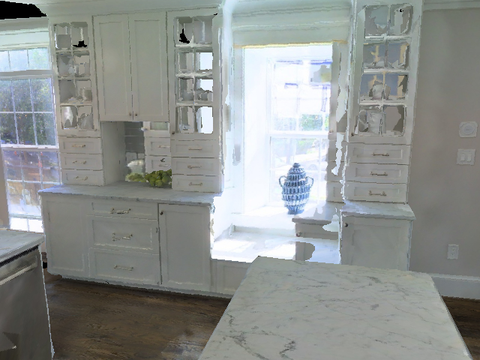}\vspace{2pt}
        \includegraphics[height=50pt]{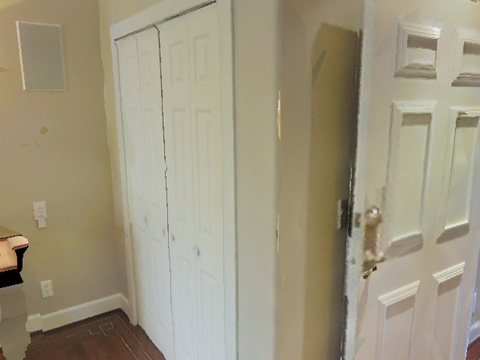}\vspace{2pt}
    \end{minipage}}
    \subfigure[Ours Plane Correspondences]{
    \begin{minipage}[b]{0.275\linewidth}
        \includegraphics[height=50pt]{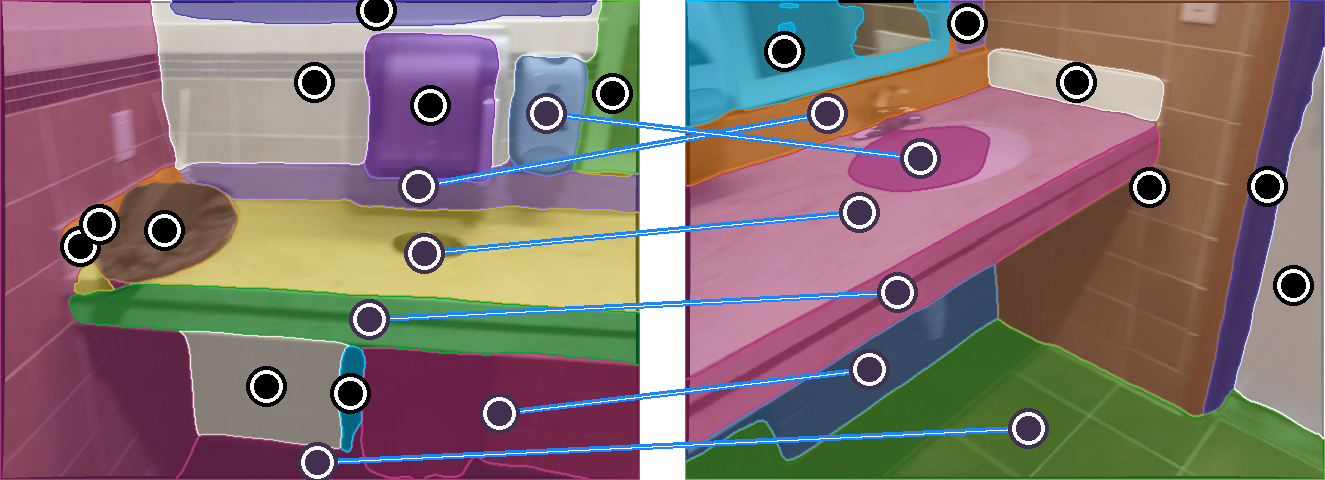}\vspace{2pt}
        \includegraphics[height=50pt]{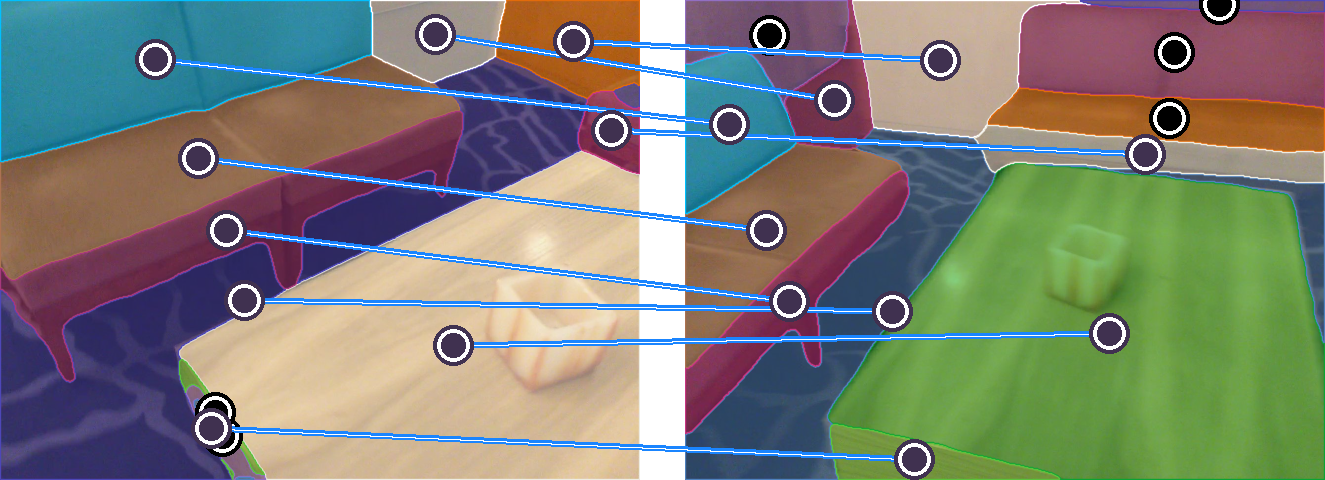}\vspace{2pt}
        \includegraphics[height=50pt]{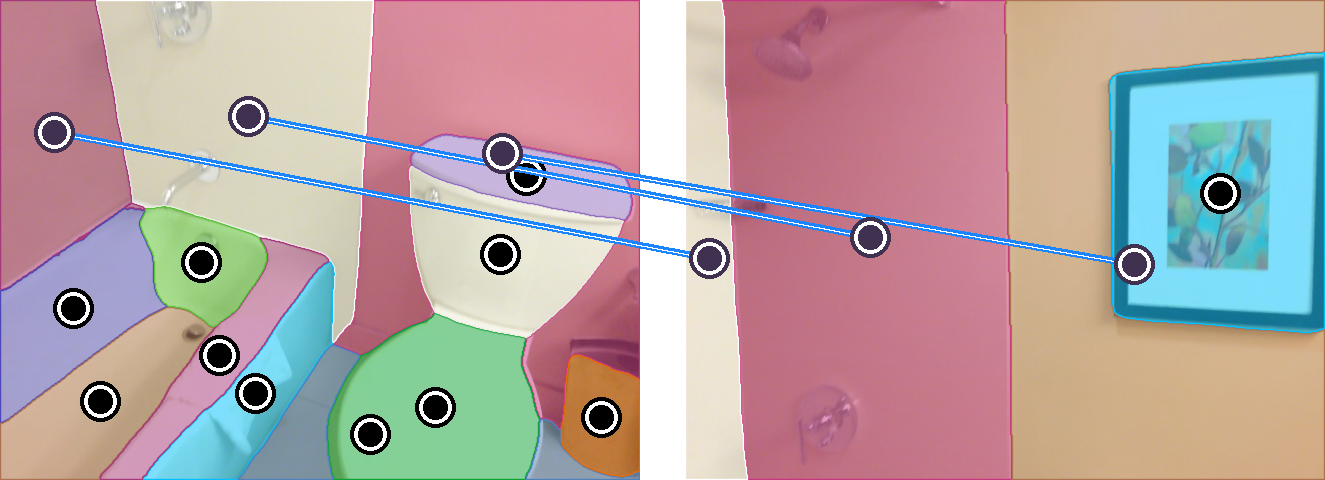}\vspace{2pt}

        \includegraphics[height=50pt]{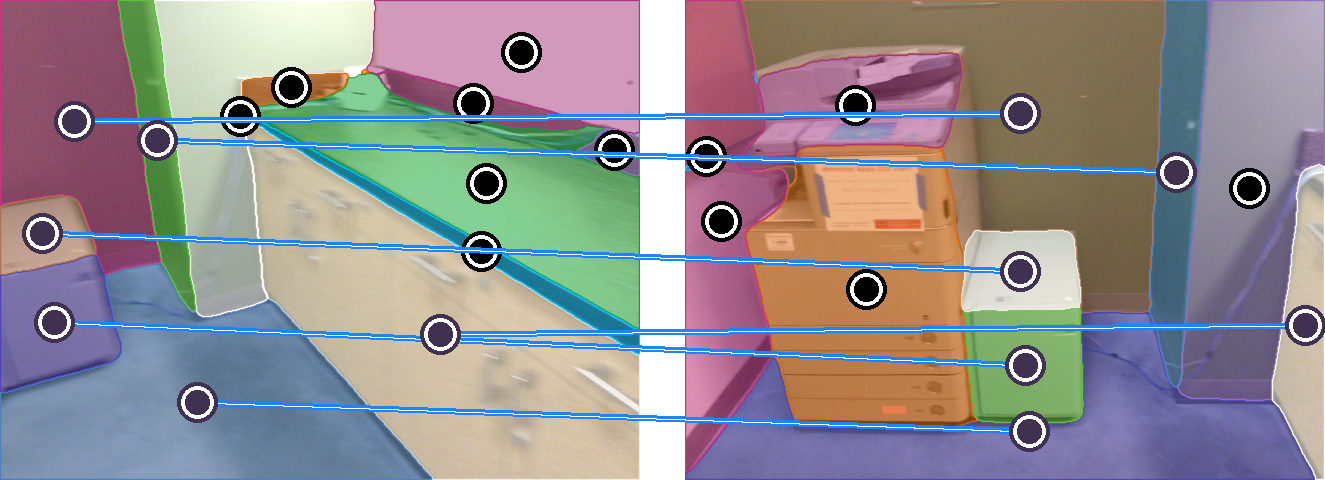}\vspace{2pt}
        \includegraphics[height=50pt]{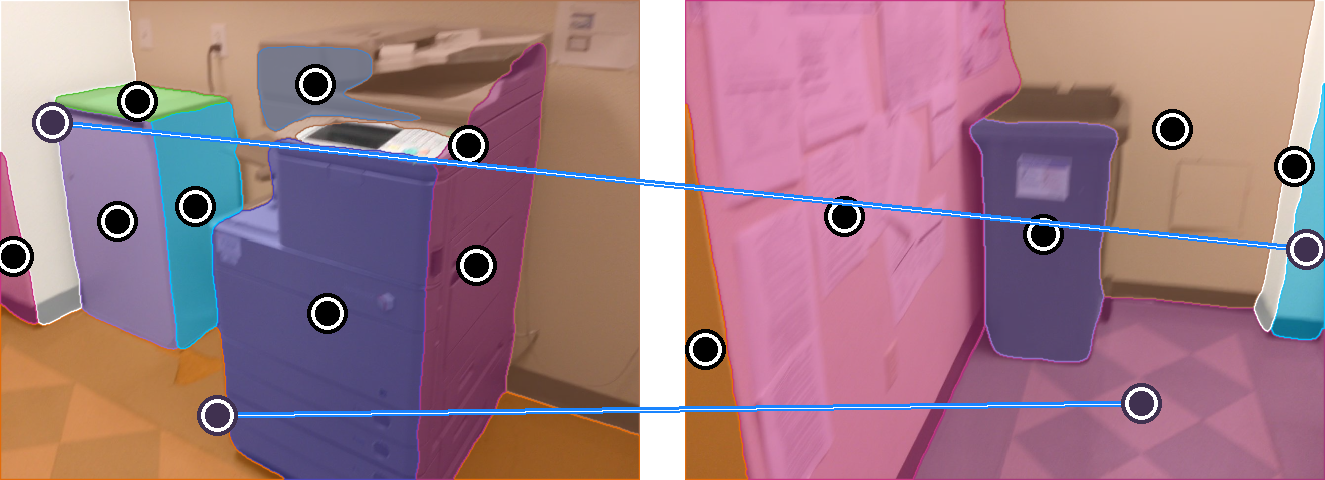}\vspace{2pt}

        \includegraphics[height=50pt]{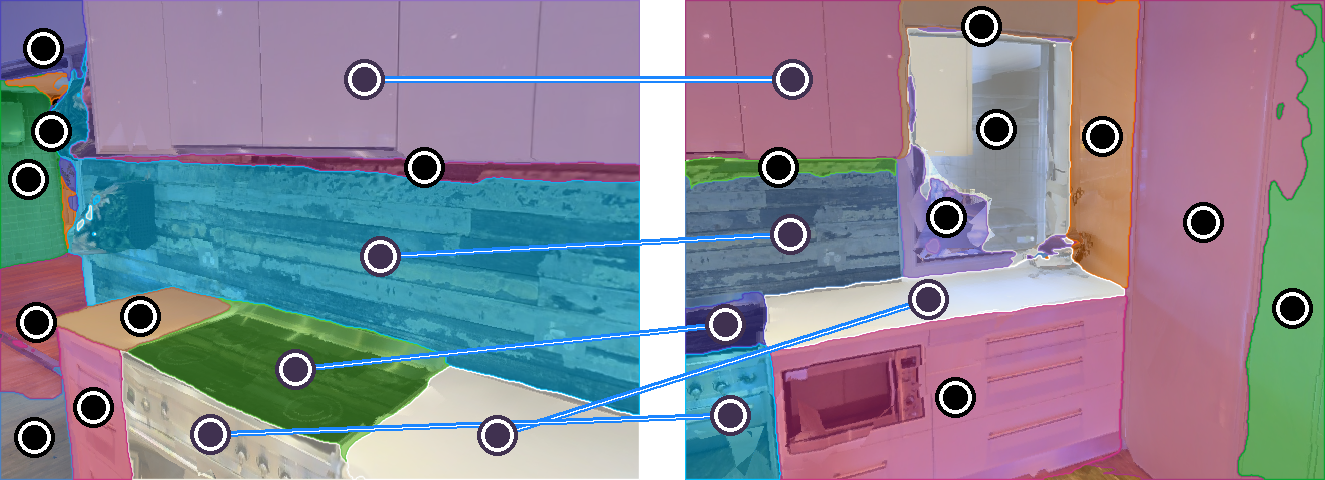}\vspace{2pt}
        \includegraphics[height=50pt]{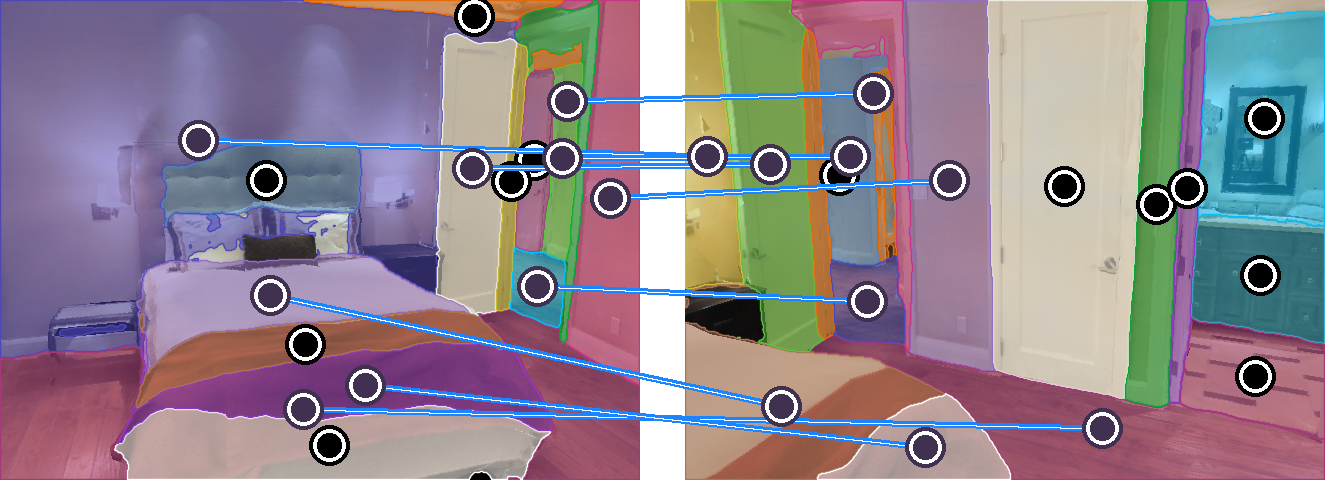}\vspace{2pt}

        \includegraphics[height=50pt]{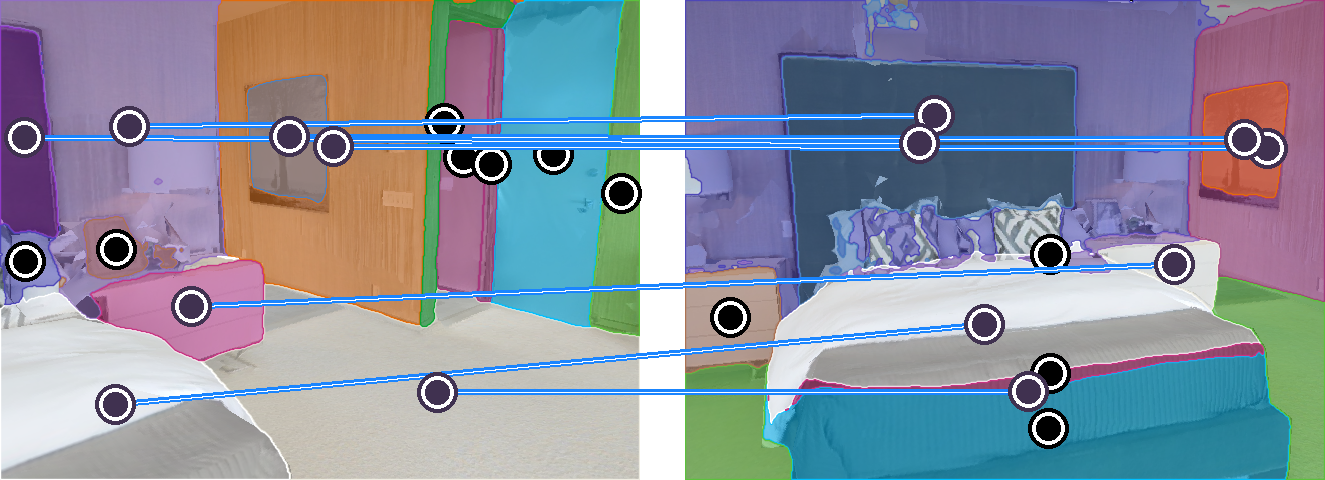}\vspace{2pt}
        \includegraphics[height=50pt]{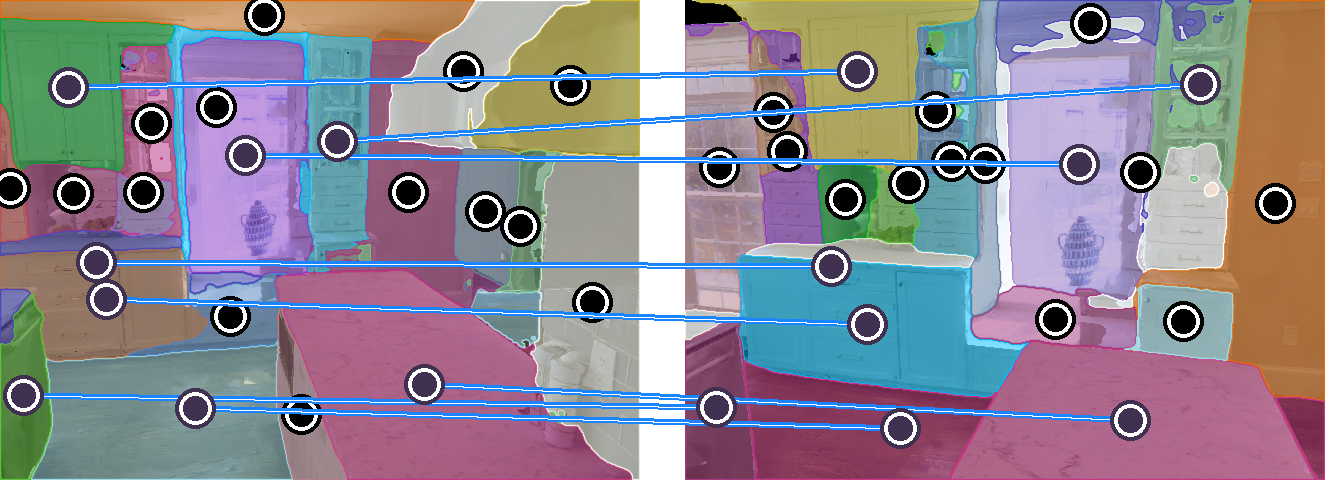}\vspace{2pt}
        \includegraphics[height=50pt]{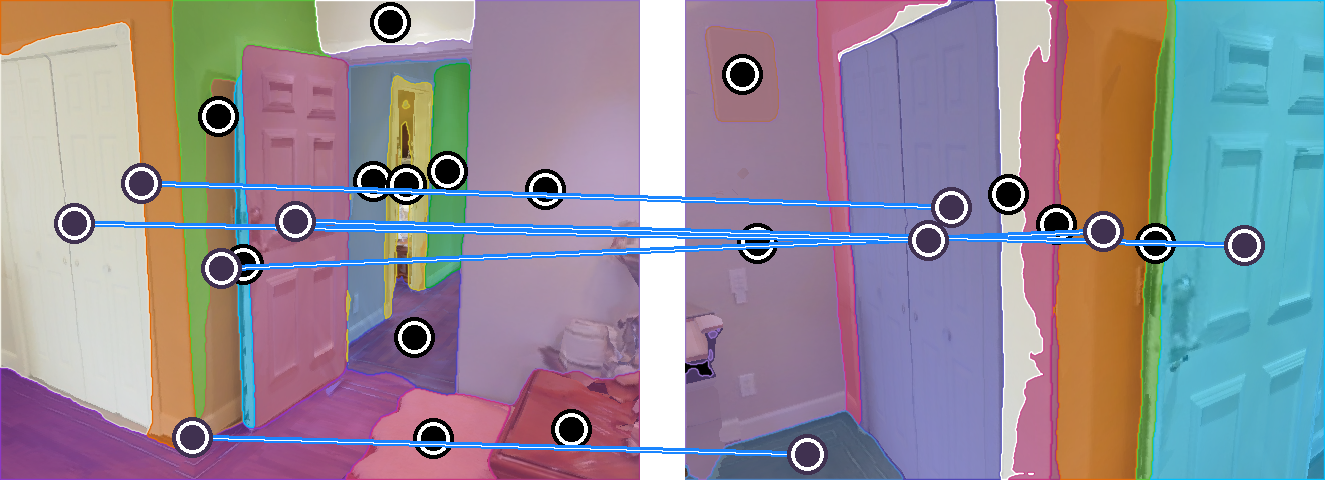}\vspace{2pt}
        
    \end{minipage}}
    \subfigure[Ours]{
    \begin{minipage}[b]{0.17\linewidth}
        \includegraphics[height=50pt]{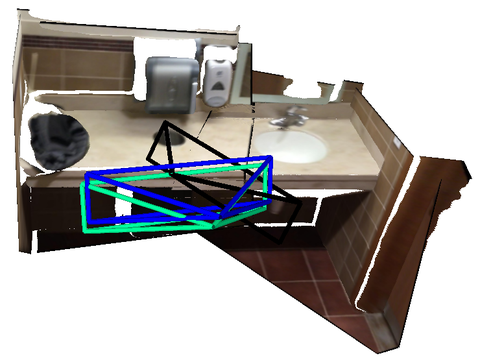}\vspace{2pt}
        \includegraphics[height=50pt]{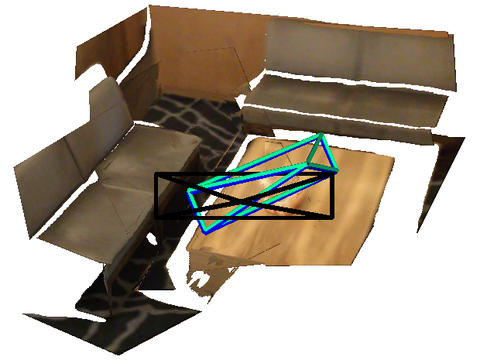}\vspace{2pt}
        \includegraphics[height=50pt]{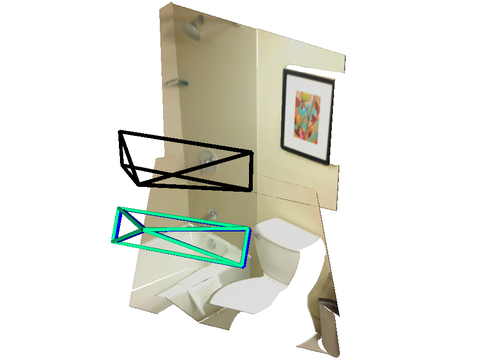}\vspace{2pt}

        \includegraphics[height=50pt]{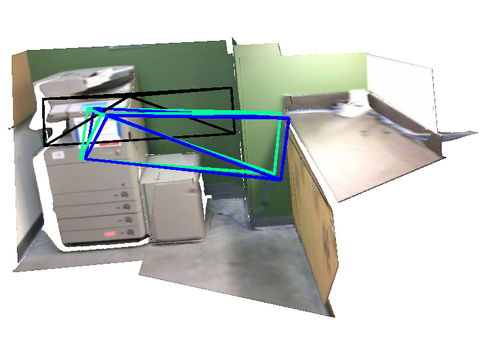}\vspace{2pt}
        \includegraphics[height=50pt]{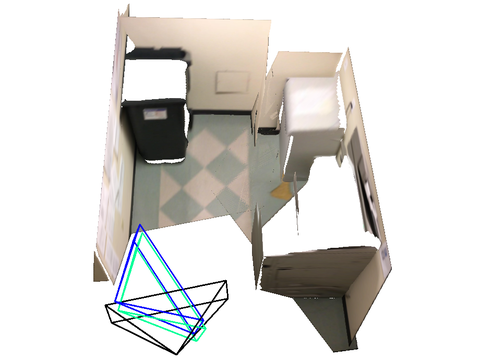}\vspace{2pt}

         \includegraphics[height=50pt]{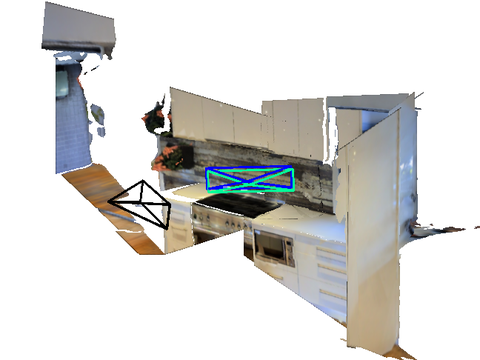}\vspace{2pt}
        \includegraphics[height=50pt]{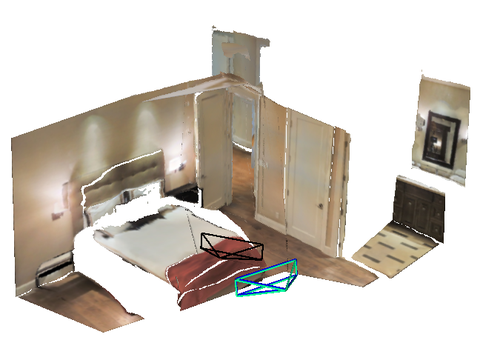}\vspace{2pt}

        \includegraphics[height=50pt]{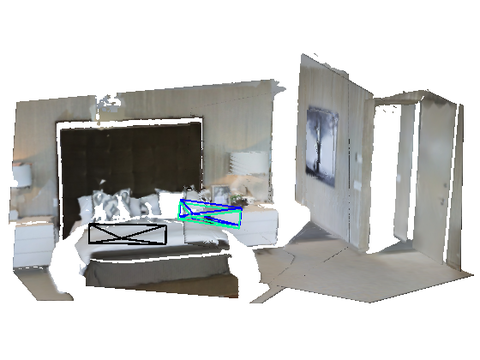}\vspace{2pt}
        \includegraphics[height=50pt]{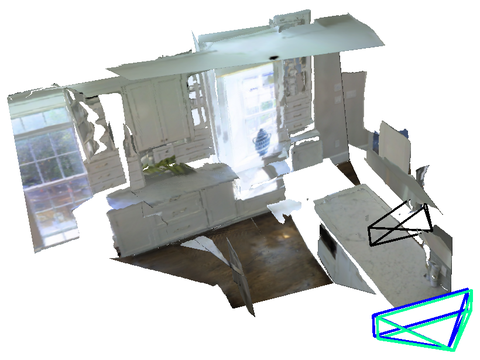}\vspace{2pt}
        \includegraphics[height=50pt]{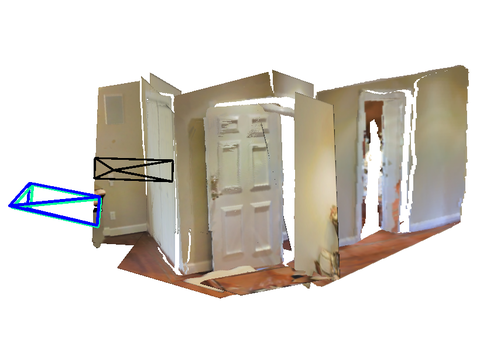}\vspace{2pt}
    \end{minipage}}
    \subfigure[GT]{
    \begin{minipage}[b]{0.17\linewidth}
        \includegraphics[height=50pt]{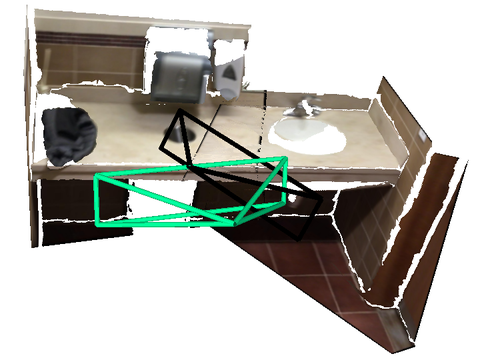}\vspace{2pt}
        \includegraphics[height=50pt]{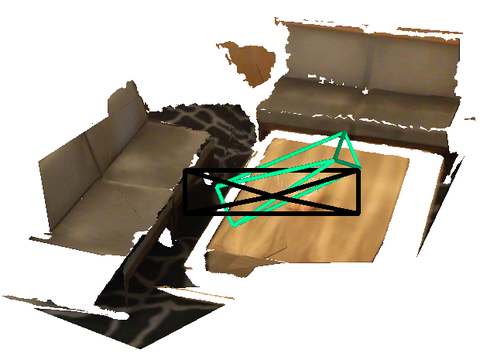}\vspace{2pt}
        \includegraphics[height=50pt]{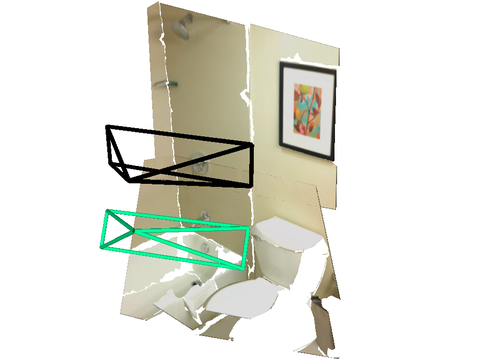}\vspace{2pt}

        \includegraphics[height=50pt]{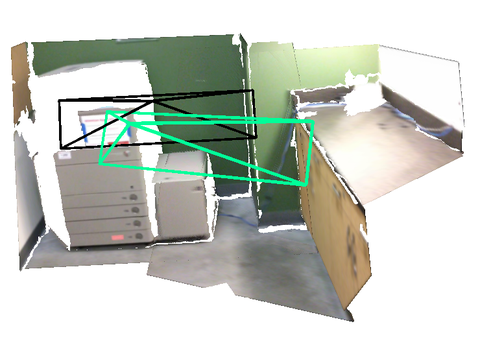}\vspace{2pt}
        \includegraphics[height=50pt]{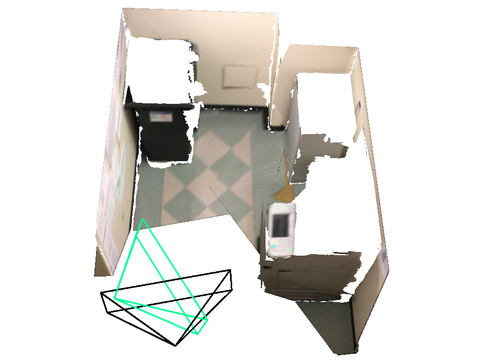}\vspace{2pt}

        \includegraphics[height=50pt]{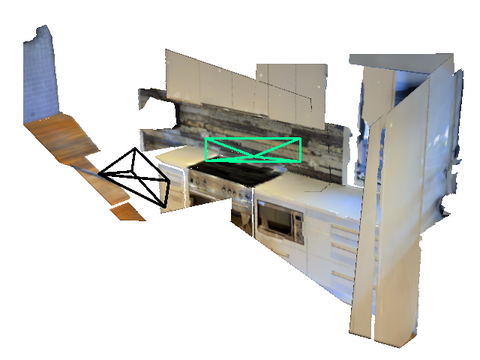}\vspace{2pt}
        \includegraphics[height=50pt]{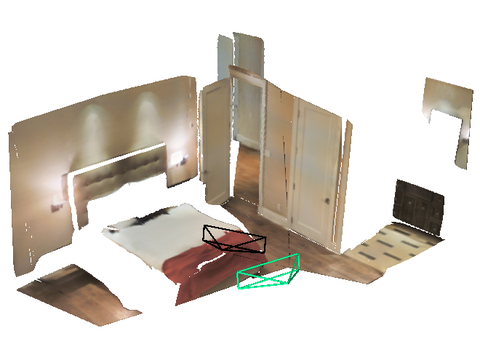}\vspace{2pt}

        \includegraphics[height=50pt]{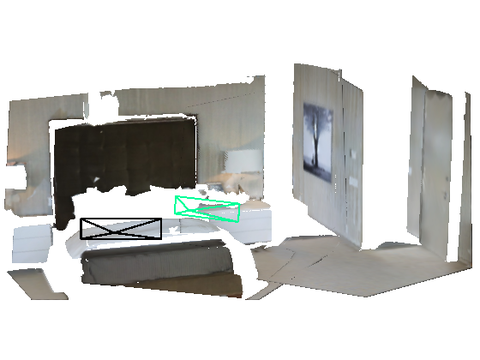}\vspace{2pt}
        \includegraphics[height=50pt]{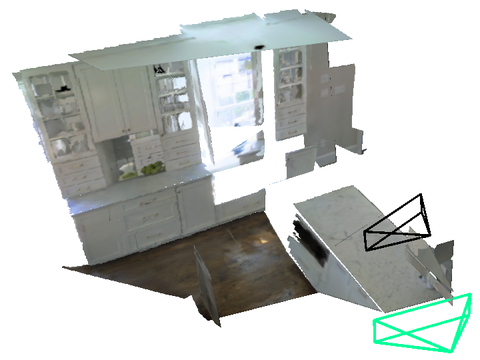}\vspace{2pt}
        \includegraphics[height=50pt]{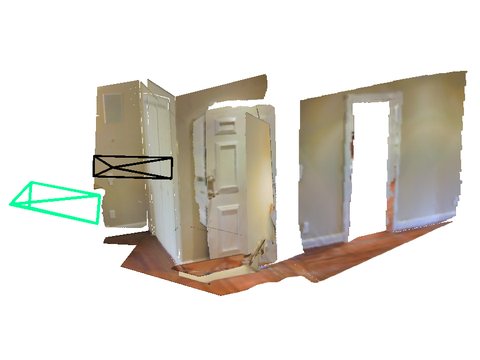}\vspace{2pt}
    \end{minipage}}
\vspace{-2pt}
\caption{Qualitative results of the plane correspondences, relative camera pose and 3d reconstructions on the ScanNetv2~\cite{Dai:etal:CVPR2017,Dai:scannetv2web} dataset (first 5 rows) and the MatterPort3D~\cite{chang2017matterport3d} dataset (last 5 rows). The \textbf{Green} frustums represent the ground truth camera of the first image while the \textbf{blue} frustums depict our predicted results. The fixed \textbf{Black} frustums show the camera of the second image.}
\vspace{-8pt}
\label{fig:2view_rec}
\end{figure*}

%% file: figures/2view_rec_compare.tex
\begin{figure*}
\centering
\renewcommand\tabcolsep{15pt}
\begin{tabular}{rccccccc}
\raisebox{20pt}{\rotatebox[origin=c]{90}{Image 1}}
\includegraphics[height=50pt]{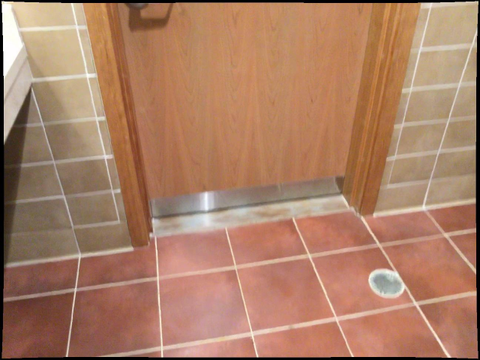} 
\includegraphics[height=50pt]{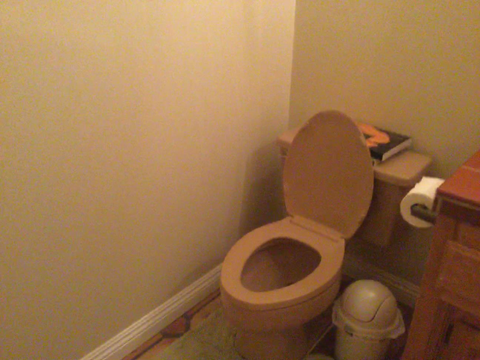} 
\includegraphics[height=50pt]{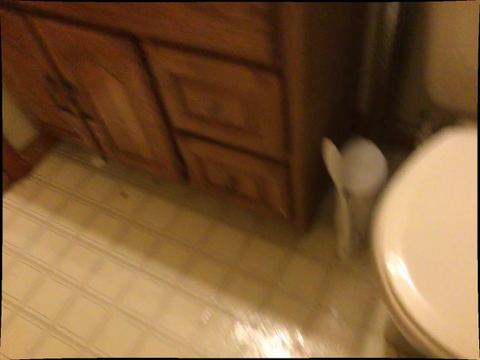}
\includegraphics[height=50pt]{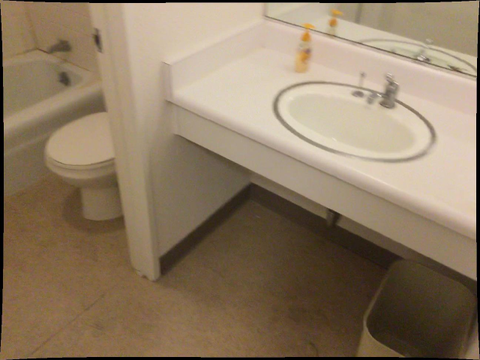}
\includegraphics[height=50pt]{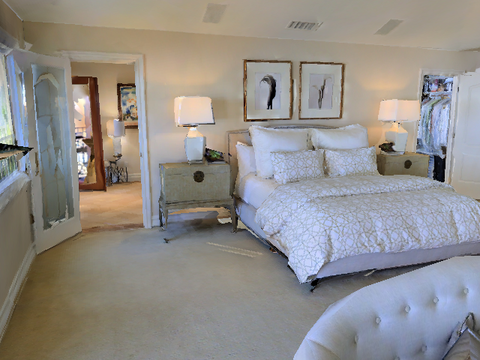}
\includegraphics[height=50pt]{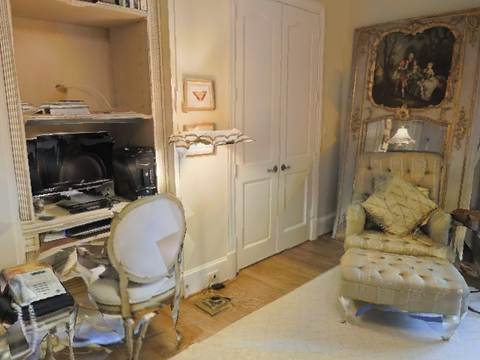}
\includegraphics[height=50pt]{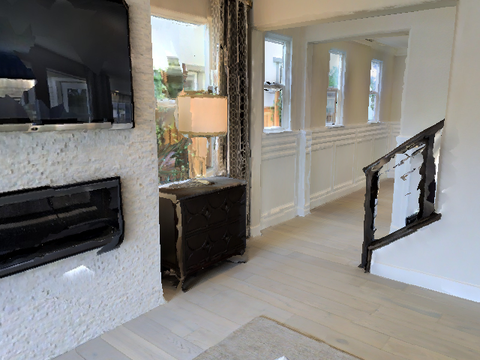}\\
\raisebox{20pt}{\rotatebox[origin=c]{90}{Image 2}}
        \includegraphics[height=50pt]{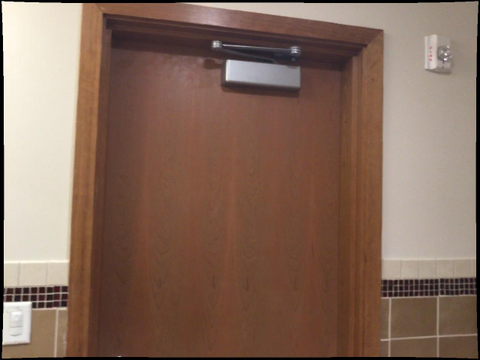}
        \includegraphics[height=50pt]{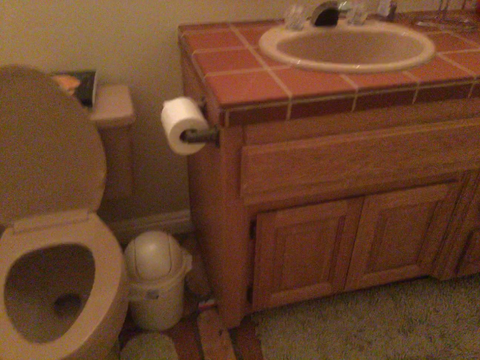}
        \includegraphics[height=50pt]{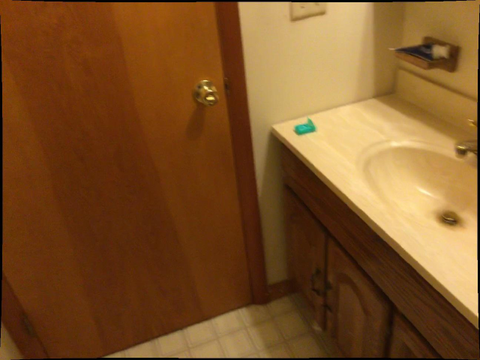}
        \includegraphics[height=50pt]{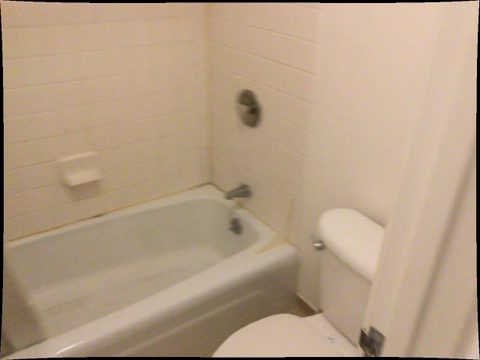}
        \includegraphics[height=50pt]{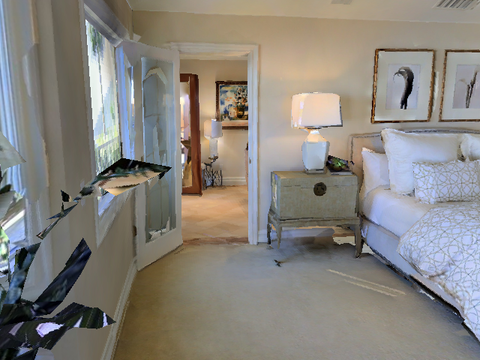}
        \includegraphics[height=50pt]{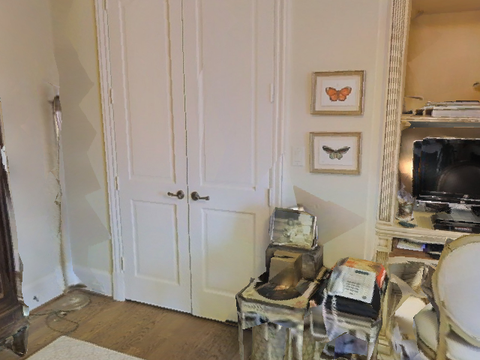}
        \includegraphics[height=50pt]{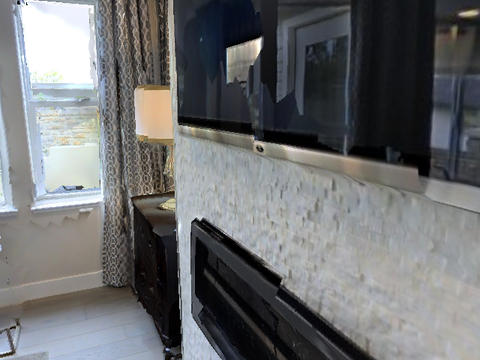}\\
\raisebox{20pt}{\rotatebox[origin=c]{90}{NOPE-SAC}}
        \includegraphics[height=50pt]{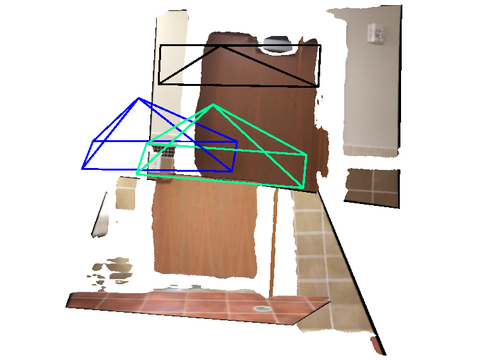}
        \includegraphics[height=50pt]{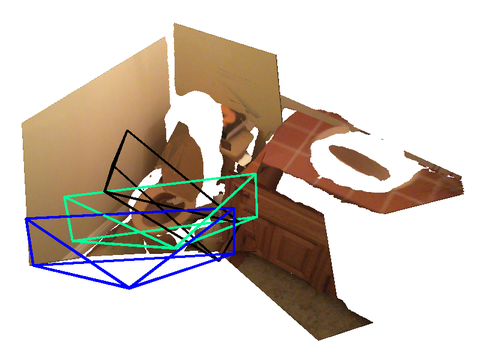}
        \includegraphics[height=50pt]{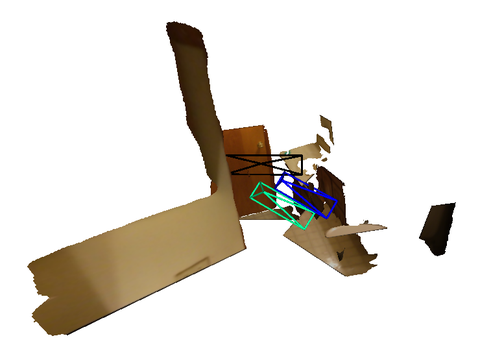}
        \includegraphics[height=50pt]{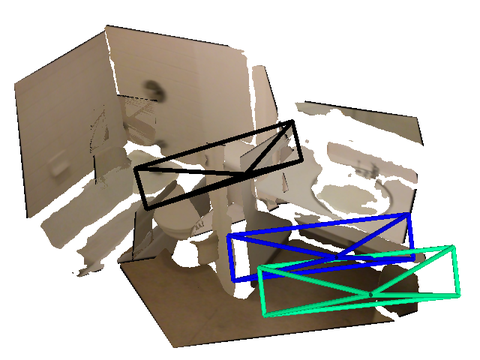}
        \includegraphics[height=50pt]{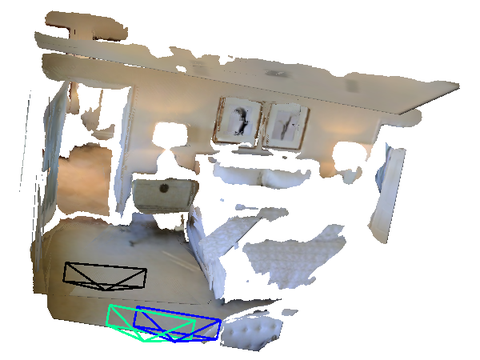}
        \includegraphics[height=50pt]{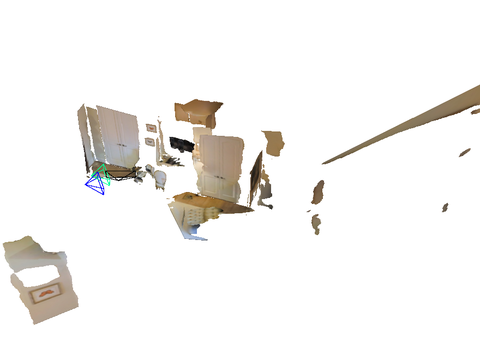}
        \includegraphics[height=50pt]{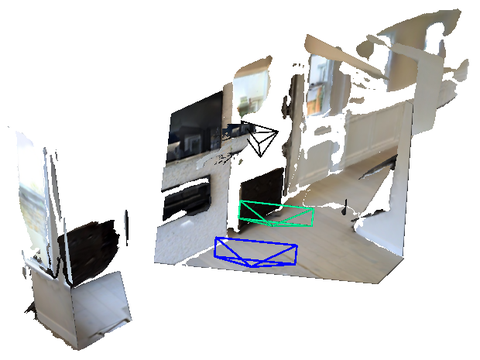}\\
\raisebox{20pt}{\rotatebox[origin=c]{90}{Ours}}
        \includegraphics[height=50pt]{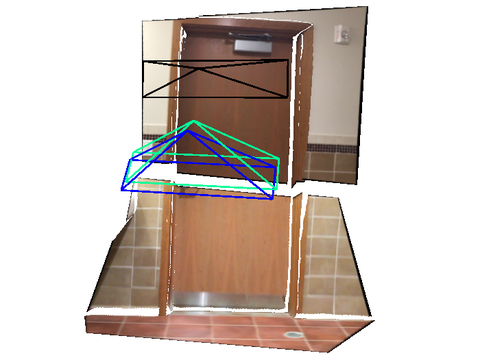}
        \includegraphics[height=50pt]{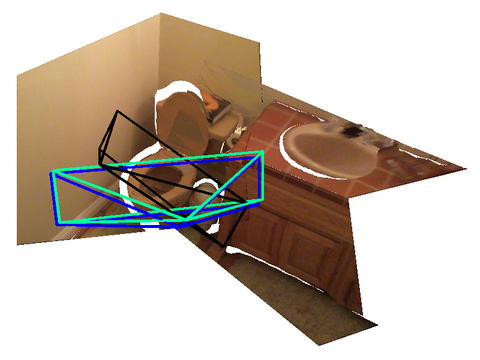}
        \includegraphics[height=50pt]{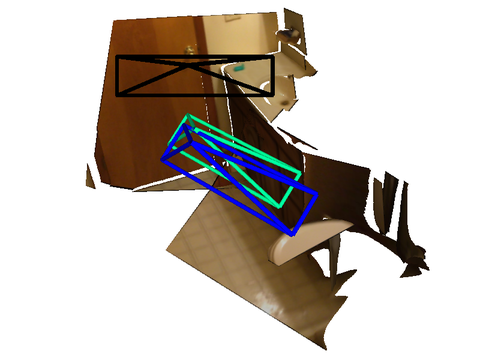}
        \includegraphics[height=50pt]{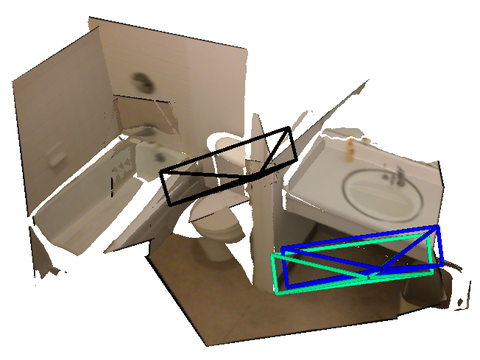}
        \includegraphics[height=50pt]{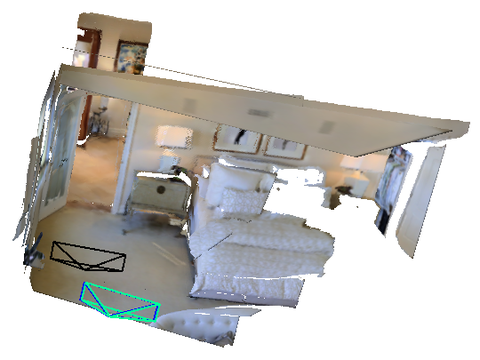}
        \includegraphics[height=50pt]{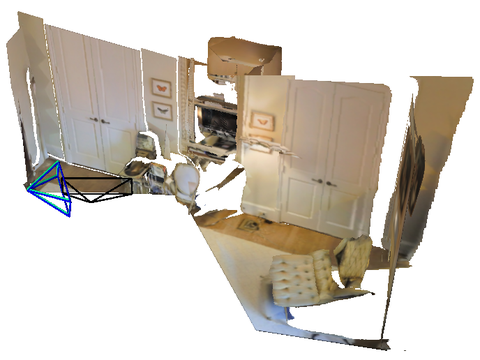}
        \includegraphics[height=50pt]{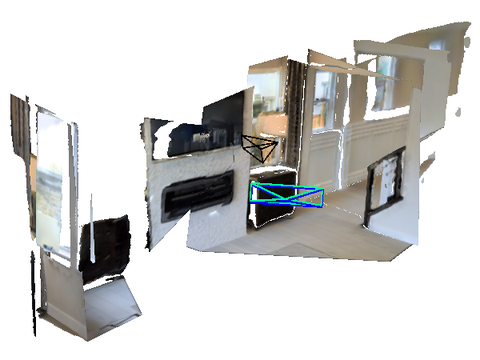}\\
\raisebox{20pt}{\rotatebox[origin=c]{90}{GT}}
        \includegraphics[height=50pt]{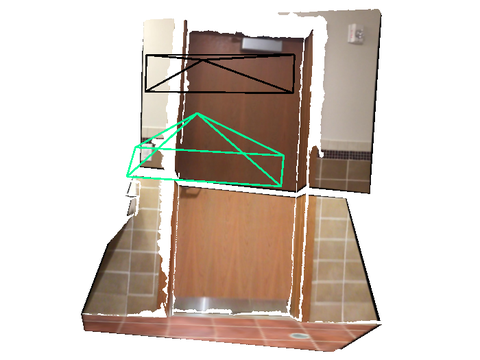}
        \includegraphics[height=50pt]{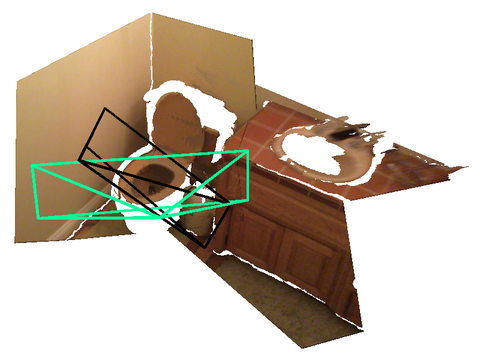}
        \includegraphics[height=50pt]{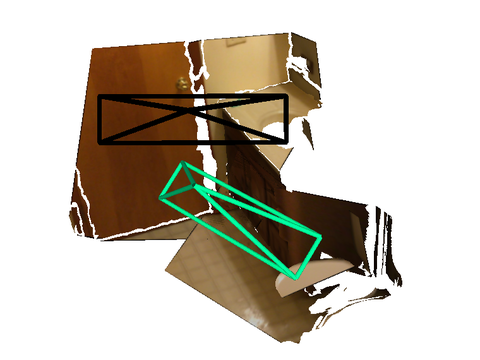}
        \includegraphics[height=50pt]{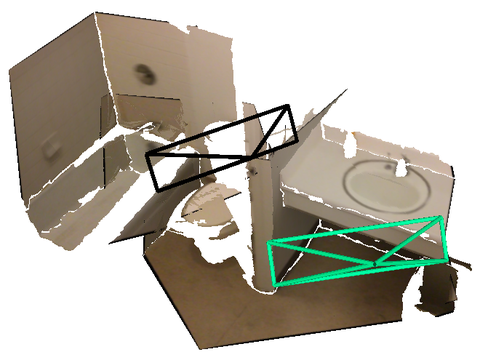}
        \includegraphics[height=50pt]{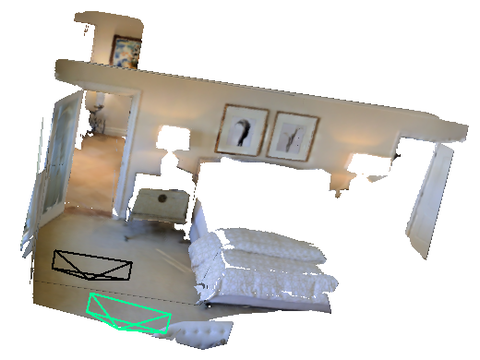}
        \includegraphics[height=50pt]{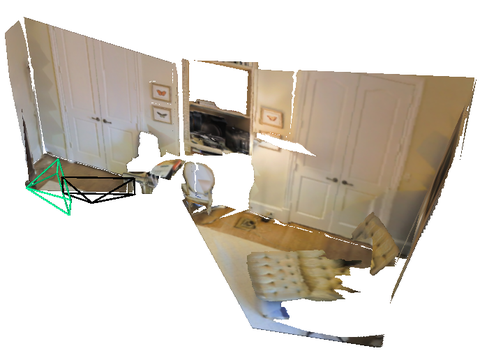}
        \includegraphics[height=50pt]{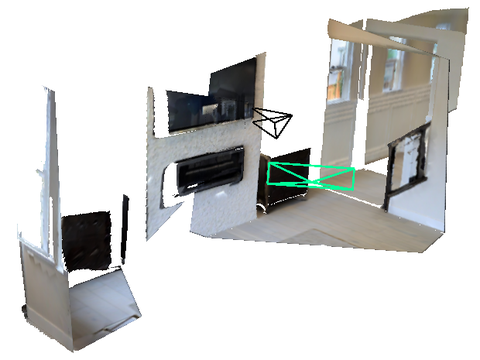}
\end{tabular}  
\vspace{-5pt}
\caption{3D planar reconstruction results on the ScanNetv2 (first 4 columns) and the MatterPort3D datasets (last 3 columns). \textbf{Green} and \textbf{Blue} frustums show the ground truth and predicted cameras of the first image. The fixed \textbf{Black} frustums show the camera of the second image.}
\vspace{-5pt}
\label{fig:2view_rec_compare}
\end{figure*}

%% file: tables/pose_model_ablation.tex
\begin{table*}[t!]
\centering
\caption{Ablation studies for the different model designs in pose estimation of PlaneRecTR++.}
\vspace{-5pt}
\resizebox{0.9\linewidth}{!}{ 
    \begin{tabular}{ccc|ccccc|ccccc}
    \toprule
        \multicolumn{3}{c|}{Settings} & \multicolumn{5}{c|}{Translation} & \multicolumn{5}{c}{Rotation} \\ 
        CE & QKNum. & VNum. & Med.~$\downarrow$ & Mean~$\downarrow$ & ($\le$1m)~$\uparrow$& 
        ($\le$0.5m)~$\uparrow$& ($\le$0.2m)~$\uparrow$ &
        Med.~$\downarrow$ & Mean~$\downarrow$ & ($\le 30^{\circ}$)~$\uparrow$ & ($\le 15^{\circ}$)~$\uparrow$ & ($\le 10^{\circ}$)~$\uparrow$ \\
        \midrule
        
        \multicolumn{13}{c}{ScanNetv2 dataset} \\
        \midrule
        
        
        $\checkmark$& 1 & 4 &\textbf{0.24}  & \textbf{0.46}  & \textbf{88.6\%}  & \textbf{76.3\%}   &   \textbf{43.2\%} &  \textbf{4.30}  & 17.16  &  \textbf{87.6\%}  &   \textbf{84.1\% } &  \textbf{79.7\%}\\
          & 1 & 4 &0.28 & 0.50 & 87.4\% & 72.3\% & 37.2\% & 5.24  &  17.09 &  85.9\%&  80.4\%&  73.0\% \\
        $\checkmark$& 4 & 4 &  0.25 & 0.47 & 88.3\% & 74.7\% & 40.2\% & 4.45  & \textbf{16.81} & 87.4\%  & 83.7\% & 78.4\% \\
        $\checkmark$& 1 & 1 &  0.27 & 0.49 & 87.9\% & 72.7 \% & 38.2 \% & 5.13  & 17.15  &  86.3\% & 80.9\% & 73.3\% \\
        \midrule

        \multicolumn{13}{c}{MatterPort3D dataset} \\
        \midrule
       $\checkmark$& 1 & 4 &  \textbf{0.39}  & \textbf{0.86}  & \textbf{77.6\%}  & \textbf{58.5\%}   &  \textbf{24.3\%} & \textbf{2.60}  & \textbf{21.19}  &  \textbf{84.6\%} &  \textbf{81.2\%}  & \textbf{78.2\%} \\
        & 1 & 4&  0.49 & 0.98 & 72.4\% & 50.7\% & 18.4\% & 4.39  &  25.72 &  80.0 \%&  74.0\%&  69.2\% \\
        $\checkmark$& 4 & 4  &  0.42 & 0.88 & 76.1\% & 56.5\% & 23.4\% & 3.06  & 21.90 & 83.6\%  & 79.6\% & 76.1\% \\
        $\checkmark$& 1 & 1  &  0.51 & 1.00 & 71.5\% & 49.5\% & 17.5\% & 4.70  & 26.06 & 79.4\% & 72.7\% & 67.1\% \\
         
        \bottomrule
    \end{tabular}
}
\label{tab:pose_model_ablation}
\vspace{-5pt}
\end{table*}

%% file: tables/corr_ablation.tex
\begin{table*}[t!]
\centering
\caption{Ablation studies for the different model designs in plane correspondences and 3D reconstruction of PlaneRecTR++.}
\vspace{-2mm}
\resizebox{0.8\linewidth}{!}{ 
    \begin{tabular}{ccc|cccc|ccc|ccc}
    \toprule
        \multicolumn{3}{c|}{Settings} & \multicolumn{4}{c|}{Correspondence} & \multicolumn{3}{c|}{Offset$\le$1m,~Normal$\le 30^{\circ}$} & \multicolumn{3}{c}{Offset$\le$0.2m,~Normal$\le 5^{\circ}$} \\ 
        CE & QKNum. & VNum. &Precision & Recall & F-score &TP  & All & -Offset & -Normal  & All & -Offset & -Normal \\
        \midrule
        
        \multicolumn{13}{c}{ScanNetv2 dataset} \\
        \midrule
        $\checkmark$& 1 & 4  & \textbf{0.576} & \textbf{0.552} & \textbf{0.564}  & \textbf{10192} & \textbf{51.08} &\textbf{51.73} & \textbf{55.08} &\textbf{18.19} & \textbf{21.68} &  \textbf{36.86} \\ 
          & 1 & 4  &0.540  &0.518 & 0.529 & 9572  & 48.48 & 49.07 & 53.03 & 15.32 &18.63  &33.50\\ 
        $\checkmark$& 4 & 4  & 0.566 & 0.547 & 0.556 & 10102 &50.43 &51.06 & 54.50 & 17.77 & 21.35 & 36.05\\ 
        $\checkmark$& 1 & 1 & 0.562  & 0.538 & 0.550  & 9938 & 49.96 &50.57 &54.62 & 15.50& 18.66& 34.57 
        \\
        \midrule

        \multicolumn{13}{c}{MatterPort3D dataset} \\
        \midrule
       $\checkmark$& 1 & 4 & \textbf{0.540} & \textbf{0.476} & \textbf{0.506} & \textbf{20630} & \textbf{45.22}  &  \textbf{49.71}  & \textbf{48.23} & \textbf{13.70} & \textbf{26.03}  & \textbf{19.69} \\ 
          & 1 & 4  & 0.518 & 0.460 & 0.487 & 19915 & 42.26 & 46.85 & 45.86 & 10.60 &21.04 &16.98
  \\ 
        $\checkmark$& 4 & 4 & 0.530 & 0.469 & 0.498 & 20302 & 44.53 & 48.69 & 47.63 & 13.32 & 24.95 &19.51 \\ 
        $\checkmark$& 1 & 1  &  0.503& 0.447 & 0.473 & 19358 & 43.01 & 47.46 & 46.56 &12.18 &22.20 & 18.45
\\ 
         
        \bottomrule
    \end{tabular}
}
\label{tab:corr_ablation}
\vspace{-2mm}
\end{table*}

%% file: figures/heatmap.tex
\begin{figure}[!ht]
\centering
\subfigure[QKNum.=1]{
    \begin{minipage}[b]{0.27\linewidth}
        \includegraphics[height=145pt]{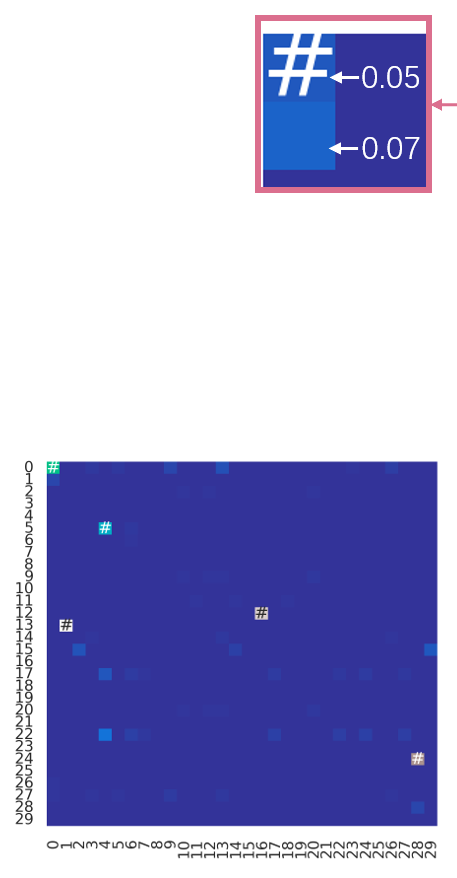}
    \label{subfig:qk1v4}
    \end{minipage}
}
\subfigure[QKNum.=4]{
    \hspace{-5pt}
    \begin{minipage}[b]{0.65\linewidth}
        \includegraphics[height=145pt]
        {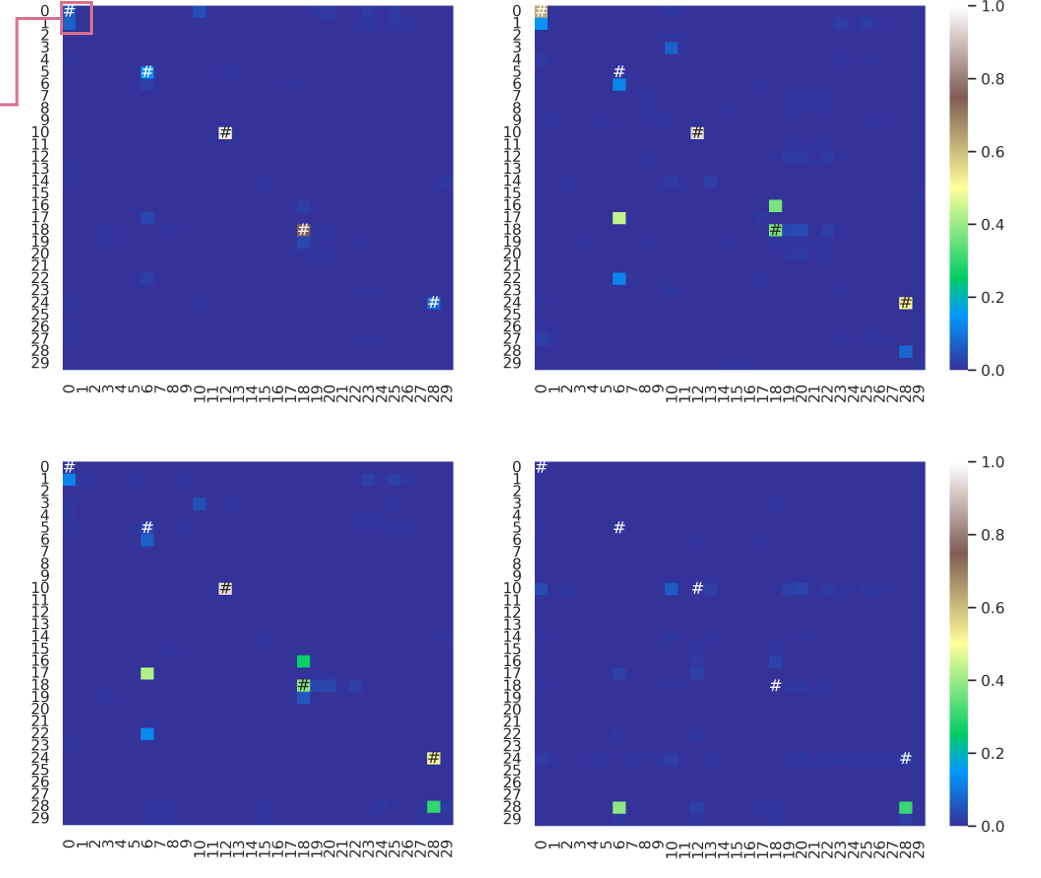}
    \label{subfig:qkv4}
    \end{minipage}
}
\caption{Visualization of attention matrices for unsplit/split query and key designs. "\#" marks the ground truth correspondence via bipartite matching in Section \ref{sec:training}. Please zoom in for more visual details.}
\vspace{-7pt}
\end{figure}



%% file: figures/multiview_3dres.tex
\begin{figure*}
\centering
\renewcommand\tabcolsep{15pt}
\begin{tabular}{rccccccc}
\raisebox{72pt}{\rotatebox[origin=c]{90}{\jiajia{Images}}} 
\includegraphics[width=65pt]{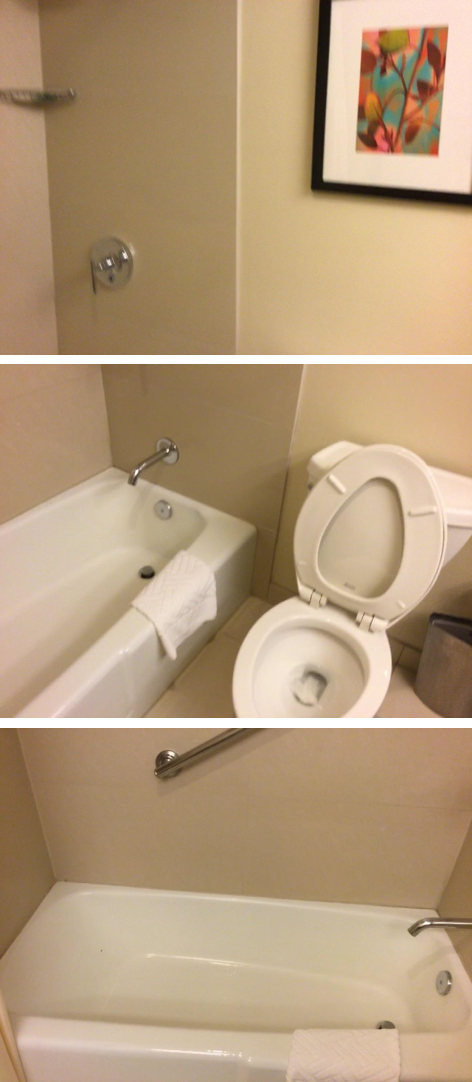} 
\includegraphics[width=65pt]{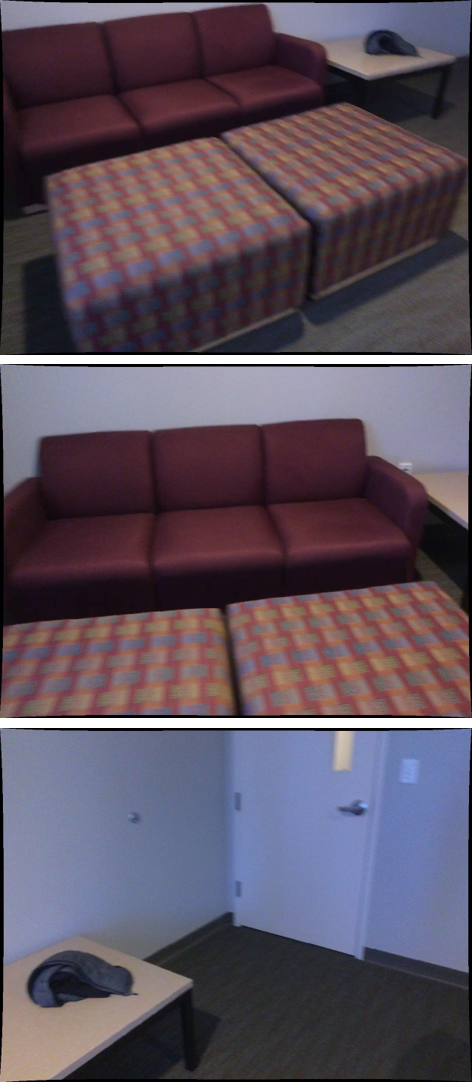}
\includegraphics[width=65pt]{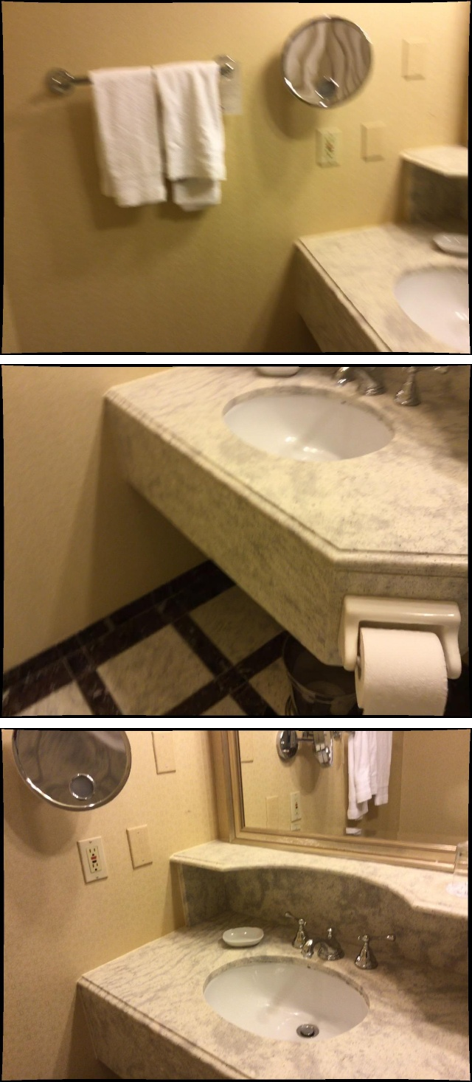} 
\includegraphics[width=65pt]{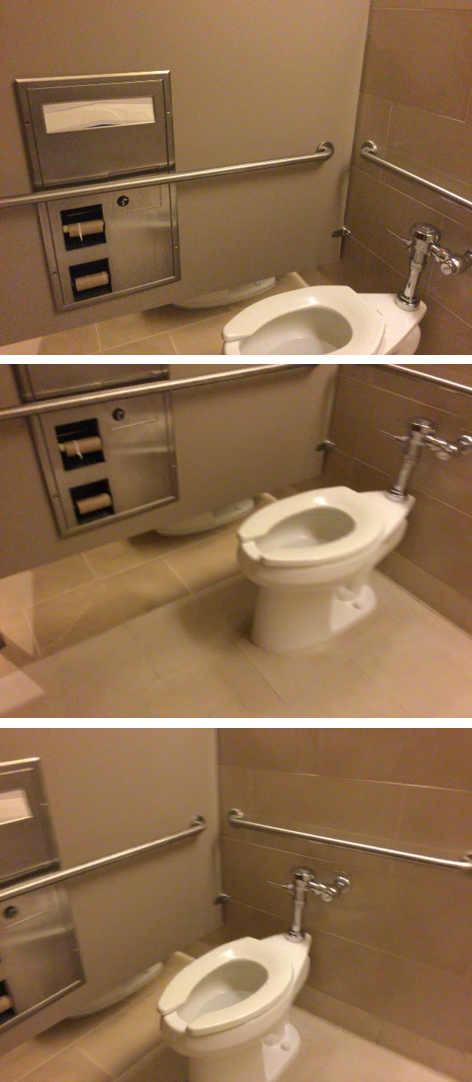}
\includegraphics[width=65pt]{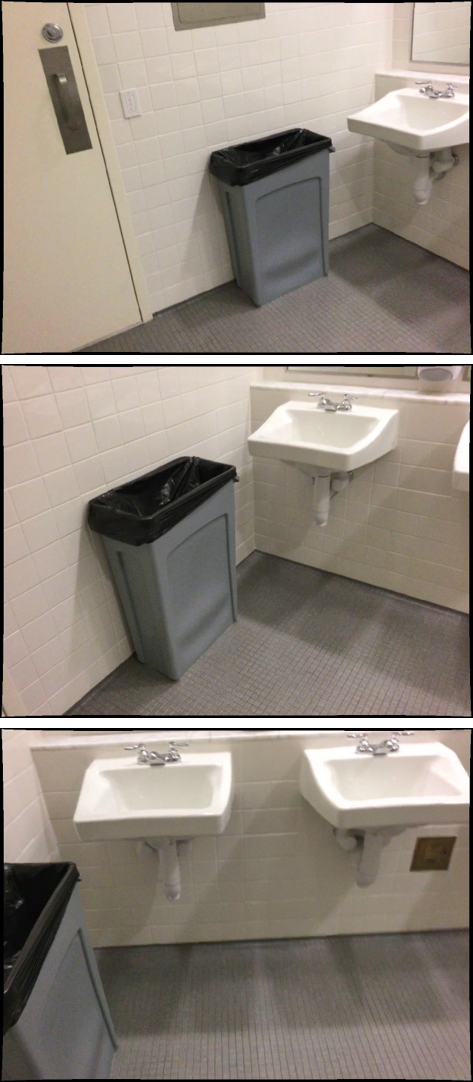}
\includegraphics[width=65pt]{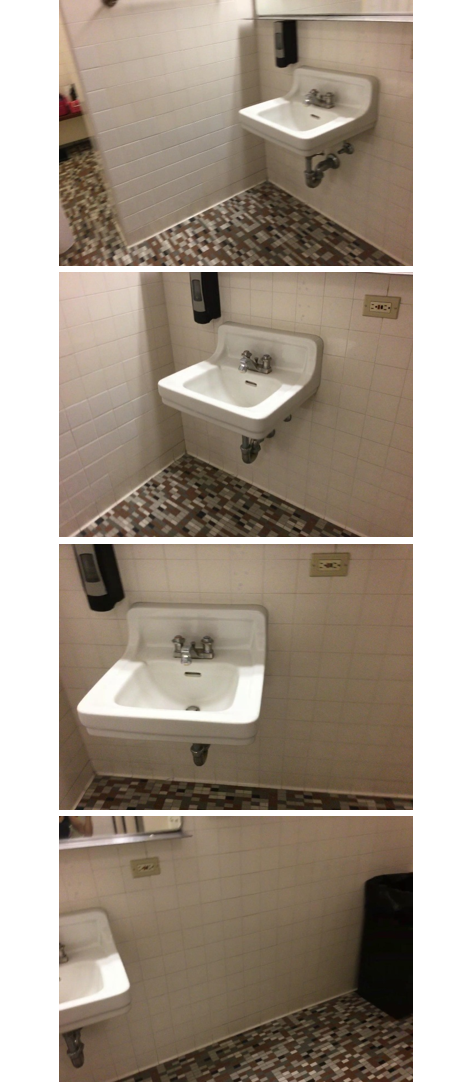} 
\includegraphics[width=65pt]{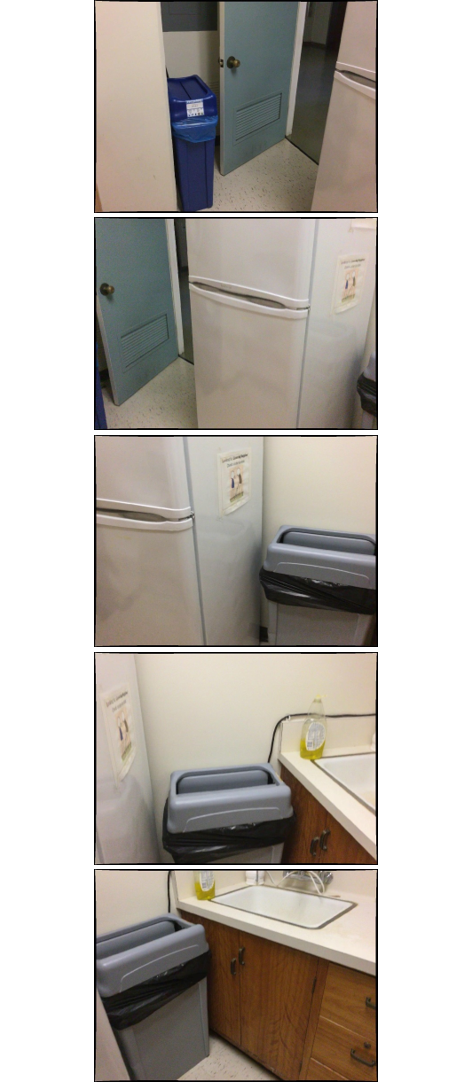}\\%
\raisebox{20pt}{\rotatebox[origin=c]{90}{\jiajia{NOPE-SAC}}}
 \includegraphics[width=65pt]{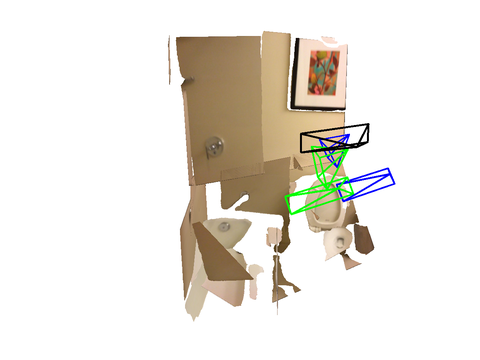}
        \includegraphics[width=65pt]{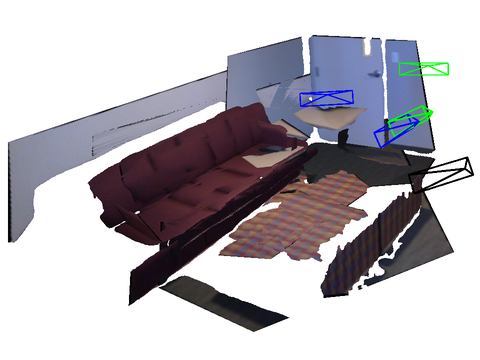}
        \includegraphics[width=65pt]{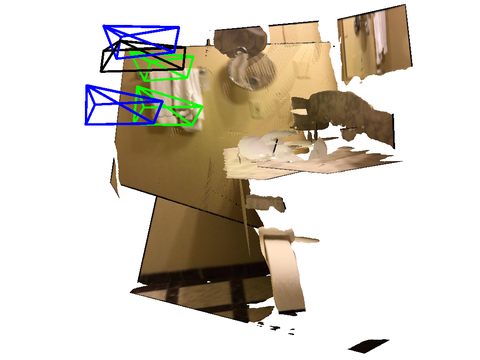}
        \includegraphics[width=65pt]{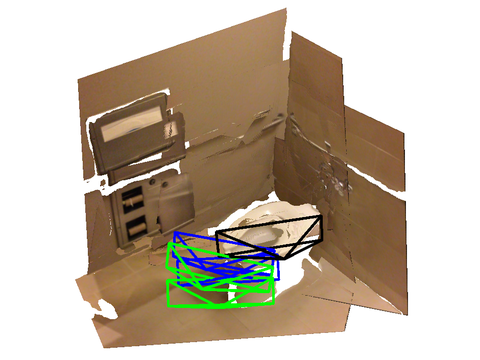}
        \includegraphics[width=65pt]{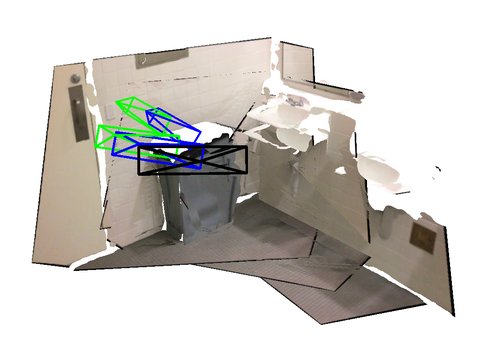}
        \includegraphics[width=65pt]{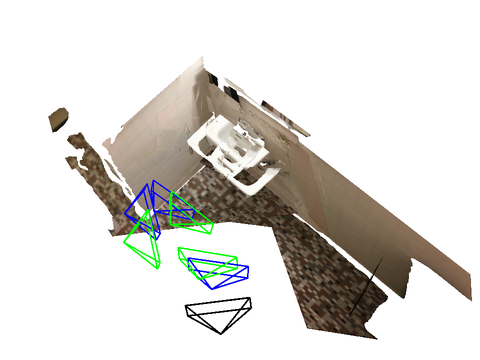}
        \includegraphics[width=65pt]{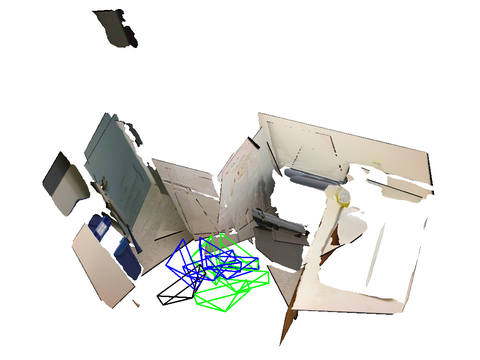}\\
\raisebox{20pt}{\rotatebox[origin=c]{90}{\jiajia{Ours}}}
\includegraphics[width=65pt]{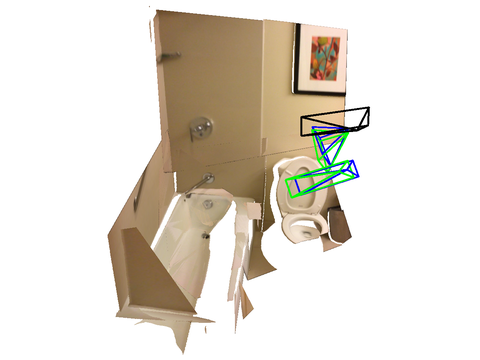}
        \includegraphics[width=65pt]{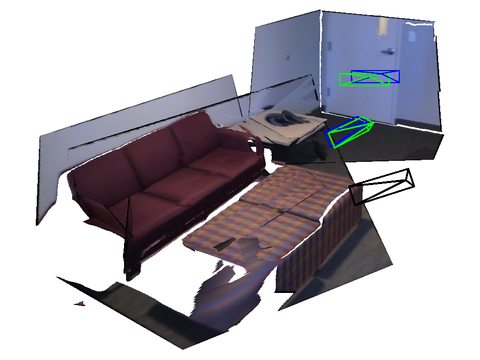}
        \includegraphics[width=65pt]{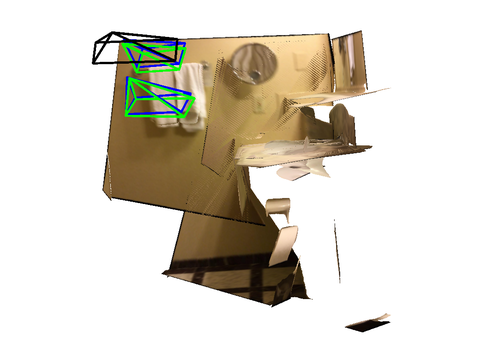}
        \includegraphics[width=65pt]{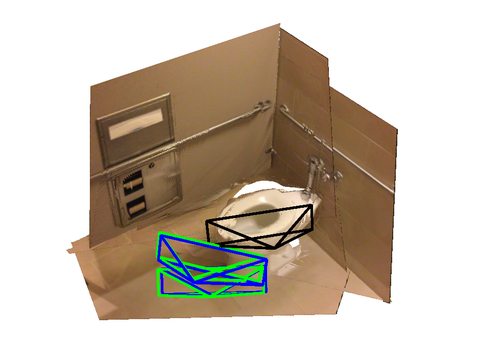}
        \includegraphics[width=65pt]{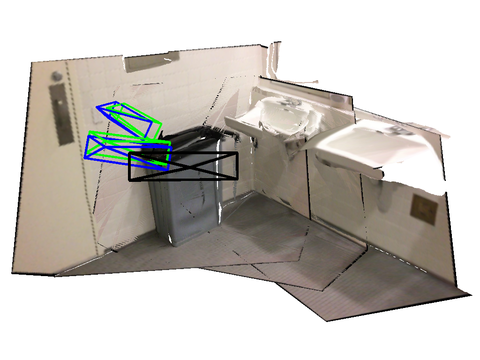}
        \includegraphics[width=65pt]{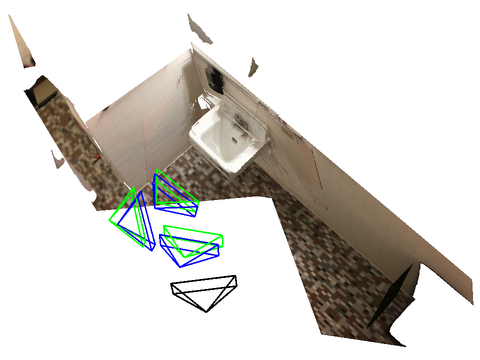}
        \includegraphics[width=65pt]{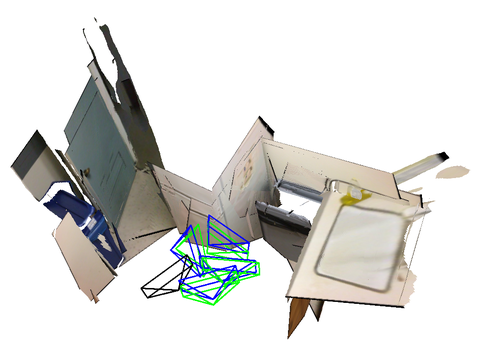}\\
\raisebox{20pt}{\rotatebox[origin=c]{90}{\jiajia{GT}}}
\includegraphics[width=65pt]{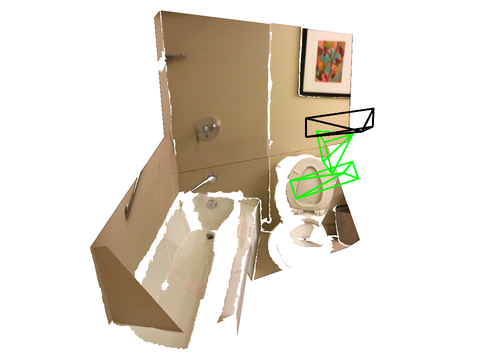}
        \includegraphics[width=65pt]{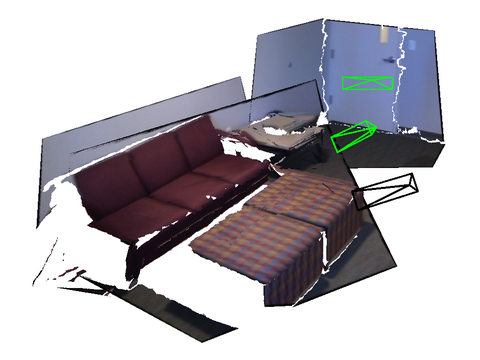}
        \includegraphics[width=65pt]{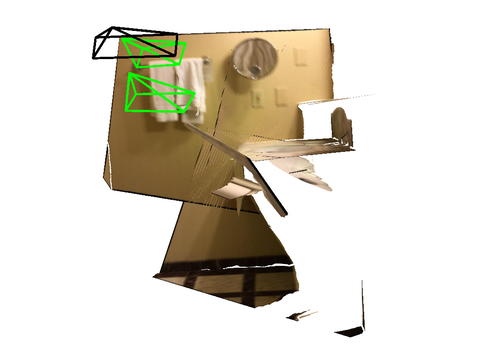}
        \includegraphics[width=65pt]{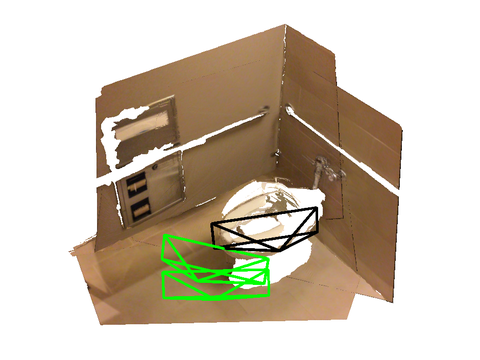}
        \includegraphics[width=65pt]{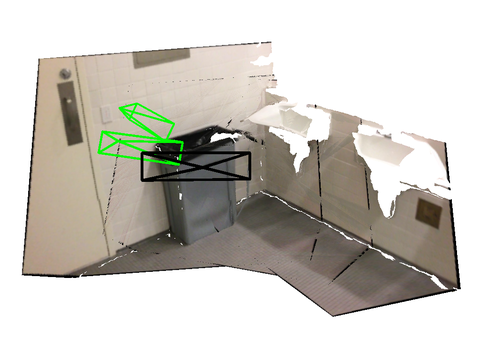}
        \includegraphics[width=65pt]{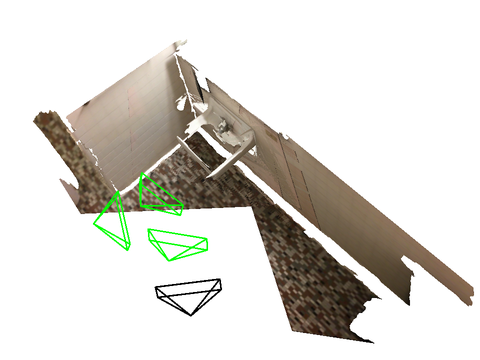}
        \includegraphics[width=65pt]{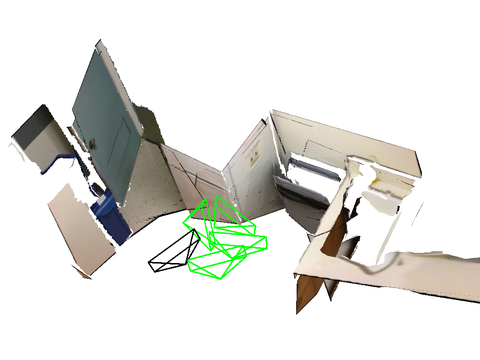}
\end{tabular}  
\vspace{-5pt}
\caption{\jiajia{Comparison of camera poses and 3D reconstruction with $\geq3$ views on the ScanNetv2 dataset. The entire reconstruction process of $n$ views is obtained by executing $n-1$ inferences between every two adjacent views. The fixed \textbf{Black} frustums show the camera of the first image. The \textbf{Green} frustums represent the ground truth camera of the other images while the \textbf{Blue} frustums depict the predicted results of the leading NOPE-SAC or ours.}}
\vspace{-6pt}
\label{fig:multiview_rec}
\end{figure*}

%% file: figures/tsne.tex
\begin{figure}[t]
  \centering
  \begin{overpic}[width=0.89\linewidth]{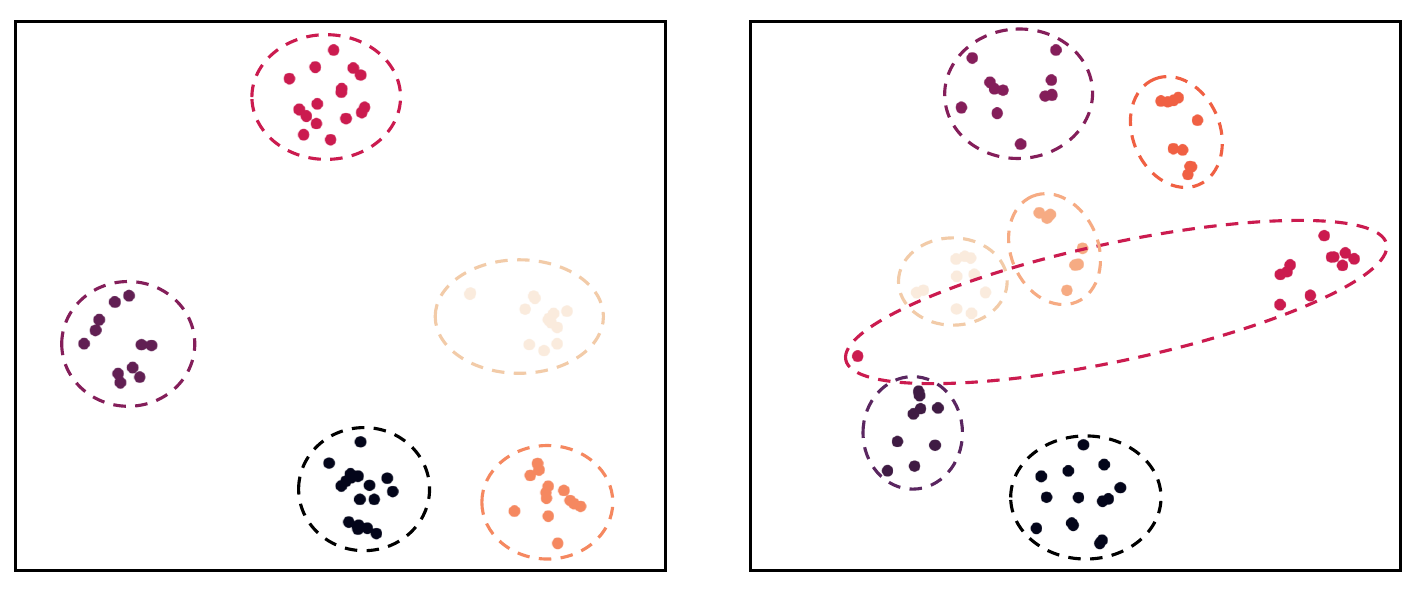}
  \end{overpic}
  \vspace{-5pt}
  \caption{Visualization of unified plane embeddings from multiple views. We sample frames from two scenes in the sparse view testing set and visualize the embeddings of frequently occurring plane instances for each scene using a t-SNE plot.   The consistent colors are assigned to represent the same instances of planes across the frames, which are enclosed within corresponding colored circles.}
  \vspace{-10pt}
  \label{fig:tsne}
\end{figure}

%% file: tables/corr_comparison.tex
\begin{table}
\centering
\caption{Comparison of Plane correspondence between supervised optimal transport (OT) using GT Correspondence, initial pose, and Ours. 
Their raw correspondence without post-filtering is denoted as "-R".
}
\resizebox{1.0\linewidth}{!}{ 
    \begin{tabular}{ccc|cccc}
    \toprule
        Method & Corr. Sup. & Init. Pose & Precision &Recall  &F-score &TP  \\ 
       \midrule

         \multicolumn{7}{c}{ScanNetv2 dataset} \\
        \midrule
        OT-R \cite{sarlin2020superglue, tan2023nopesac} &$\checkmark$ & & 0.305 &  0.438  & 0.359 & 8087 \\ 
        OT-R \cite{sarlin2020superglue, tan2023nopesac} &$\checkmark$ &$\checkmark$ & 0.443  &0.480 & 0.461 &8873  \\ 
        Ours (Monocular)-R &  & & 0.382 & 0.367 & 0.374 & 6786 \\ 
        Ours-R & & & \textbf{0.534} & \textbf{0.562} & \textbf{0.547} & \textbf{10390} \\ 
         \midrule
        
        OT \cite{sarlin2020superglue, tan2023nopesac} &$\checkmark$ &$\checkmark$    & 0.473    & 0.467   &  0.470   &8627  \\    
        Ours & & & \textbf{0.576} & \textbf{0.552} & \textbf{0.564}  & \textbf{10192} \\ 
        \midrule

         \multicolumn{7}{c}{MatterPort3D dataset} \\
        \midrule
        OT-R \cite{sarlin2020superglue, tan2023nopesac} &$\checkmark$ & &0.371 & 0.500  &0.426   & 21670 \\
        OT-R \cite{sarlin2020superglue, tan2023nopesac} &$\checkmark$ &$\checkmark$ & \textbf{0.499}    &  \textbf{0.515}   &    \textbf{0.507}  &  \textbf{22285} \\ 
        Ours (Monocular)-R & & & 0.316 &  0.283 &  0.298 & 12238 \\ 
        Ours-R & & & 0.456 &  0.509 & 0.481 & 22022 \\ 
         \midrule
        
        OT \cite{sarlin2020superglue, tan2023nopesac} &$\checkmark$ &$\checkmark$   & 0.531    &  \textbf{0.501}   &   \textbf{0.515}   & \textbf{21677}  \\     
        Ours & & &\textbf{0.540} &0.476 &0.506 & 20630 \\  
         \bottomrule
    \end{tabular}
}
\vspace{-5pt}
\label{tab:corr_compare}
\end{table}

%% file: conclusion.tex
\section{Conclusion}

We have presented PlaneRecTR++, a unified query learning framework, to learn robust 3D plane recovery and relative camera pose estimations. Through encoding planar attributes among unified latent embeddings, our method captures the correlations of diverse sub-tasks of 3D plane reconstruction using a single and compact Transformer architecture, and achieves state-of-the-art performance on four public benchmark datasets. Thanks to the tight interconnection of all sub-tasks, different from all existing multi-stage paradigms, our method realizes mutual benefits of coupled predictions in a single-shot prediction, and is able to automatically discover across-view plane correspondences, even without requiring any external initialization and correspondence supervisions. Furthermore, we have conducted extensive ablative experiments to demonstrate the efficacy of PlaneRecTR++.

\paraspace
\ptitle{Limitations and Future work.}
It still remains unexplored how to extend PlaneRecTR++ to efficiently process \jiajia{more} sequential images for a larger scale of \jiajia{planar} recontruction in an online fashion. An exciting future venue would be to leverage the learned correspondences for long-term plane-level tracking, possibly drawing recent advancements in Transformer-based video panoptic segmentation.

%% file: acknowledgement.tex

\ifCLASSOPTIONcompsoc
  \section*{Acknowledgments}
\else
  \section*{Acknowledgment}
\fi

We thank the anonymous reviewers and editors for their valuable comments. This work was supported in part by the NSFC (No. 62325211, No. 62132021, No. 62201603, No. 62571535), the HNNSF (No. 2025JJ40060), and the CPSF (No. 2023TQ0088, No. GZC20233539).

%% file: biography.tex

%
\vspace{-1.15cm}
\begin{IEEEbiography}[{\includegraphics[width=1in,height=1.25in,clip,keepaspectratio]{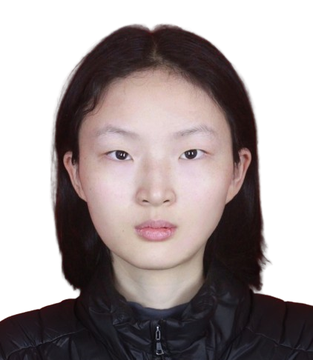}}]{Jingjia Shi} is a PhD student at the College of Computer, National University of Defense Technology (NUDT), Changsha, China. Her research interests focus on learning based 3D vision, including structure-aware reconstruction, pose estimation, and 3D representation learning.
\end{IEEEbiography}
\vspace{-1.15cm}
\begin{IEEEbiography}[{\includegraphics[width=1in,height=1.25in,clip,keepaspectratio]{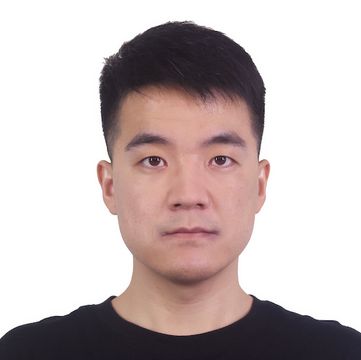}}]{Shuaifeng Zhi}  is currently a Lecturer (Assistant Professor) at the College of Electronic Science and Technology, National University of Defense Technology (NUDT), Changsha, China. He received his Ph.D. degree in Computing Research at the Dyson Robotics Laboratory, Imperial College London, UK, in 2021. He was a 6-month visiting student in 5GIC, University of Surrey, UK, in 2015. His current research interests focus on robot vision, particularly on scene understanding, neural scene representation, and semantic SLAM. He also serves on the editorial board of The Visual Computer.
\end{IEEEbiography}
\vspace{-1.15cm}
\begin{IEEEbiography}[{\includegraphics[width=1in,height=1.25in,clip,keepaspectratio]{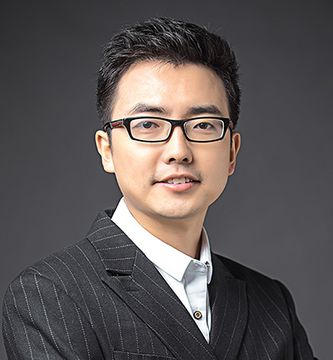}}]{Kai Xu}
is a Professor at the College of Computer, NUDT, where he received his Ph.D. in 2011. He conducted visiting research at Simon Fraser University and Princeton University. His research interests include geometric modeling and shape analysis, especially on data-driven approaches to the problems in those directions, as well as 3D vision and its robotic applications. He has published over 80 research papers, including 20+ SIGGRAPH/TOG papers. He has co-organized several SIGGRAPH Asia courses and Eurographics STAR tutorials. He serves on the editorial board of ACM Transactions on Graphics, Computer Graphics Forum, Computers \& Graphics, and The Visual Computer. He also served as program co-chair of CAD/Graphics 2017, ICVRV 2017 and ISVC 2018, as well as PC member for several prestigious conferences including SIGGRAPH, SIGGRAPH Asia, Eurographics, SGP, PG, etc. His research work can be found in his personal website: www.kevinkaixu.net.
\end{IEEEbiography}


